\documentclass[accepted]{main}

\makeatletter \let\NAT@parse\undefined \makeatother  

\usepackage{amsmath,amssymb,amsfonts}  
\usepackage{mathtools}                 
\usepackage{pifont}                    
\usepackage{textcomp}                  
\usepackage{graphicx}                  
\usepackage{tikz}                      
\usepackage{algorithm,algorithmic}     
\usepackage{xcolor}                    
    \definecolor{links}{HTML}{1462e0}
    \definecolor{urls}{HTML}{1462e0}
    \definecolor{cites}{HTML}{1462e0}
    \hypersetup{colorlinks, linkcolor=links, citecolor=cites, urlcolor=urls}
\usepackage{booktabs}                  
\usepackage{xtab}                      
\usepackage{multirow}                  
\usepackage{threeparttable}            
\usepackage{cite}                      
\usepackage{hyperref}                  
\usepackage{csquotes}                  
\usepackage{soul}                      
\usepackage{comment}                   
\usepackage{arydshln}                  
\usepackage[normalem]{ulem}            
\usepackage{subcaption}
    \DeclareCaptionLabelSeparator{period}{. }
\begin{document}

\title{A Vector-Quantized Foundation Model for Patient Behavior Monitoring}

\author[1,2]{Rodrigo~Oliver}
\author[1,2]{Josu\'e~P\'erez-Sabater}
\author[1,2]{Leire~Paz-Arbaizar}
\author[3]{Diego~Herrero-Quevedo}
\author[1,2,4,5]{Antonio~Art\'es-Rodr\'iguez}
\author[1,2]{Alejandro~Lancho}
\author[1,2]{Pablo~M.~Olmos}

\affil[1]{Signal Theory and Communications Department\\
    University Carlos III of Madrid\\
    Legan\'es, Spain}
\affil[2]{Instituto de Investigaci\'on Sanitaria Gregorio Mara\~n\'on\\
    Madrid, Spain}
\affil[3]{Center for Data Science\\
    New York University\\
    New York, NY, USA}
\affil[4]{Evidence-Based Behavior\\
    Legan\'es, Spain}
\affil[5]{CIBERSAM, ISCIII\\
    Madrid, Spain}

\maketitle

\begin{abstract}
\textbf{Foundation models have achieved remarkable success across various domains, yet their adoption in healthcare remains limited. While significant advances have been made in medical imaging, genetic biomarkers, and time series from electronic health records, the potential of foundation models for patient behavior monitoring through personal digital devices remains underexplored. The data generated by these devices are inherently heterogeneous, multisource, and often exhibit high rates of missing data, posing unique challenges. This paper introduces a novel foundation model based on a modified vector quantized variational autoencoder, specifically designed to process real-world data from smartphones and wearable devices. We leveraged the discrete latent representation of this model to effectively perform two downstream tasks, suicide risk assessment and emotional state prediction, on different held-out clinical cohorts without the need of fine-tuning. We also highlight the existence of a trade-off between discrete and continuous latent structures, suggesting that hybrid models may be optimal for balancing accuracy across various supervised and unsupervised tasks.}
\end{abstract}



\section{Introduction} \label{sec:introduction}

The advent of foundation models (FMs) has catalyzed transformative advancements across various domains, from natural language processing to computer vision, achieving remarkable generalization across diverse tasks \cite{bommasani2021}. However, their integration into healthcare has been comparatively slower. This delay can be attributed to clinical data's inherent complexity and variability and the challenges posed by heterogeneous, high-dimensional, and often incomplete datasets, such as electronic health records (EHR) \cite{Moor23}. Moreover, FMs introduce significant challenges in terms of privacy, validation mechanisms, and overconfidence.

An underexplored but crucial area in healthcare is the analysis of time-series data from mobile phones and wearable devices, which are increasingly used in daily life and provide a vast amount of data \cite{Rat24}. This has enabled the passive collection of behavioral metrics, such as the pattern of mobile apps used, distance traveled, time spent at home, and sleep patterns, among others. This method, known as digital phenotyping (DP), allows for continuous, unobtrusive monitoring without requiring active user input, making it ideal for long-term monitoring \cite{insel_2017}. These data have proven valuable for characterizing and tracking psychiatric patients \cite{Moreno2020,Romero2023,Buscher2024-uh}. Recent research has applied DP to detect behavioral shifts that may indicate serious mental health risks \cite{Berrouiguet2019}.

Behavioral data from DP devices presents several challenges: it is multisource (e.g., heart rate, motion, sleep patterns), heterogeneous (coming from different sensors with varying formats and time scales), and often incomplete, with significant portions missing due to device issues or user behavior \cite{Wu22, Lin20}. Importantly, these missing data points might hold valuable insights into patient behavior, so properly modeling them is crucial \cite{barrejon_2022}. For instance, a wearable device that stops collecting data intermittently during certain times may indicate behavioral patterns such as sleep disturbances or irregular daily routines relevant to mental health monitoring.

The development of FMs specific for behavioral data from smartphones and wearable sensors is just commencing to emerge \cite{narayanswamy_2024,yuan_2024}. Expanding the research on this field is the primary contribution of our work. We demonstrate that state-of-the-art FMs for time series can struggle to handle the complexity of such data and fail to fully capture the rich information embedded within these datasets. In particular, we show that the dominant approach for designing time series foundation models (TSFMs)—based on autoregressive transformers with continuous embeddings, which have proven effective in downstream tasks like sentiment analysis \cite{bashiri_2024}—is inadequate for the unsupervised detection of statistical changes in the embedding spaces from an individual’s recent history. This limitation is particularly critical in fields like computational psychiatry, where identifying behavioral shifts is essential.

\begin{figure*}
    \centering
    \includegraphics[width=\textwidth]{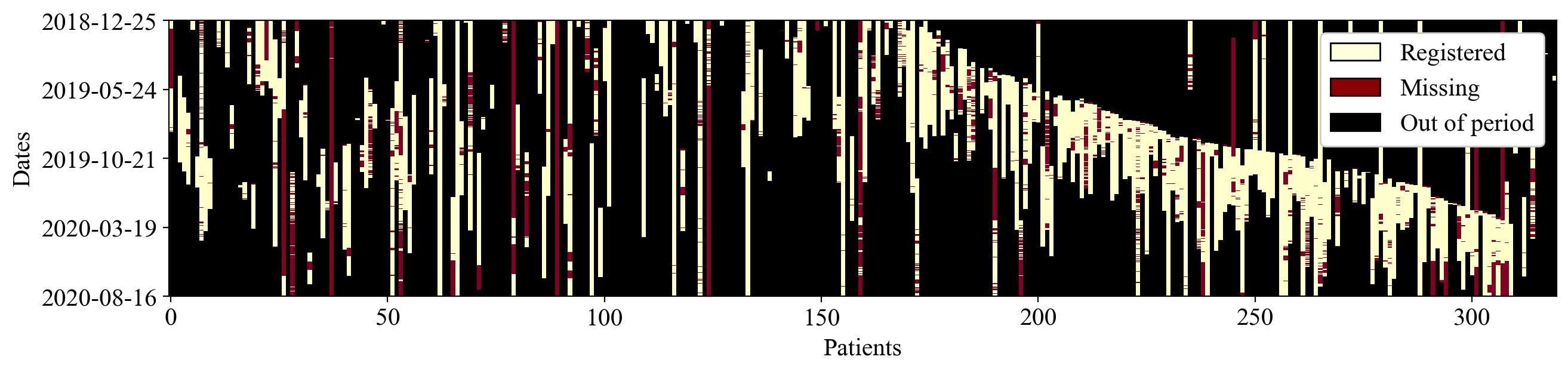}
    \caption{Visualization of data missingness. The availability of step count data is shown over approximately 18 months. The length of registered periods varies from patient to patient, and most contain scattered days or sequences with no data.}
    \label{fig:missingness_graph}
\end{figure*}

Discrete representations have proven effective in enhancing interpretability and capturing distinct patterns, which is particularly valuable in applications where human understanding of the model’s outputs is critical \cite{jin_2020,miao_2017}. Variational autoencoders can be leveraged with vector quantization and nearest-neighbor lookup to map features into discrete latent vectors, effectively storing relevant information and capturing complex relationships within the data \cite{van2018neural}. This approach is especially beneficial when representing discrete states, such as varying health conditions or behavioral patterns. In this work, using a dataset comprising different clinical cohorts, we demonstrate that a vector quantized-variational autoencoder (VQ-VAE), trained as a TSFM via an auto-encoding self-supervision mechanism to impute missing data—a common challenge in data collected from wearable and mobile devices—can successfully perform both supervised and unsupervised detection of behavioral changes. Our model, termed VQ-TSFM and pretrained to reconstruct multisource, heterogeneous time-series data, is designed to model missingness patterns and provides a discrete latent codebook that enables successful performance on downstream medical tasks without task-specific training.

Our model's capabilities are demonstrated across two distinct tasks, highlighting its potential in personalized healthcare. One critical application is unsupervised suicide risk assessment, where our approach achieves an AUC of 0.87 in predicting suicidal events through a change-point detection algorithm \cite{Adams2007} applied to the pretrained latent codebook of VQ-TSFM. This significantly advances previous methodologies using a patient-specific heterogeneous mixture model (HetMM) \cite{barrigon_2023}, which, despite an AUC of 0.88, suffers from scalability and efficiency issues due to its per-individual approach. The VQ-TSFM offers a more scalable solution by extracting patient profiles across a population with a single model.

We also showcase the VQ-TSFM's utility in a supervised task: predicting emotional states (positive, neutral, and negative). This aligns with ongoing research \cite{arbaizar2025} on the potential of transformer-based approaches for emotion forecasting using passive behavioral data. We compare VQ-TSFM with a continuous Informer-based TSFM (I-TSFM) that achieves high accuracy and a ROC AUC of 0.988 for emotional valence classification. While introducing quantization, our VQ-TSFM still reached an AUC of 0.909, demonstrating strong performance and indicating its capacity for accurate emotion prediction.

Our experiments uncover a previously unreported trade-off: while the continuous I-TSFM exhibits strong predictive performance on this task, the VQ-TSFM requires increasing the VQ-VAE resolution (i.e., expanding the discrete alphabet and embedding dimensions) in order to enhance its predictive accuracy and perform closer to the I-TSFM. However, this improvement comes at the cost of degraded CPD performance, as detecting statistical changes becomes more challenging. This tension between supervised and unsupervised tasks suggests that future FMs for general artificial intelligence may benefit from integrating hybrid discrete-continuous structures to balance accuracy across diverse applications.

\section{Behavioral Dataset} \label{sec:dataset}

The data used in this study was collected through a DP-enabled mobile application provided by the Evidence-Based Behavior (eB2) company\footnote{https://eb2.tech/.}, and they comprise a total of 5,532 patients enrolled across 39 independent clinical programs. The collection of datasets contains data acquired across a diverse array of devices, including smartphones from multiple manufacturers, smartwatches, and fitness bands. Notably, the majority of participants used their own personal device, thereby mitigating potential device-related biases. On the other hand, while the studies were primarily oriented to healthcare settings, they present unique selection criteria and thus the collective participant sample exhibits significant heterogeneity in terms of clinical context and target populations. This heterogeneity is reflected in a substantial variability in socio-demographic backgrounds (sex, age, physical and mental autonomy, etc.) and a wide range of clinical conditions, including different mental health disorders, oncology, eating disorders, and cognitive impairment. Thus, each program conforms an independent dataset characterized by its own behavioral distributions and hence can be considered as different domains. The collection of all datasets comprises 1,122,233 entries across 64 variables, spanning from January 1, 2016, to March 13, 2024. Each entry encapsulates aggregated daily metrics from original time-stamped recordings captured at 30-minute intervals across multiple sensors. \autoref{tab:missingness} overviews the variables selected to train the VQ-VAE, their types, and the corresponding missingness rates.

\begin{table}[t]
    \centering
    \small
    \begin{threeparttable}
    \caption{Type and relative missingness of selected variables}
    \label{tab:missingness}
    \begin{tabular}{@{}llcc@{}}
    \toprule
    \textbf{Category} & \textbf{Variable name} & \textbf{Type} & \textbf{Missing rate(\%)} \\ \midrule
    \textbf{Activity} & Time Walking (s) & $\mathbb{R}_{\geq 0}$   & 62.79   \\
                      & App Usage (s)  & $\mathbb{R}_{\geq 0}$  & 83.15   \\
                      & Practiced Sport\tnote{a}  & $\{0,1\}$   & 0.00    \\
                      & Total Steps & $\mathbb{N}_0$ & 55.30   \\ \midrule
    \textbf{Location} & Location Clusters\tnote{b} & $\mathbb{N}_0$ & 72.53   \\
                      & Distance (m)  & $\mathbb{R}_{\geq 0}$ & 73.01   \\
                      & Time at Home (m) & $\mathbb{R}_{\geq 0}$ & 82.53   \\ \midrule
    \textbf{Other}    & Weekend\tnote{c}  & $\{0,1\}$        & 0.00    \\ \midrule
    \textbf{Sleep}    & Sleep Duration (s) & $\mathbb{R}_{\geq 0}$    & 66.76   \\
                      & Sleep Start (s)\tnote{d}   & $\mathbb{R}$     & 66.11   \\ \bottomrule
    \end{tabular}
    \begin{tablenotes}
        \item[a] Sports activity is flagged if the combined time spent walking, running, bicycling, and other sports exceeds one hour.
        \item[b] Locations are dynamically defined by clustering algorithms grouping related geographical positions.
        \item[c] $1$ represents weekend data, whereas $0$ represents weekday data.
        \item[d] The reference time is 23:00. Negative values indicate seconds before this time, and positive values indicate seconds after.
        \end{tablenotes}
    \end{threeparttable}
\end{table}

A common challenge in studies involving digital phenotyping is missing data, often caused by smartphone operating systems terminating background processes or patients intentionally discontinuing the use of their digital devices. These disruptions, essential for passive data collection, result in significant gaps in the data stream, compromising the quality and completeness of the dataset (see \autoref{fig:missingness_graph} for a representative example). To address this, we focused on a subset of variables with a missingness rate below 85\%. Additionally, the collected data are heterogeneous: some variables are recorded as daily summaries with limited dimensions (e.g., variable sleep is encoded in duration, start time and end time), whereas others provide more granular, time-segmented information, such as physical activity or app usage time. The dataset also contains significant noise and outliers, likely due to sensor malfunctions, inconsistent user behavior, environmental factors, and hardware or software issues. A detailed description of the dataset and its preprocessing is provided in \autoref{app:vqvae_preprocessing} of the supplementary material.

\section{Foundation Models for Behavioral Time Series} \label{sec:fm}

We now describe the FMs that were pre-trained over the collection of multiple DP datasets.

\subsection{I-TSFM: A Transformer Approach} \label{sec:fm_informer}

The transformer model we employ as a baseline follows an encoder-decoder architecture, leveraging the efficiency of the Informer model for time series forecasting \cite{Zhou21}. The model is pretrained through next token prediction (NTP), where the output is compared to the actual output shifted one day ahead. A hyperparameter grid was used to test various configurations over the NTP loss, with the best-performing model, trained on 50-day sequences (30 days for the encoder and 20 days for the decoder), utilizing an embedding dimension of 64, 8 attention heads, 3 layers, a feedforward dimension of 256, and a dropout rate of 0.3. Once trained, the model can predict future days autoregressively.

The architecture adheres to the encoder structure introduced in Informer \cite{Informer21}, integrating the ProbSparse self-attention mechanism to optimize computational complexity from $O\left(L^2\right)$ to $O(L \log L)$. Unlike the original Informer model, this implementation maintains the full sequence length to retain detailed temporal dependencies, allowing to preserve high-fidelity sequence dependency alignment. The decoder is based on the standard transformer architecture \cite{NIPS2017_3f5ee243}, consisting of a masked self-attention mechanism for autoregressive constraints (prevents attending to future positions), a cross-attention mechanism for encoder-decoder information exchange, and a feed-forward network for feature transformation into the final output.

To feed the heterogeneous data streams with missing entries to the Informer network, time-series embeddings are obtained using a heterogeneous hidden Markov model (het-HMM) \cite{moreno2022pyhhmm}. This model is capable of handling both continuous and categorical features and is employed to address the problem of missing data by marginalizing over unobserved values. Once trained, the het-HMM provides posterior probabilities over the hidden states of each day. These vectors serve as time-embeddings and are fed to the Informer architecture, allowing the system to infer a representation from the observed behavioral patterns of a given day. The employed het-HMM consists of seven latent states, selected based on the Bayesian information criterion.

\begin{figure*}[t]
        \centering
        \includegraphics[width=\textwidth]{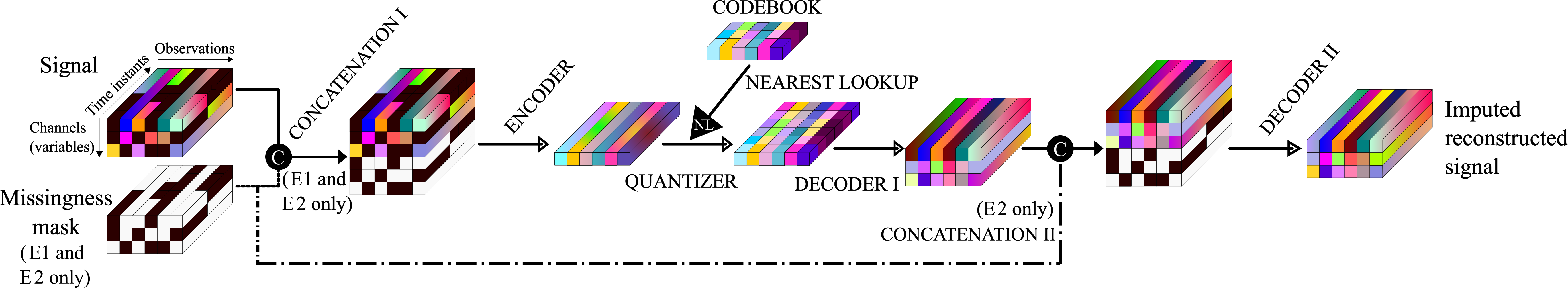}
        \caption{Overview of the VQ-TSFM structure. The complete set corresponds to the extended version E2 of the VQ-TSFM. Model E1 only features encoder conditioning and VQ-TSFM does not present any missingness mask concatenations, operating solely on the signal.}
        \label{fig:vqvae_overview}
\end{figure*}

As a result, the temporal sequence used as input to the model has dimensions $L \times S$, where $L$ represents the total sequence length and $S$, the probability for each of the hidden states. The input for the transformer encoder and decoder were reshaped into observation batches with dimensions $\left[B,\, L_e,\, S\right]$ and $\left[B,\, L_d,\, S\right]$, where $B$ denotes the batch size, $L_e$ and $L_d$ the encoder and decoder sequence lengths, respectively, and $S$ is the feature dimension. The architecture of the transformer model is described in \autoref{tab:transformer_architecture} of the Supplementary Material.

\subsection{Our VQ-TSFM: A Quantized VAE Approach} \label{sec:fm_vqvae}

The vector quantized-variational autoencoder \cite{van2018neural} extends the traditional VAE by incorporating a discrete latent space, addressing some of the limitations of continuous representations. In VQ-VAE, the latent space is composed of $K$ discrete embeddings, $\mathbf{e}_j \in \mathbb{R}^D$, where $D$ is the embedding dimension and $j \in \{1, 2, \ldots, K\}$, forming the codebook $E = \{\mathbf{e}_j\}_{j=1}^K$. The encoder produces a continuous latent output $\mathbf{z}_e(\mathbf{x})$, which is quantized to the nearest embedding $\mathbf{e}_k$, with $k=1,\dots,K$, using nearest-neighbor lookup:
\begin{equation}
    \label{eq:nearest_lookup}
    q(z = k|\mathbf{x}) = 
    \begin{cases} 
    1 & \text{for } k = \arg\min_j \left\|\mathbf{z}_e(x) - \mathbf{e}_j\right\|_2 \\
    0 & \text{otherwise}
    \end{cases}
\end{equation}
where $z=k$ indicates that $\mathbf{z}_q(\mathbf{x}) = \mathbf{e}_k$ and $\mathbf{z}_q(\mathbf{x})$ denotes the decoder input. The loss function takes the form
\begin{multline}
    \label{eq:vanilla_vqvae_loss}
    L = \underbrace{\log p(\mathbf{x}|\mathbf{z}_q(\mathbf{x}))}_{\text{Reconstruction loss}}
    + \underbrace{\left\|\text{sg}\left[\mathbf{z}_e(\mathbf{x})\right] - \mathbf{e}_k\right\|_2^2}_{\text{Codebook loss}} \\
    + \beta \underbrace{\left|\mathbf{z}_e(\mathbf{x}) - \text{sg}\left[\mathbf{e}_k\right]\right\|_2^2}_{\text{Commitment loss}},
\end{multline}
where $\text{sg}[\cdot]$ denotes the stop-gradient operator. The reconstruction loss is optimized by both the encoder and the decoder, forcing them to provide relevant data representations. The codebook loss ensures that the embeddings capture such representations. The commitment loss enforces stability during training by limiting the updates in the encoder output to match current embeddings. As described in \cite{van2018neural}, the codebook loss can be replaced by exponential moving averages of $\mathbf{z}_e(x)$, which is the implementation used for the experiments in this work.

\textbf{Missing-aware VQ-VAE architectures.}
We propose three variants of VQ-VAE to handle missing data in multivariate time series. Our primary model is an implicit mask variant (VQ-TSFM), which learns to represent missingness without feeding the binary mask into the network explicitly. Empirically, VQ-TSFM achieves reconstruction and imputation performance on par with architectures that take the mask as an additional input, so we adopt it as our default. To verify and validate VQ-TSFM's robust behavior, we also implement two explicit mask variants---VQ-TSFM E1 and VQ-TSFM E2 (hereafter referred to as E1 and E2, respectively)---described below.

Let $\mathbf{x}_d^{(i)} \in \mathbb{R}^T$ represent the time-series data vector of length $T$ for patient $i$ and variable $d$, where each component corresponds to a data entry in a sampled time instant and $d\in\{1,\ldots,D\}$. Recall that the set of possible variables is summarized in \autoref{tab:missingness}.  Let $\mathbf{m}_d^{(i)} \in \{0,1\}^T$ denote a binary mask vector where each entry indicates whether the corresponding entry is observed (entry value equal to $1$) or missing (entry value equal to $0$). The corrupted signal, after applying the binary mask, $\mathbf{m}_d^{(i)}$, is defined as:
\begin{equation}
    \tilde{\mathbf{x}}_d^{(i)} = \mathbf{m}_d^{(i)} \odot \mathbf{x}_d^{(i)}, 
\end{equation}
where $\odot$ denotes the element-wise product. When fed to the encoder $\mathbf{z}_e(\cdot)$, this formulation applies zero-imputation, ensuring missing data points do not introduce misleading information, as gradients related to imputed values remain zero during backpropagation \cite{NAZABAL2020107501}.

\begin{figure*}[t]
    \centering
    \begin{subfigure}[b]{0.45\linewidth}
        \includegraphics[width=\linewidth]{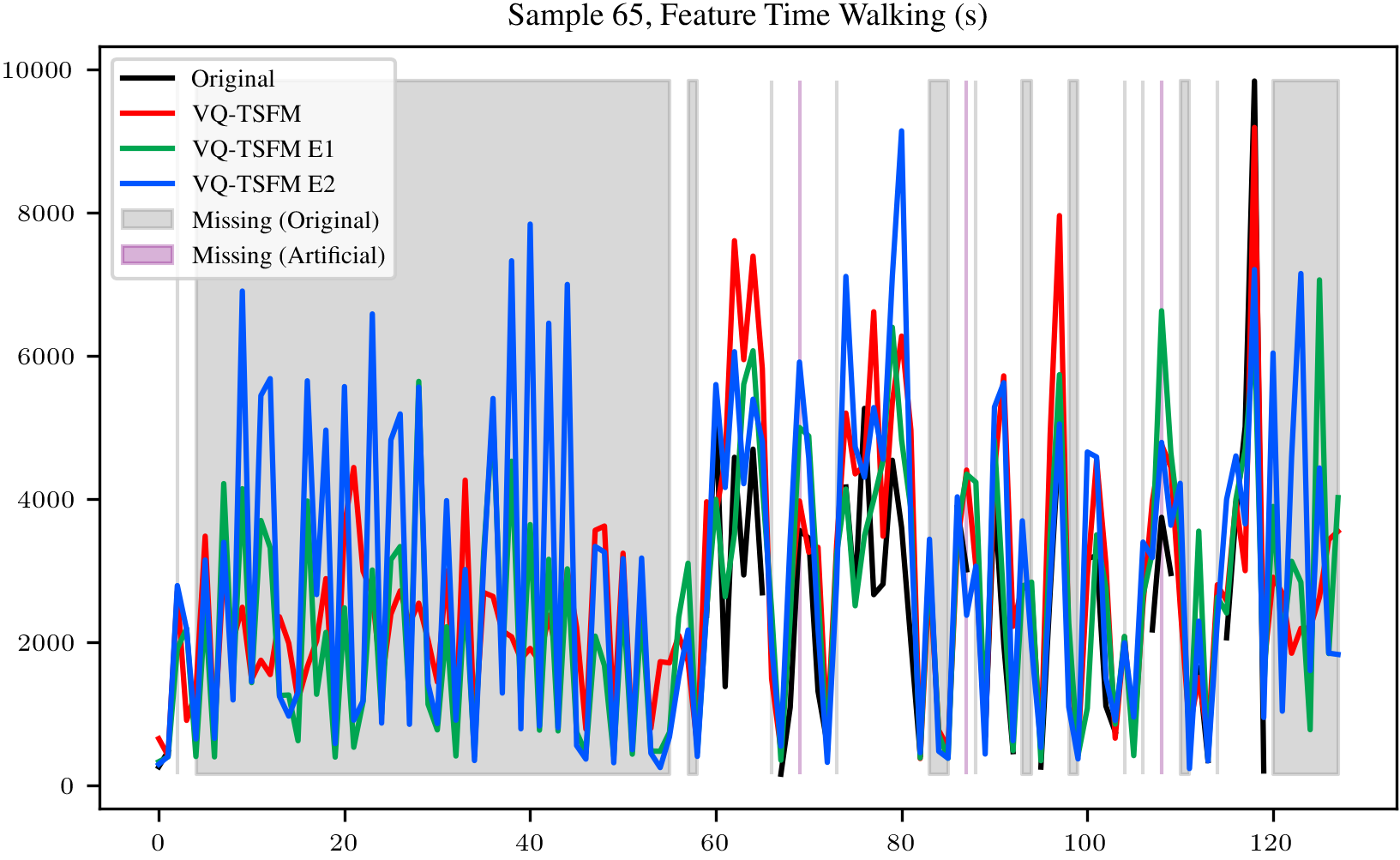}
        \caption{Reconstruction of sample 65 for Time Walking.}
        \label{fig:recon3}
    \end{subfigure}
    \begin{subfigure}[b]{0.45\linewidth}
        \includegraphics[width=\linewidth]{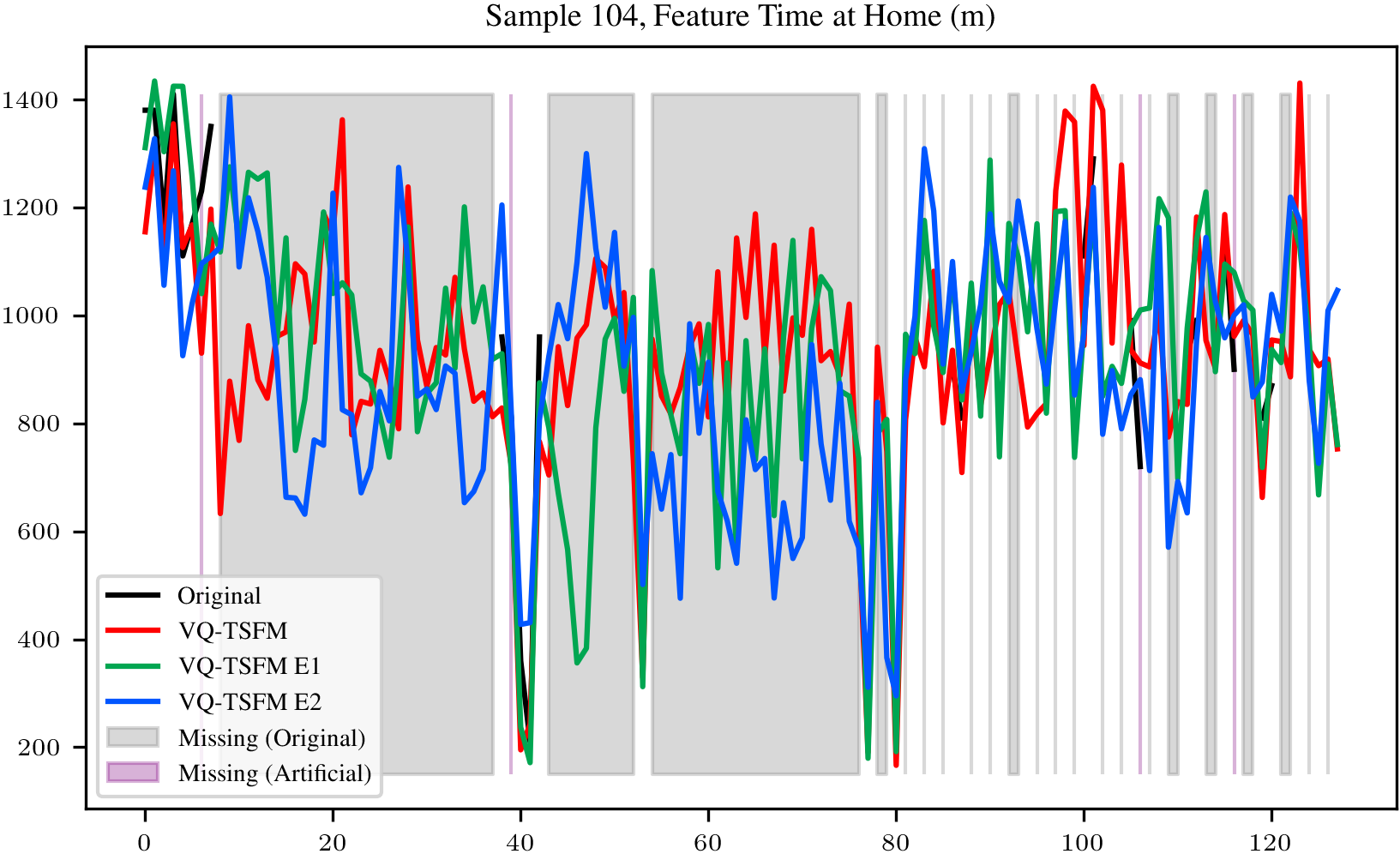}
        \caption{Reconstruction of sample 104 for Time at Home.}
        \label{fig:recon4}
    \end{subfigure}
    
    \caption{Representative signal reconstructions for observed and imputed instances. In cases where the original signal is not explicitly shown, it is because one or more of the models (whose reconstructions are plotted) overlap the true signal precisely, obscuring the original data. Additional signal reconstructions for other data types are available in Appendix \ref{app:signal_recons_imput}.}
    \label{fig:signal_reconstructions}
\end{figure*}

Inspired by \cite{collier2021vaes}, we propose two variants to explicitly incorporate the missing mask within the VQ-TSFM structure (see \autoref{fig:vqvae_overview} for a joint overview and \autoref{fig:vqvae_model_architectures} in \autoref{app:vqvae_details} of the supplementary section for specific descriptions): (i) Model {E1}: Missingness mask conditioning in the encoder only; (ii) Model {E2}: Missingness mask conditioning in both encoder and decoder. In contrast, the implicit VQ-TSFM follows a simpler architecture with no missingness mask conditioning in either the encoder or decoder stages (only the input signal is processed). As a result, VQ-TSFM relies solely on the zero-imputed signal.

In models E1 and E2, both the input signal and missingness mask are integrated within the encoder. The missingness mask is preprocessed through $M$ convolutional layers, which allow the model to capture dependencies in the missing data patterns across variables. The processed mask is concatenated with the input signal along the channel axis, and the combined data is passed through $N$ convolutional layers, resulting in a continuous latent representation. This latent representation is then quantized via a nearest-neighbor lookup in the codebook before being passed to the decoder.

In version E1, the quantized embeddings are further processed through $O$ deconvolutional layers, followed by variable-specific activation functions tailored to the data type. In contrast, model E2 employs a more complex structure: the quantized embeddings are concatenated with the separately processed missingness mask (which is transformed via $L$ convolutional layers) along the channel axis before passing through additional $P$ convolutional layers. The output is fed into variable-specific activation functions.

We trained the models on the behavioral dataset described in \autoref{sec:dataset}. Each data modality was modeled by selecting an appropriate likelihood function tailored to its distributional characteristics. For real-valued variables, we employed a Gaussian likelihood, whereas for binary features, a Bernoulli likelihood was used. Count data were presented over a sufficiently extended array of values, and the Gaussian likelihood was also applied to them. For more information on data preprocessing, see \autoref{app:vqvae_preprocessing}.

\textbf{Self-supervision through missing data imputation.} Models were trained according to their reconstruction performance on observed entries only, without explicit missing-value objectives, and they were evaluated on their ability to impute artificially-introduced missing data. This approach prioritizes the quality of reconstructing available data without explicitly optimizing for imputing missing values. Consequently, evaluating their performance on data imputation under various missingness mechanisms provides a more rigorous test of their generalization capabilities in handling unobserved data, which they were not directly trained to predict.

We assessed the models' performance on both reconstruction and imputation tasks, which are crucial for evaluating their effectiveness in scenarios involving both observed and unobserved data. Reconstruction refers to recovering known values based on latent representations, whereas imputation involves estimating values that were not observed during training. For the imputation task, the models were exposed to synthetic missingness, simulating both missing completely at random (MCAR) and missing not at random (MNAR) mechanisms. In the MCAR setting, missing instances were introduced uniformly at random, whereas in the MNAR scenario, missingness was conditioned on the values of the target variables. This setup provides a comprehensive evaluation of the models' capabilities in both random and structured missingness settings.

\autoref{fig:signal_reconstructions} presents two examples of signal reconstructions for both observed and imputed instances. These visualizations highlight the studied VQ-TSFM models' ability to accurately recover data. Additional signal reconstructions and performance metrics showing results on reconstruction and imputation quality are provided in Appendix \ref{app:signal_recons_imput} due to space constraints. Furthermore, our results show that the codebook usage per sample is usually very sparse for most patients, as can be checked in Appendix \ref{app:embed_usage_hist}. 

Because VQ-TSFM matches or exceeds the imputation accuracy of E1 and E2 whilst being comparatively simpler, it constitutes our chosen architecture for all downstream tasks, with explicit variants used only to validate these findings and to verify that no implicit bias harms performance. However, our ablation study in \autoref{app:cpd_details} shows eventual improvements with E1 and E2, but they are not consistent in all cases.

\section{Change-Point Detection} \label{sec:cpd}

CPD involves identifying abrupt shifts in a time series. The objective is to segment sequential data into partitions generated under different underlying conditions, without prior knowledge of when these changes occur \cite{Page1955}. Presenting CPD as a non-supervised downstream task over the internal structure of a FM is a novel relevant problem in the literature and, as we will demonstrate, has important implications on the FM design.

A Bayesian CPD online approach, presented by \cite{Adams2007}, confronts the problem from a probabilistic perspective. This framework assumes that the observed data at sample $t$ are generated by some probability distribution with unknown parameters $\theta_t$. Each assumed partition is independent of the others and defined by unique parameters. At the same time, observations are regarded as samples drawn from those partitions in an independent and identically distributed (i.i.d.) manner. A significant shift in the base parameters of the distribution will be considered a change point. In the following, subscripts refer to a specific element or sequence from temporal variables. For example, the term $\mathbf{z}_t$ refers to the $t$-th element of the corresponding sequence, whereas $\mathbf{z}_{1:t}$ indicates the span from the first observed day until the current date $t$. We introduce the counting variable $r_t\in\mathbb{N}_0$ to denote the \textit{run length} at time $t$, representing the time (in units, e.g., days in our digital phenotyping setting) that elapsed since the last change point. For a given time $t$, the run length can either increase by one if no change is detected or drop to zero otherwise. Hence, our model focuses on inferring the posterior distribution of this variable, given by
\begin{equation} \label{eq:runlength_posterior}
    p\left(r_t | \mathbf{z}_{1:t}\right) = \frac{p\left(r_t,\mathbf{z}_{1:t}\right)}{p\left(\mathbf{z}_{1:t}\right)}.
\end{equation}

This inference can be made in a recursive and online manner, meaning that, given all past observations, the probability that a change occurred is distributed along all previous days. By deriving this run length distribution, we can have a sense of how our signal behaves in time and when a substantial change has occurred. The run length $r_t$ and the observed data $\mathbf{z}_t$ are jointly modeled as
\begin{equation} \label{eq:runlength_joint_marginalized}
    p\left(r_t,\mathbf{z}_{1:t}\right) = \int p\left(r_t,\mathbf{z}_{1:t},\theta_t\right) \,\text{d}\theta_t,
\end{equation}
where the model parameters are marginalized. The joint density within the integral can be factorized by marginalizing over the run length of the previous day, $r_{t-1}$, which we assume has been previously obtained, as follows:

\vspace{-0.5cm}
\begin{align} \label{eq:runlength_joint}
  &p\left(r_t,\mathbf{z}_{1:t},\theta_t\right) = \sum_{r_{t-1}} p\left(r_t, r_{t-1}, \mathbf{z}_{1:t}, \theta_t\right) \\
  &=\sum_{r_{t-1}} \underbrace{p\left(r_t|r_{t-1}\right)}_{\substack{\text{change point} \\ \text{prior}}} \underbrace{p\left(\mathbf{z}_t|\theta_t\right) p\left(\theta_t|r_{t-1},\mathbf{z}_{1:t-1}\right)}_{\text{predictive posterior}} \nonumber \\ & \quad \cdot\, \underbrace{p\left(r_{t-1},\mathbf{z}_{1:t-1}\right)}_{\text{recursive term}} \nonumber.
\end{align}

The prior probability of having a change point at any moment, conditioned on past change-points, is defined by the hazard function $H(\cdot)$ \cite{Ibe2014}, which in our case was set to a constant that depends on some hyperparameter $\lambda$ such that $p\left(r_t | r_{t-1}\right) = H\left(r_{t-1}\right) = 1/\lambda$. The recursive term in \autoref{eq:runlength_joint} is independent of the model parameters and can be computed recursively. Thus, it follows that

\begin{equation} \label{eq:general}
    p\left(r_t,\mathbf{z}_{1:t}\right) = \sum_{r_{t-1}} p\left(r_t|r_{t-1}\right) \Psi_t p\left(r_{t-1},\mathbf{z}_{1:t-1}\right),
\end{equation}

where the term $\Psi_t$ denotes the predictive posterior of the next datum conditioned to past run length and observed data, which is given by

\begin{equation} \label{eq:predictive_posterior}
    \Psi_t = \int p\left(\mathbf{z}_t|\theta_t\right) p\left(\theta_t|r_{t-1},\mathbf{z}_{1:t-1}\right) \,\text{d}\theta_t.
\end{equation}

\begin{figure*}[bt]
    \centering
    \includegraphics[width=\linewidth]{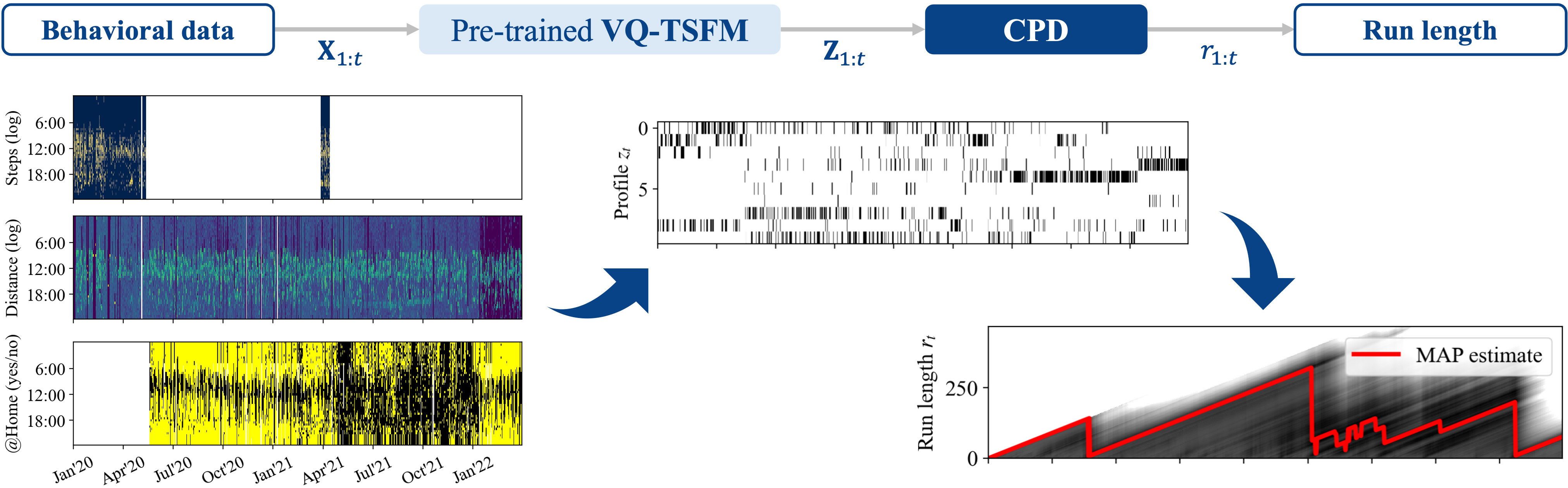}
    \caption{Diagram of the VQ-TSFM–CPD integration with the corresponding variable notation at each step: observed data ($\mathbf{X}_{1:t}$), discrete latent profiles ($\mathbf{Z}_{1:t}$) and run length prediction ($r_{1:t}$). The plots below the diagram illustrate a real-world example: three behavioral sources (step count, distance traveled and time spent at home) are compressed into a latent profile, which is then used to compute the run length, i.e., the time since the last change point. The red line shows the most probable run length for each day (\textit{maximum a posteriori}).}
    \label{fig:vqvae_cpd_integration}
\end{figure*}

The complexity of this term is determined by the choice of prior and likelihood distributions that define the data. In fact, its computation is often intractable, unless the underlying process is modeled after an exponential family with conjugate prior \cite{Turner2013}. However, other strategies can be employed to obtain an approximation of the predictive posterior, such as Markov chain Monte Carlo methods \cite{Moreno2019}. In our case, we exploit the simplicity of the VQ-TSFM patient encoding, as it yields a sequence of categorical observations, to implement a robust CPD with inference in closed-form expression.

Once all probabilities are derived, \autoref{eq:runlength_posterior} returns the run length characterization of the complete temporal sequence: for each day, a distribution explains how the probability of a potential change point is shared among all previous days. After some post-processing, the CPD output is obtained as a binary prediction vector, where 1 indicates a predicted change point and 0 otherwise. Please refer to \autoref{app:cpd_details} for a more in-depth description of the CPD algorithm.

\section{An Unsupervised Downstream Task: Suicide Risk Assessment with CPD} \label{sec:cpd_task}

We now delve into the performance of the Bayesian CPD described above to predict suicidal attempts in advance. Behavioral data from patients with risk of suicidal conduct were collected as described in \autoref{sec:dataset}, whereas clinical records provided the crisis events that the CPD aims to detect. This cohort of suicidal patients —one of the 39 datasets in our collection— was unseen by the VQ-TSFM as it was separated from the others prior to training the model. The heterogeneity, high dimensionality and high missing rate of behavioral data complicates the estimation of underlying parameters and the posterior probability of the run length. To address this problem, a form of profiling step needs to be introduced prior to the CPD stage. We compare three different integrations and their effect on CPD performance:

\begin{itemize}
    \item[(i)] CPD over a patient-specific heterogeneous mixture model (HetMM), where each time sample is independently encoded into a discrete latent posterior distribution and the CPD processes the sequence of such distributions.
    \item[(ii)] CPD coupled to the proposed VQ-TSFM discrete internal structure.
    \item[(iii)] CPD over the continuous embeddings provided by I-TSFM explained.
\end{itemize}

Regarding (i), note that it lacks scalability and efficiency: each individual is represented by a separate model, increasing computational needs and hindering the ability to identify shared patterns across a population. In (ii) and (iii), we use a single model to project every time-series in the internal structure. By training a single model on the whole population, it is able to capture a richer perspective of human behavior across many datasets, without requiring any fine-tuning.

\begin{figure*}[t]
    \centering
    \begin{subfigure}{0.2625\linewidth}
        \includegraphics[width=\linewidth]{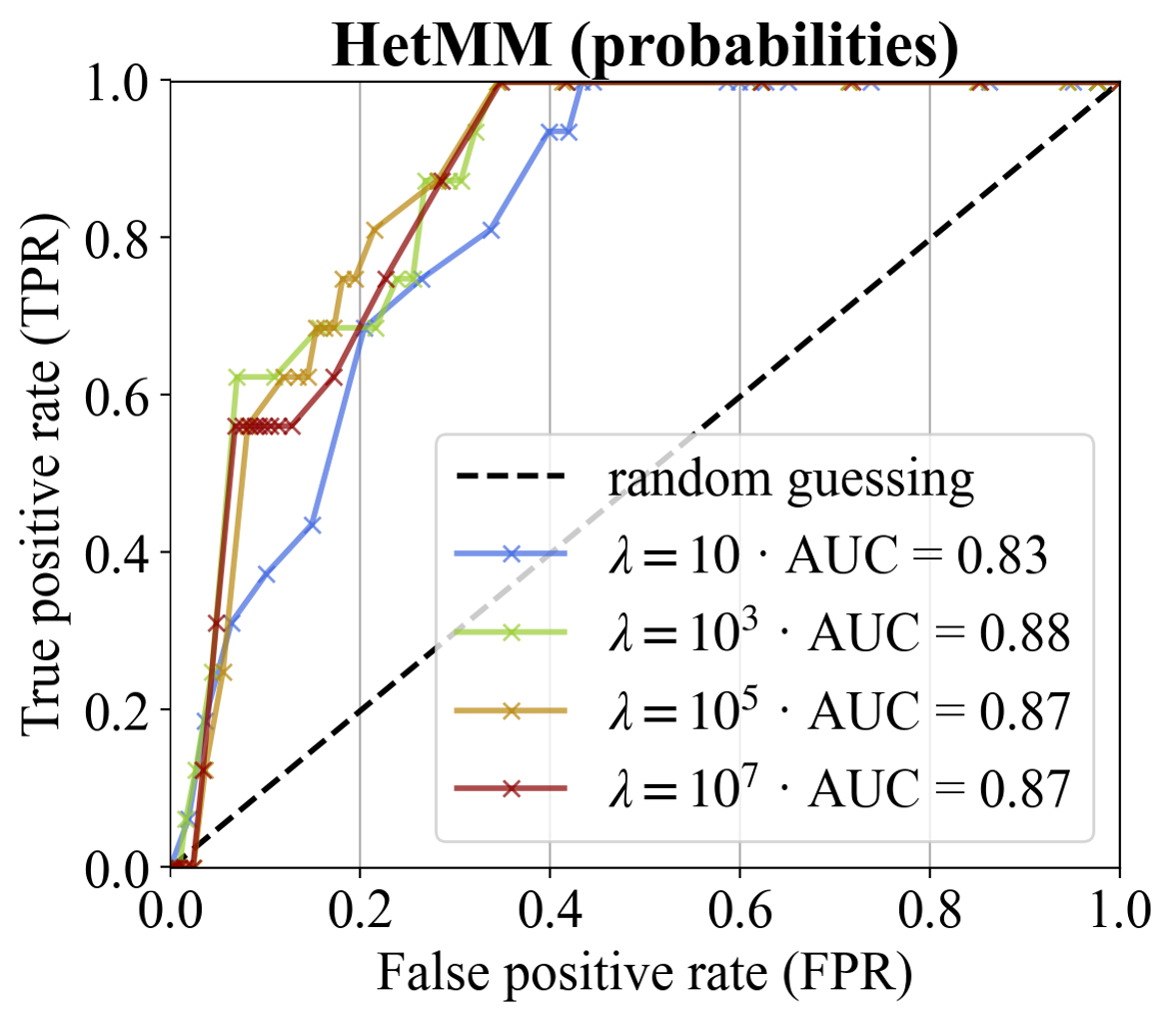}
        \caption{}\label{fig:cpd_hetmm}
    \end{subfigure}
    \hfill
    \begin{subfigure}{0.239\linewidth}
        \includegraphics[width=\linewidth]{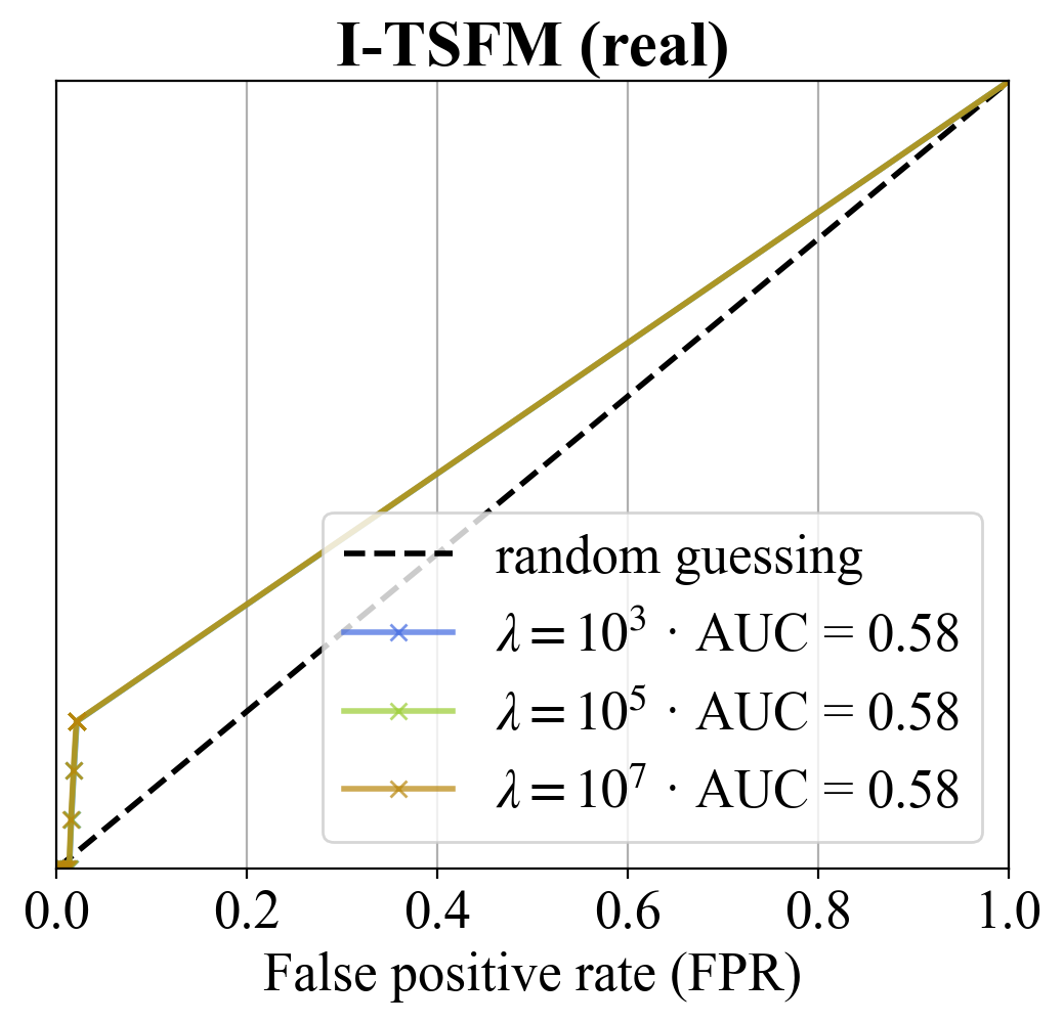}
        \caption{}\label{fig:cpd_informer}
    \end{subfigure}
    \begin{subfigure}{0.239\linewidth}
        \includegraphics[width=\linewidth]{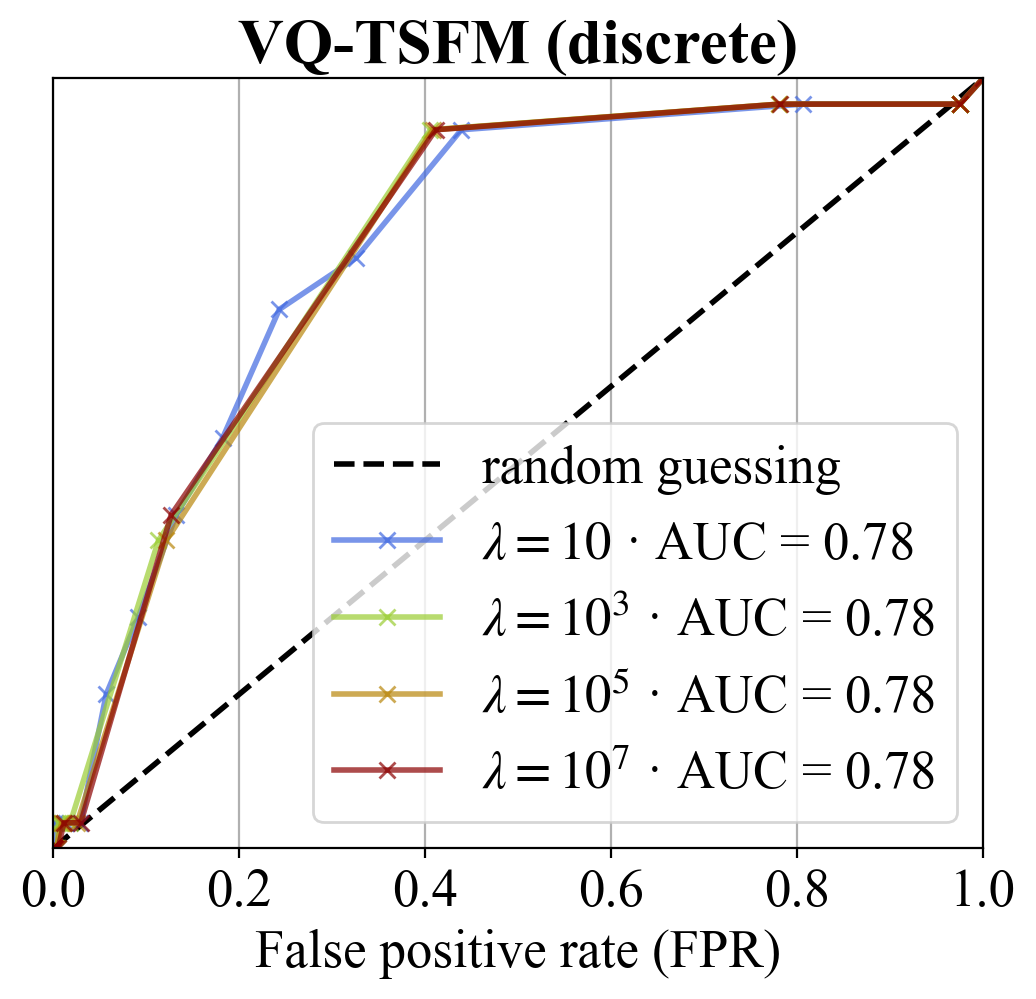}
        \caption{}\label{fig:cpd_vqvae_disc}
    \end{subfigure}
    \begin{subfigure}{0.239\linewidth}
        \includegraphics[width=\linewidth]{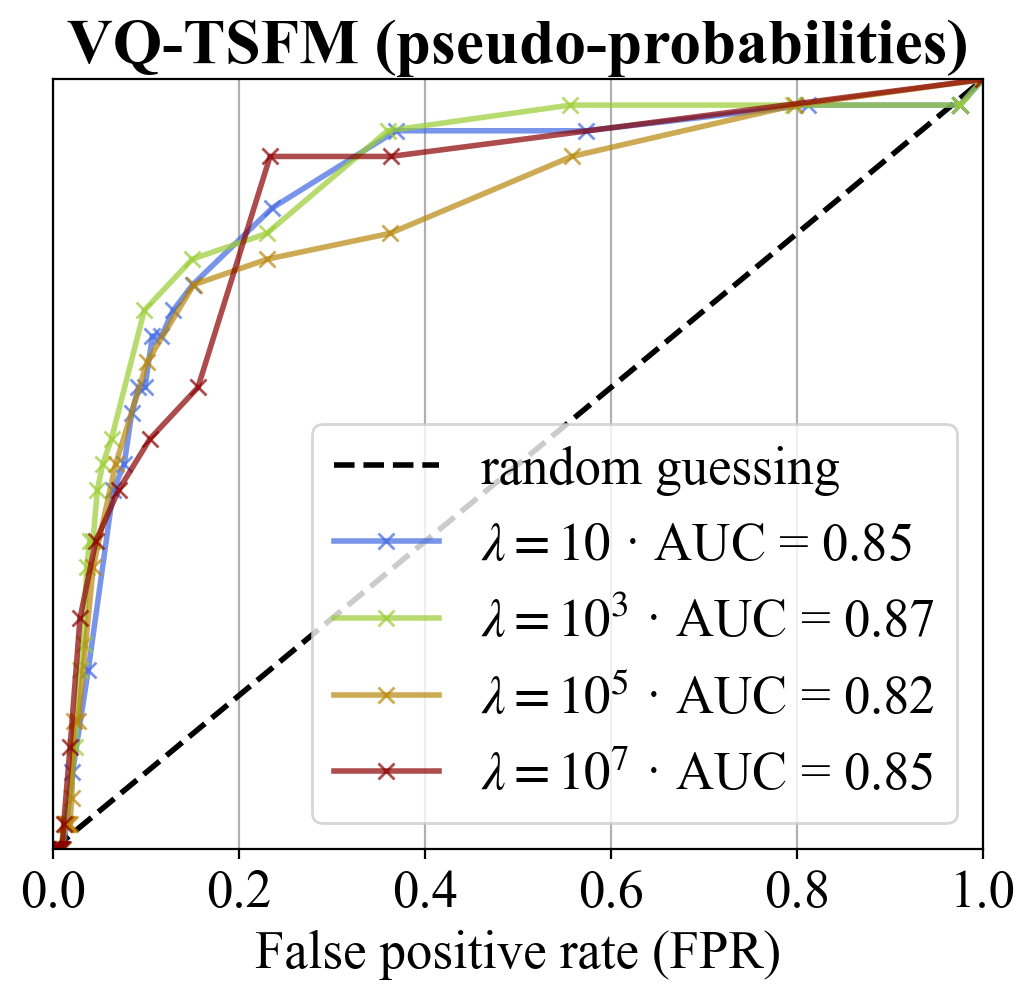}
        \caption{}\label{fig:cpd_vqvae_prob}
    \end{subfigure}
    \caption{ROC curves comparing the performance of the CPD with four different versions of the prior profiling stage: (a) CPD over a patient-specific HetMM, (b) CPD over I-TSFM, (c) CPD over VQ-TSFM, and (d) CPD over VQ-TSFM using pseudo-probabilities. The four colored lines in each plot correspond to four different values of hazard hyperparameter $\lambda$.}
    \label{fig:cpd_results}
\end{figure*}

For each method, we computed the cumulative run-length over a window of seven days, defining an \enquote{instability} estimator. An alarm is returned if the instability rises above some threshold that can be modified to control the CPD sensibility. Alarms were then validated against real events. This threshold was swept to produce a receiver-operating characteristic (ROC) curve, which we used to assess the model trade-off between sensitivity (ability to correctly identify crisis events) and specificity (ability to not raise false alarms, i.e., not returning a positive when there are no events). These metrics, together with the commonly used area under the curve (AUC), were used to compare the different model outputs, which are shown in \autoref{fig:cpd_results}.

The CPD implementation accepts either discrete (integer labels for daily profiles), probabilistic (profile probabilities for each day) or real-valued sequences. While HetMM naturally returns probabilistic profiles, VQ-TSFM provides discrete profiles, which can increase noise when the confidence is low (i.e., the profile distribution is flat) \cite{Romero2022}. Hence, the encoder output was modified to also provide a pseudo-probabilistic interpretation of the latent embeddings (details in \autoref{app:vqvae_latent_cpd}). On the other hand, the Informer architecture returns real-valued embeddings of 64 dimensions. While a multivariate version of the CPD can handle real data, the high-dimensionality of the input leads to a collapse of the run length: the CPD needs to track patterns across several dimensions, and the resulting predictive distributions of when the last change point occurred are extremely weak. To address this problem, a prior step was introduced to reduce the embedding size to 3 with principal component analysis and, while the run length no longer collapsed completely, it still was not certain enough to accurately predict events. The experiment was run for different values of hyperparameter $\lambda$, involved in the so-called hazard function that defines the prior probability of having a change point at any given time instant.

The reference mixture model (\autoref{fig:cpd_hetmm}) maintained AUC scores between 0.83–0.88 for every value of $\lambda$. Remarkably, the VQ-TSFM method achieved comparable results using the discrete profiles. Some of the tested models display false positive rates below 0.25 (i.e., less than 25\% of false alarms) while still maintaining their sensitivity near 90\%. The VQ-TSFM model with the best AUC score was the one using pseudo-probabilities for the patient profiling with $\lambda = 10^3$, achieving an AUC score of 0.87. We emphasize the significance of this result, as the VQ-TSFM approach uses a single model to extract patient profiles that are then used as inputs for the CPD algorithm, establishing a novel and scalable approach for the detection of suicidal events.

\begin{figure*}[t]
    \centering
    \includegraphics[width=0.99\textwidth]{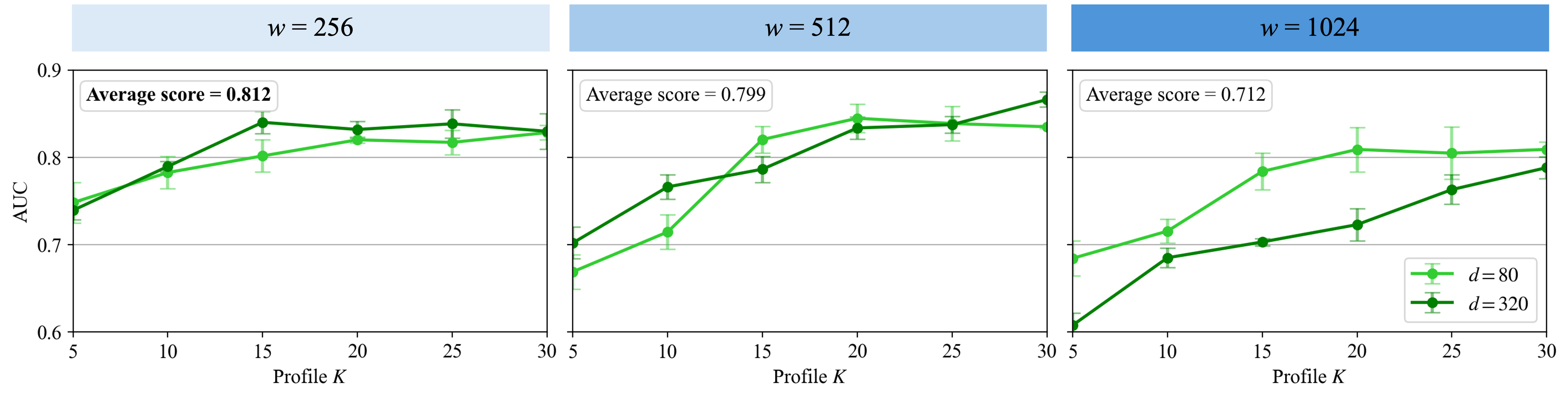}
    \caption{AUC scores of suicide prediction obtained with different VQ-TSFM embedding configurations (using the pseudo-probability output). Each subplot compares the embedding size $d$ and the number of profiles $K$, whereas the three columns display the results for different dictionary size ($w$). A point in the plot represents, for the corresponding model, the average AUC score from the ROC curves using $\lambda=\{10,10^3,10^5,10^7\}$. All points in one subplot are averaged to compute the \textit{Average score} in the top left-hand corner.}
    \label{fig:ablation_vqvae}
\end{figure*}

\autoref{fig:ablation_vqvae} displays additional averaged results for the different VQ-TSFM embedding configurations. Critically, these results demonstrate that the CPD AUC over the VQ-TSFM degrades when increasing the dictionary length $w$, especially for $w=1024$ where this score drops below 0.8. In this regard, we conclude that increasing the discrete resolution of the VQ-TSFM encoder makes it harder to find statistically relevant evidence for behavioral changes. In the continuous limit, this conclusion is supported by the poor performance of the CPD in the I-TSFM case.

The ablation study in \autoref{fig:ablation_vqvae} is further elaborated in \autoref{app:cpd_details}, where an extended figure incorporates results for the VQ-TSFM explicit variations, E1 and E2, capable of integrating the missing mask. The discussion offered in that appendix concludes that VQ-TSFM consistently exhibits robust results across different embedding configurations, and therefore the use of the extended versions E1 and E2 is not justified.

\section{A Supervised Downstream Task: Emotion Prediction} \label{sec:emotion}

Monitoring the emotional state of psychiatric patients is challenging due to discontinuous assessments, environmental influences, and subjective evaluation tools. Given the variability of mental states, modeling emotions through behavioral data enables real-time objective tracking, aiding in risk prevention and treatment \cite{arbaizar2025}. The methodology described in \autoref{sec:dataset} also collects reports made by the patients regarding their emotional state. Following Russell’s 2D model, which defines emotions based on valence (positive to negative) and arousal (high to low), each reported emotion is assigned a valence score: negative (0), neutral (1), or positive (2), providing an easy target on emotion prediction. The presence of such variables in our collection of datasets, which users must enter actively, is very scarce though (96.34\% of missing entries in daily summaries).

\begin{table}[t]
    \centering \small
    \caption{Average AUC scores in suicide event detection, for different configurations of the VQ-TSFM embeddings}
    \label{tab:cpd_trade_off}
    \begin{tabular}{ccccc}
    \toprule
    Embedding dimension ($d$) & \multicolumn{3}{c}{Dictionary length ($w$)} \\
    \midrule
    & 256 & 512 & 1024 \\
    \cmidrule(lr){2-4}
    80  & 0.820 & \textbf{0.845} & 0.809 \\
    320 & 0.832 & 0.834 & 0.723 \\
    \bottomrule
    \multicolumn{4}{p{175pt}}{Scores are the average AUC for $\lambda$ values of $10$, $10^3$, $10^5$ and $10^7$. The number of profiles of the VQ-TSFM was $K=20$.}\\
    \end{tabular}
\end{table}

We now compare both TSFM approaches on this supervised task. The joint population from all clinical programs was split into two partitions, using one of them to train the models and the other one to test their performance. The experiment consisted of using an internal representation of the passive data (the encoded vectors from either the I-TSFM or the VQ-TSFM) from 7-day sequences as input to a classifier which predicted the emotion on the eighth day. These predictions were then contrasted with the actual emotions reported by the patient on the same day. The I-TSFM employed in this study was specifically developed for emotion forecasting, and its capabilities have been validated and presented in \cite{arbaizar2025}. In that prior work, the transformer architecture was integrated with an XGBoost classifier after studying multiple options. This I-TSFM and XGBoost pair serves as reference for this study.

\begin{figure}[b!]
    \centering
    \includegraphics[width=\linewidth]{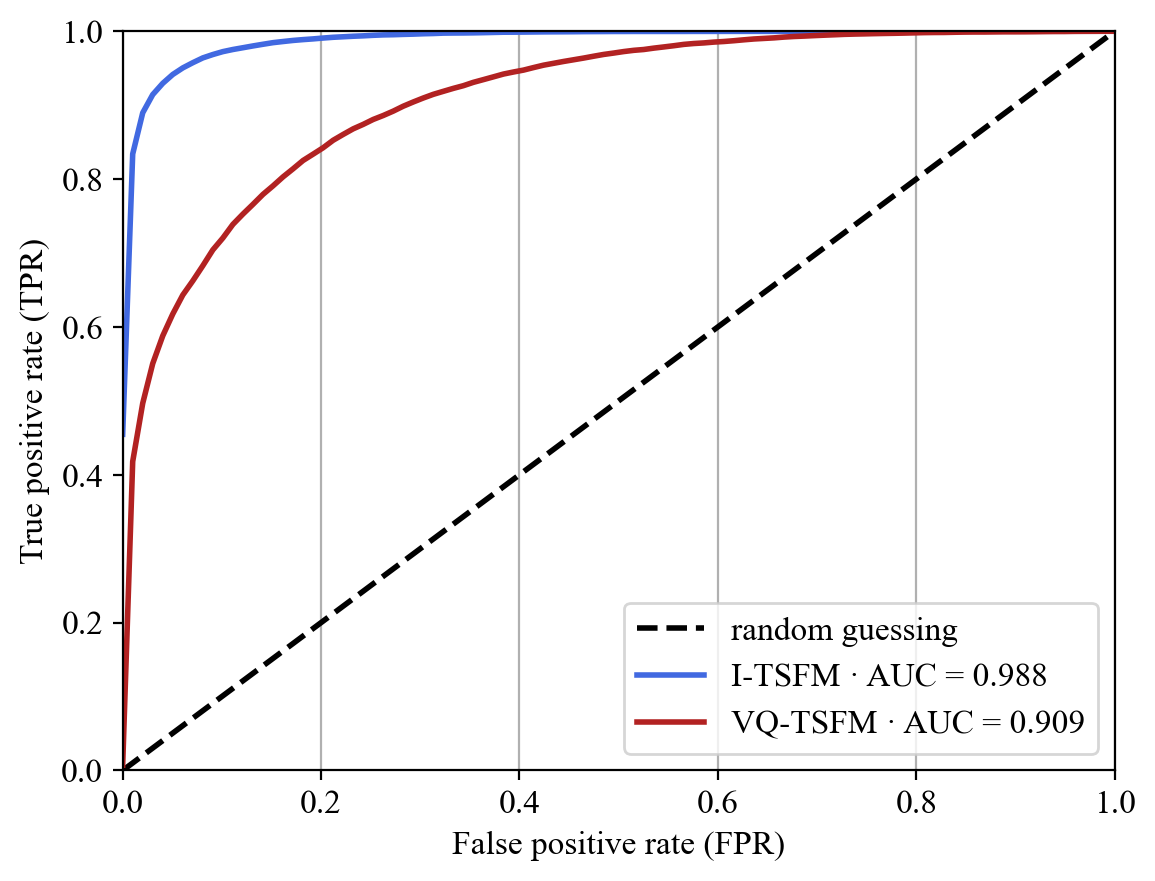}
    \caption{Results of emotion prediction based on 7 days of passive behavior data, processed into a latent space. Two methods are compared: the Informer TSFM coupled to an XGBoost classifier and the VQ-TSFM combined with a 1D CNN that behaves as an integrated downstream task.}
    \label{fig:emotion_results}
\end{figure}

Regarding VQ-TSFM, the dictionary embeddings $\mathbf{e}_{j}$ were fed to the classifying stage, which after some experimentation was adopted to be a one-dimensional convolutional neural network (1D CNN). We acknowledge that optimal downstream processing may vary across different foundation model architectures, and so we conducted a comparative analysis that maximizes the inherent strengths of each TSFM by pairing it with its most effective classification counterpart (XGBoost for the I-TSFM and 1D CNN with VQ-TSFM). Details on the 1D CNN are provided in \autoref{app:emotion_details}.

The validated performance on the test set is compared against the reference XGBoost model using the continuous I-TSFM method in \autoref{fig:emotion_results}. Unlike the unsupervised CPD task, the supervised emotion prediction task benefits substantially from the high-resolution representation capabilities of the I-TSFM, achieving a near-perfect AUC of $0.988$. As expected, introducing quantization through the VQ-TSFM leads to a drop in performance, with the best tested configuration reaching an AUC of $0.909$. Importantly, despite this drop, the results in \autoref{tab:emotion_trade_off} reveal a clear trend: increasing the embedding dimension consistently improves performance, and using a sufficiently large dictionary size (e.g., $w = 512$ or $1024$) also enhances the results. This suggests that the representational capacity of the VQ-TSFM plays a key role in downstream task performance. Interestingly, this trend contrasts with what we observed in the unsupervised CPD task, where increasing the dictionary size led to degraded performance. These findings underline the adaptability of the VQ-TSFM embeddings across tasks, despite the foundation model being trained in a task-agnostic manner.

\begin{table}[t]
    \centering \small
    \caption{Weighted AUC scores in emotion prediction, for different configurations of the VQ-TSFM embeddings}
    \label{tab:emotion_trade_off}
    \begin{tabular}{ccccc}
    \toprule
    Embedding dimension ($d$) & \multicolumn{3}{c}{Dictionary length ($w$)} \\
    \midrule
    & 256 & 512 & 1024 \\
    \cmidrule(lr){2-4}
    80  & 0.827 & 0.895 & 0.901 \\
    320 & 0.895 & \textbf{0.909} & 0.902 \\
    \bottomrule
    \multicolumn{4}{p{175pt}}{AUC scores have been weighted to account for class imbalance. The number of profiles of the VQ-TSFM was $K=20$.}
    \end{tabular}
\end{table}

\section{Discussion on Trade-Off and Conclusion} \label{sec:discussion}

This paper presents a significant advancement in applying foundation models to the analysis of heterogeneous, multisource time-series data collected from wearable devices in healthcare. By leveraging the modified VQ-VAE architecture, our model addresses key challenges such as high rates of missing data and the complex nature of multisource inputs. The model's capacity to reconstruct missing entries and capture critical behavioral patterns through discrete latent representations enhances interpretability, positioning it as a powerful tool for healthcare applications. Our results demonstrate that the model, even without patient-specific fine-tuning, performs remarkably well in tasks such as change-point detection, accurately identifying critical events like suicide attempts. This highlights its potential in monitoring patient behavior and supporting early interventions in healthcare.

While VQ-TSFM excels in unsupervised anomaly detection tasks, our comparison against the Informer-based TSFM (I-TSFM) reveals that continuous embeddings offer distinct advantages in predictive tasks. Specifically, I-TSFM demonstrated superior performance in supervised emotion classification, suggesting that continuous representations provide finer granularity for modeling subtle behavioral patterns over time. In contrast, VQ-TSFM required an increased resolution—expanding the discrete alphabet and embedding dimensions—to approach I-TSFM’s predictive accuracy. However, this enhancement led to a trade-off, as the increased resolution weakened the CPD’s ability to detect statistical changes, illustrating the fundamental tension between optimizing for supervised and unsupervised tasks.

This trade-off underscores the need for future FMs to integrate both discrete and continuous representations, enabling them to effectively balance predictive accuracy with statistical anomaly detection capabilities. A promising avenue for research lies in the development of hybrid architectures that dynamically adapt their latent space based on task-specific requirements. Such models could leverage discrete representations for robust anomaly detection while employing continuous embeddings for fine-grained prediction tasks. Additionally, investigating mechanisms for adaptive resolution tuning within a single FM framework could further enhance flexibility and performance across diverse applications in healthcare.

\section{Ethical Considerations} \label{sec:ethical_considerations}

Each clinical study received approval from the relevant institutional review board in compliance with ethical standards and the Declaration of Helsinki. Institutional review board approval numbers are indicated in brackets and correspond to the center where approval was obtained for each project and country. Patients at high risk of suicide were identified through collaborations with the Jim\'enez D\'iaz Foundation (FJD, EC005-21), Montpellier University Hospital (CPP Ouest IV 20/18\_2), and Cl\'inica Nuestra Se\~nora de la Paz. Patients with common mental disorders were recruited from FJD (PIC148-22), while those with eating disorders were monitored at specialized mental health centers, including Adalmed and Ita mental health clinics. The study also includes patients with cancer monitored in partnership with Gregorio Mara\~n\'on Hospital (EB2COLON2023), Centro Nacional de Investigaciones Oncol\'ogicas, and Fuenlabrada Hospital; patients with HIV/AIDS from Gregorio Mara\~n\'on (MICRO.HGUGM.2022-002); patients with heart problems from Cl\'inico San Carlos Hospital (19/239-O\_P); and patients with obstructive sleep apnea monitored at FJD (PIC163-22). Informed consent was obtained from every participant at the time of inclusion, ensuring adherence to ethical guidelines and participant rights.

\section{Acknowledgments}
This work was funded by Comunidad de Madrid IND2022/TIC-23550, IDEA-CM project (TEC-2024/COM-89) and the ELLIS Unit Madrid (European Laboratory for Learning and Intelligent Systems). Alejandro Lancho was supported by the Comunidad de Madrid through the C\'esar Nombela 2023 program (2023-T1/COM-29065), as well as from the Spanish Ministry of Science, Innovation and Universities (GENICOM project, PID2023-148856OA-I00). Antonio Art\'es-Rodr\'iguez was supported by Instituto de Salud Carlos III and the European Union (NextGenerationEU/PRTR) through the projects PMP22/00032 and PMP24/00026, by MCIN/AEI under grants CPP2023-010721, and by LaCaixa Foundation through the project SMARTCRISIS. Pablo M. Olmos was supported by the 2024 Leonardo Grant for Scientific Research and Cultural Creation from the BBVA Foundation, and by projects MICIU/AEI/10.13039/501100011033/FEDER and UE (PID2021-123182OB-I00; EPiCENTER).

{\small
\bibliographystyle{IEEEtran}
\bibliography{main}}

\newpage \appendix \onecolumn


\makeatletter
    \vbox to 1.0in
    \bgroup
        \hrule height4pt
        \vskip .10in
        \centering

        \Huge\bfseries
        Supplementary Material

        \vskip .15in
        \hrule height1pt
        \vskip .10in
        \vskip 0.1in plus 1fil
    \egroup
\makeatother


\section{Data Preprocessing for the VQ-TSFM} \label{app:vqvae_preprocessing}

As outlined in \autoref{sec:dataset}, the original dataset comprises 64 variables, many of which exhibit high levels of missing data. This poses a significant challenge for standard deep learning techniques, which typically require large datasets to generalize effectively. Thus, an extensive data processing pipeline was necessary and is described in detail here.

In order to rigorously assess the performance of the three proposed models (VQ-TSFM and its extensions E1 and E2), we implemented a robust evaluation strategy based on an $n$-partition scheme of the original dataset. Each partition was systematically allocated for training, validation, and testing---along with reconstructed signal plots---across all models. Importantly, this design ensured that the data partitions were consistent across all models, precluding any leakage of patient data between partitions within a given $n$-partition configuration. This strict partitioning protocol enabled a fair comparison between the mask-conditioned architectures (E1, E2), and the non-conditioned model, ensuring identical experimental conditions across different, randomly sampled sections of the dataset.

A key challenge in modeling time-series data is the transformation of the tabular dataset into a format suitable for deep learning techniques. Specifically, we reshaped the data into observation batches with dimensions $[B, F, L]$, where $B$ denotes the batch size, $F$ the number of features, and $L$ the sequence length. The initial preprocessing step involved the removal of uninformative or redundant variables, coupled with a stringent constraint ensuring that patient records were not split across training, validation, and test within any $n$-partition. Instead, all data from a single patient were placed within the same partition to preserve temporal and contextual consistency.

Several variables were excluded from the analysis due to inconsistencies in missing data reporting. For instance, features such as the variables measuring the minimum/maximum/average heart rate used a placeholder value of $-1$ to indicate missing data, whereas other variables adhered to the standard Numpy convention of using NaN. Date-related variables also required normalization to a consistent format. Additionally, certain variables contained erroneous or outlier values, likely due to faulty sensors or other external factors, as discussed in \autoref{sec:dataset}. While it was not possible to completely eliminate all erroneous entries due to the absence of key contextual variables, we removed the majority of manifestly inaccurate data points. For example, the \textit{Sleep Duration} variable is known to be device-dependent, with different vendors applying varying algorithms to detect sleep patterns. Similarly, the \textit{Total Steps} variable can be influenced by non-step movements, such as hand gestures, whereas the \textit{App Usage} variable is constrained by vendor-specific limitations. The \textit{Location Clusters} variable, being derived from external algorithms that process raw geolocation data, also exhibited potential inaccuracies.

To mitigate these issues and improve model stability, we applied the constraints shown in \autoref{tab:clipinfo}, where the columns \enquote{Minimum Bound} and \enquote{Maximum Bound} specify the ranges to clip the values in \enquote{Original Minimum} and \enquote{Original Maximum}. Any value outside these bounds was marked as missing.

\begin{table}[ht]
    \centering
    \caption{Clipping constraints applied to ensure model stability}
    \label{tab:clipinfo}
    \begin{tabular}{@{}lcccc@{}}
    \toprule
    \textbf{Variable} & \textbf{Original Minimum} & \textbf{Original Maximum} & \textbf{Minimum Bound} & \textbf{Maximum Bound} \\
    \midrule
    Sleep Start (s) & -11,657,590 & 7,430,400 & -22,500 & 25,000 \\
    Distance (m) & 7.891e-10 & 9,945,435.20 & 20 & 95,000 \\
    Time at Home (m)& 0.0 & 1,440 & 120 & --- \\
    Sleep Duration (s) & 1.0 & 86,400.0 & 3,600 & 54,000 \\
    Time Walking (s) & 0.0 & 3,098,824.0 & 120 & 15,000 \\
    App Usage (s) & 0.0 & 630,478.0 & 180 & 35,000 \\
    Location Clusters & 0 & 40 & 1 & 15 \\
    Total Steps & 1 & 99,734 & 150 & 25,000 \\
    \bottomrule
    
    \multicolumn{5}{@{}p{0.9\linewidth}@{}}
    {The \textit{Original Minimum} and \textit{Original Maximum} columns represent the range of raw variable values in the dataset, whereas the \textit{Minimum Bound} and \textit{Maximum Bound} columns define the clipping thresholds. Values falling outside these bounds were treated as missing to avoid outliers, erroneous data, and ensure more reliable model training.}
    \end{tabular}
\end{table}

After the initial preprocessing steps, we ensured that each patient's time-series data remained temporally contiguous. Specifically, if a patient's records spanned from March 15, 2019, to May 2, 2019, but included a gap until May 15, 2019, the data were split into two distinct sequences: one from March 15 to May 2, and the other from May 15 to the end of the recording period (e.g., June 24). Sequences that were shorter than the predefined minimum length, were discarded to maintain consistency in sequence length across the dataset. This was not applied to the final subset of held-out psychiatric patients whose time-series---varying in length--- were processed in full.

Next, we addressed differences in scale across continuous and counting variables by applying appropriate transformations. For real-valued continuous features, we utilized scikit-learn's \texttt{RobustScaler}, which is well-suited for handling data with outliers by centering the data around the median and scaling it based on the interquantile range (IQR). These transformations were fitted on the training set and subsequently applied to the validation and test sets to ensure consistency across all partitions.

It is important to note that all metrics and signal reconstructions reported in this work reflect the original feature space. To achieve this, we reversed the scaling transformations prior to computing evaluation metrics and generating signal plots. This approach ensures that the reported results are both interpretable and faithful to the original data distributions.

For each model instance, a missingness mask was dynamically generated for each patient sequence, with synthetic missingness introduced to simulate unobserved data. This missingness mask consisted of three distinct values: \enquote{0} for originally missing data, \enquote{1} for observed data, and \enquote{2} for synthetically induced missing data. However, for model input, the mask was binarized by collapsing \enquote{2} into \enquote{0}, as the model was designed to treat all missing entries uniformly, regardless of whether the missingness was natural or synthetically generated.

To simulate missing data, we employ two distinct strategies: MCAR (missing completely at random) and MNAR (missing not at random). Each mode is constructed to introduce missingness in ways that reflect both random and structure data loss. In the MCAR setting, missingness is introduced through a random process designed to target approximately 10\% of the observed entries. However, a series of safeguard conditions modulate this target to ensure data integrity. Specifically:

\begin{itemize}
    \item If more than 85\% of the data for any feature is already missing, no additional missingness is introduced.
    \item A flat rate of 10\% is tentatively introduced if there is not prior existing missingness for a given sample.
    \item For each feature, missing values are added by randomly selecting from the observed entries, ensuring that only those entries are affected.
\end{itemize}

The result is a systematic, yet random, distribution of missingness that prevents over-saturation while maintaining stochasticity.

In contrast, MNAR employs a feature-drive approach, introducing missingness based on relationships between variables and their values. Structured missingness is inserted through a combination of non-linear conditions and thresholds. The MNAR process unfolds as follows:

\begin{itemize}
    \item If more than 85\% of the data for any feature is already missing, no additional missingness is introduced.
    \item Non-linear conditions are applied to enforce missingness. For example, if a feature consistently deviates from its typical range (e.g., extreme values of a continuous variable), missingness is introduced.
\end{itemize}

To avoid excessive data sparsity, the same 85\% ceiling on missingness per feature is applied, ensuring that no single features becomes overwhelmingly absent. Furthermore, a small percentage of random missingness (approximately 2\%) is introduced to account for incidental data loss not captured by the MNAR corruption process.

Finally, a wrapper class for resolution augmentation was developed but was not used in the final experiments. This method was found to exacerbate existing missingness streaks, complicating model training. To handle varying sequence lengths, random cropping was applied to select sub-sequences for analysis.

\newpage
\section{Informer Architectural Details} \label{app:informer}

The Informer approach described in \autoref{sec:fm_informer} follows an encoder-decoder transformer architecture. \autoref{tab:transformer_architecture} shows the details of each layer of such model. The input for the encoder and decoder were reshaped into batches of dimensions [$B$, $L_e$, $S$] and [$B$, $L_d$, $S$], respectively, where $B$ denotes the batch size, $L_e$ and $L_d$ the encoder and decoder sequence lengths, and $S$ is the feature dimension.

\begin{table*}[ht]
  \centering
  \setlength{\tabcolsep}{3pt}
  \renewcommand{\arraystretch}{0.85}
  \begin{threeparttable}
    \caption{Transformer Model Architecture}
    \label{tab:transformer_architecture}
    \begin{tabular}{@{}c c c l@{}}
      \toprule
      \multicolumn{4}{c}{\textbf{Encoder}} \\ 
      \midrule
      \textbf{Layer Type} 
        & \textbf{Input Dimensions} 
        & \textbf{Output Dimensions} 
        & \textbf{Details} \\ 
      \midrule

      Input (Signal) 
        & $[B, L_e, S]$ 
        & --- 
        & Encoder input (sequence of features) \\

      Embedding Transformation 
        & $[B, L_e, S]$ 
        & $[B, L_e, d_{\text{embedding}}]$ 
        & $\mathbf{X}_{\text{enc}}\in\mathbb{R}^{L_e\times S}\to\mathbb{R}^{L_e\times d_{\text{embedding}}}$ \\

      Positional Encoding Addition 
        & $[B, L_e, d_{\text{embedding}}]$ 
        & $[B, L_e, d_{\text{embedding}}]$ 
        & Add sinusoidal positional encoding \\

      \addlinespace[0.5ex]
      \cdashline{1-4}
      \addlinespace[0.5ex]
      \multicolumn{4}{c}{3 $\times$ Attention Blocks} \\
      \addlinespace[0.5ex]
      \cdashline{1-4}
      \addlinespace[0.5ex]

      ProbSparse Self-Attention 
        & $[B, L_e, d_{\text{embedding}}]$ 
        & $[B, L_e, d_{\text{embedding}}]$ 
        & Efficient self-attention (O($L\log L$)) \\

      Add and Normalization 
        & $[B, L_e, d_{\text{embedding}}]$ 
        & $[B, L_e, d_{\text{embedding}}]$ 
        & Residual add + LayerNorm \\

      Feed-Forward Layer 
        & $[B, L_e, d_{\text{embedding}}]$ 
        & $[B, L_e, d_{\text{embedding}}]$ 
        & FC + ReLU \\

      Add and Normalization 
        & $[B, L_e, d_{\text{embedding}}]$ 
        & $[B, L_e, d_{\text{embedding}}]$ 
        & Residual add + LayerNorm \\

      Conv1D + ELU Activation 
        & $[B, L_e, d_{\text{embedding}}]$ 
        & $[B, L_e, d_{\text{embedding}}]$ 
        & 1D conv (kernel=3, stride=1) + ELU \\

      \midrule
      \multicolumn{4}{c}{\textbf{Decoder}} \\ 
      \midrule

      Input (Signal) 
        & $[B, L_d, S]$ 
        & --- 
        & Decoder input (sequence of features) \\

      Embedding Transformation 
        & $[B, L_d, S]$ 
        & $[B, L_d, d_{\text{embedding}}]$ 
        & $\mathbf{X}_{\text{dec}}\in\mathbb{R}^{L_d\times S}\to\mathbb{R}^{L_d\times d_{\text{embedding}}}$ \\

      Positional Encoding Addition 
        & $[B, L_d, d_{\text{embedding}}]$ 
        & $[B, L_d, d_{\text{embedding}}]$ 
        & Add sinusoidal positional encoding \\

      \addlinespace[0.5ex]
      \cdashline{1-4}
      \addlinespace[0.5ex]
      \multicolumn{4}{c}{3 $\times$ Attention Blocks} \\
      \addlinespace[0.5ex]
      \cdashline{1-4}
      \addlinespace[0.5ex]

      Masked Self-Attention 
        & $[B, L_d, d_{\text{embedding}}]$ 
        & $[B, L_d, d_{\text{embedding}}]$ 
        & Prevents attending to future tokens \\

      Add and Normalization 
        & $[B, L_d, d_{\text{embedding}}]$ 
        & $[B, L_d, d_{\text{embedding}}]$ 
        & Residual add + LayerNorm \\

      Cross-Attention 
        & $[B, L_d, d_{\text{embedding}}]$ 
        & $[B, L_d, d_{\text{embedding}}]$ 
        & Attends to encoder outputs \\

      Add and Normalization 
        & $[B, L_d, d_{\text{embedding}}]$ 
        & $[B, L_d, d_{\text{embedding}}]$ 
        & Residual add + LayerNorm \\

      Feed-Forward Layer 
        & $[B, L_d, d_{\text{embedding}}]$ 
        & $[B, L_d, d_{\text{embedding}}]$ 
        & FC + ReLU \\

      Add and Normalization 
        & $[B, L_d, d_{\text{embedding}}]$ 
        & $[B, L_d, d_{\text{embedding}}]$ 
        & Residual add + LayerNorm \\

      Linear Projection 
        & $[B, L_d, d_{\text{embedding}}]$ 
        & $[B, L_d, S]$ 
        & Project to output space \\

      Softmax 
        & $[B, L_d, S]$ 
        & $[B, L_d, S]$ 
        & Probability distribution \\

      \bottomrule
    \end{tabular}
  \end{threeparttable}
\end{table*}

\newpage
\section{VQ-TSFM Architectural Details} \label{app:vqvae_details}

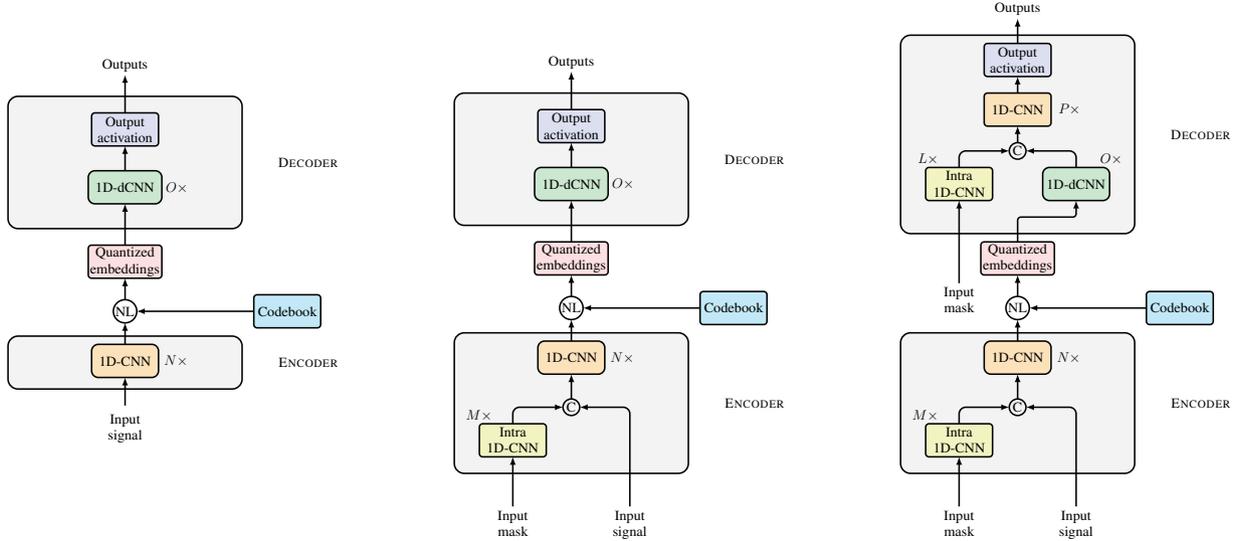
\begin{figure*}[h]
    \centering
    \begin{subfigure}{0.315\textwidth}
        \centering
        \scalebox{0.44}{\usetikzlibrary{arrows.meta}

\begin{tikzpicture}[baseline={(0,0)}, every node/.style={font=\large}]

\definecolor{emb_color}{RGB}{252,224,225}
\definecolor{multi_head_attention_color}{RGB}{252,226,187}
\definecolor{add_norm_color}{RGB}{242,243,193}
\definecolor{ff_color}{RGB}{194,232,247}
\definecolor{softmax_color}{RGB}{203,231,207}
\definecolor{linear_color}{RGB}{220,223,240}
\definecolor{gray_bbox_color}{RGB}{243,243,244}

\draw[fill=gray_bbox_color, line width=0.046875cm, rounded corners=0.300000cm] (1, 6.25) -- (8, 6.25) -- (8, 4.65) -- (1, 4.65) -- cycle;
\node[anchor=east] at (11,5.45) {\textsc{Encoder}};

\node[text width=2.500000cm, anchor=north, align=center] at (4.5,4) {Input\\signal};
\draw[line width=0.046875cm, -latex, rounded corners=0.2cm] (4.5, 4.15) -- (4.5, 5);

\draw[line width=0.046875cm, fill=multi_head_attention_color, rounded corners=0.2cm] (3.5, 6) -- (5.5, 6) -- (5.5, 5) -- (3.5, 5) -- cycle;
\node[align=center] at (4.5,5.5) {1D-CNN};

\node[anchor=east] at (6.5,5.5) {$N\times$};

\draw[line width=0.046875cm, -latex] (4.5, 6) -- (4.5, 6.65);

\node[align=center] at (4.5,7) {NL};
\draw[line width=0.046875cm] (4.5, 7) circle (0.35);

\draw[line width=0.046875cm, fill=ff_color, rounded corners=0.1cm] (8.35, 7.5) -- (10.35, 7.5) -- (10.35, 6.5) -- (8.35, 6.5) -- cycle;
\node[text width=2.500000cm, align=center] at (9.35, 7) {Codebook};
\draw[line width=0.046875cm, -latex] (8.35, 7) -- (4.85, 7);

\draw[line width=0.046875cm, -latex] (4.5, 7.35) -- (4.5, 8);
\draw[line width=0.046875cm, fill=emb_color, rounded corners=0.1cm] (3.4, 9) -- (5.6, 9) -- (5.6, 8) -- (3.4, 8) -- cycle;
\node[text width=2.500000cm, align=center] at (4.5,8.5) {Quantized\\embeddings};

\draw[fill=gray_bbox_color, line width=0.046875cm, rounded corners=0.300000cm] (1, 13.5) -- (8, 13.5) -- (8, 9.5) -- (1, 9.5) -- cycle;
\draw[line width=0.046875cm, -latex, rounded corners=0.1cm] (4.5, 9) -- (4.5, 10.25);
\node[anchor=east] at (11,11.5) {\textsc{Decoder}};

\draw[line width=0.046875cm, fill=softmax_color, rounded corners=0.2cm] (3.4, 11.25) -- (5.6, 11.25) -- (5.6, 10.25) -- (3.4, 10.25) -- cycle;
\node[align=center] at (4.5,10.75) {1D-dCNN};
\node[anchor=east] at (6.5,10.75) {$O\times$};
\draw[line width=0.046875cm, -latex, rounded corners=0.2cm] (4.5, 11.25) -- (4.5, 12);


\draw[line width=0.046875cm, fill=linear_color, rounded corners=0.1cm] (3.5, 13) -- (5.5, 13) -- (5.5, 12) -- (3.5, 12) -- cycle;
\node[align=center] at (4.5, 12.5) {Output\\activation};
\draw[-latex, line width=0.046875cm, rounded corners=0.2cm] (4.5, 13) -- (4.5, 14.15);
\node[text width=2.500000cm, anchor=north, align=center, ] at (4.5, 14.75) {Outputs};

\end{tikzpicture}}
        \caption{Implicit VQ-TSFM (without missingness mask conditioning).}
        \label{fig:model_a0}
    \end{subfigure}%
    \hfill
    \begin{subfigure}{0.315\textwidth}
        \centering
        \scalebox{0.44}{\usetikzlibrary{arrows.meta}

\begin{tikzpicture}[baseline={(0,0)}, every node/.style={font=\large}]

\definecolor{emb_color}{RGB}{252,224,225}
\definecolor{multi_head_attention_color}{RGB}{252,226,187}
\definecolor{add_norm_color}{RGB}{242,243,193}
\definecolor{ff_color}{RGB}{194,232,247}
\definecolor{softmax_color}{RGB}{203,231,207}
\definecolor{linear_color}{RGB}{220,223,240}
\definecolor{gray_bbox_color}{RGB}{243,243,244}

\draw[fill=gray_bbox_color, line width=0.046875cm, rounded corners=0.300000cm] (1, 6.25) -- (8, 6.25) -- (8, 2) -- (1, 2) -- cycle;
\node[anchor=east] at (11, 4.125) {\textsc{Encoder}};

\node[text width=2.500000cm, anchor=north, align=center, ] at (2.75,1) {Input\\mask};
\draw[line width=0.046875cm, -latex] (2.75, 1) -- (2.75, 2.5);

\node[text width=2.500000cm, anchor=north, align=center] at (6.25,1) {Input\\signal};
\draw[line width=0.046875cm, -latex, rounded corners=0.2cm] (6.25, 1) -- (6.25, 4) -- (4.75, 4);

\draw[line width=0.046875cm, fill=add_norm_color, rounded corners=0.1cm] (1.75, 3.5) -- (3.75, 3.5) -- (3.75, 2.5) -- (1.75, 2.5) -- cycle;
\node[align=center] at (2.75, 3) {Intra\\1D-CNN};

\node[anchor=east] at (2.25,3.75) {$M\times$};

\draw[-latex, line width=0.046875cm, rounded corners=0.2cm] (2.75, 3.5) -- (2.75, 4) -- (4.25, 4);
\node[align=center] at (4.5,4) {C};
\draw[line width=0.046875cm] (4.5, 4) circle (0.25);

\draw[-latex, line width=0.046875cm, rounded corners=0.200000cm] (4.5, 4.25) -- (4.5, 5);
\draw[line width=0.046875cm, fill=multi_head_attention_color, rounded corners=0.2cm] (3.5, 6) -- (5.5, 6) -- (5.5, 5) -- (3.5, 5) -- cycle;
\node[align=center] at (4.5,5.5) {1D-CNN};

\node[anchor=east] at (6.5,5.5) {$N\times$};

\draw[line width=0.046875cm, -latex] (4.5, 6) -- (4.5, 6.65);

\node[align=center] at (4.5,7) {NL};
\draw[line width=0.046875cm] (4.5, 7) circle (0.35);

\draw[line width=0.046875cm, fill=ff_color, rounded corners=0.1cm] (8.35, 7.5) -- (10.35, 7.5) -- (10.35, 6.5) -- (8.35, 6.5) -- cycle;
\node[text width=2.500000cm, align=center] at (9.35, 7) {Codebook};
\draw[line width=0.046875cm, -latex] (8.35, 7) -- (4.85, 7);

\draw[line width=0.046875cm, -latex] (4.5, 7.35) -- (4.5, 8);
\draw[line width=0.046875cm, fill=emb_color, rounded corners=0.1cm] (3.4, 9) -- (5.6, 9) -- (5.6, 8) -- (3.4, 8) -- cycle;
\node[text width=2.500000cm, align=center] at (4.5,8.5) {Quantized\\embeddings};

\draw[fill=gray_bbox_color, line width=0.046875cm, rounded corners=0.300000cm] (1, 13.5) -- (8, 13.5) -- (8, 9.5) -- (1, 9.5) -- cycle;
\draw[line width=0.046875cm, -latex, rounded corners=0.1cm] (4.5, 9) -- (4.5, 10.25);
\node[anchor=east] at (11,11.5) {\textsc{Decoder}};

\draw[line width=0.046875cm, fill=softmax_color, rounded corners=0.2cm] (3.4, 11.25) -- (5.6, 11.25) -- (5.6, 10.25) -- (3.4, 10.25) -- cycle;
\node[align=center] at (4.5,10.75) {1D-dCNN};
\node[anchor=east] at (6.5,10.75) {$O\times$};
\draw[line width=0.046875cm, -latex, rounded corners=0.2cm] (4.5, 11.25) -- (4.5, 12);


\draw[line width=0.046875cm, fill=linear_color, rounded corners=0.1cm] (3.5, 13) -- (5.5, 13) -- (5.5, 12) -- (3.5, 12) -- cycle;
\node[align=center] at (4.5, 12.5) {Output\\activation};
\draw[-latex, line width=0.046875cm, rounded corners=0.2cm] (4.5, 13) -- (4.5, 14.15);
\node[text width=2.500000cm, anchor=north, align=center, ] at (4.5, 14.75) {Outputs};

\end{tikzpicture}}
        \caption{VQ-TSFM E1 (encoder-only missingness mask conditioning).}
        \label{fig:model_a1}
    \end{subfigure}%
    \hfill
    \begin{subfigure}{0.315\textwidth}
        \centering
        \scalebox{0.44}{\usetikzlibrary{arrows.meta}

\begin{tikzpicture}[baseline={(0,0)}, every node/.style={font=\large}]

\definecolor{emb_color}{RGB}{252,224,225}
\definecolor{multi_head_attention_color}{RGB}{252,226,187}
\definecolor{add_norm_color}{RGB}{242,243,193}
\definecolor{ff_color}{RGB}{194,232,247}
\definecolor{softmax_color}{RGB}{203,231,207}
\definecolor{linear_color}{RGB}{220,223,240}
\definecolor{gray_bbox_color}{RGB}{243,243,244}

\draw[fill=gray_bbox_color, line width=0.046875cm, rounded corners=0.300000cm] (1, 6.25) -- (8, 6.25) -- (8, 2) -- (1, 2) -- cycle;
\node[anchor=east] at (11, 4.125) {\textsc{Encoder}};

\node[text width=2.500000cm, anchor=north, align=center, ] at (2.75,1) {Input\\mask};
\draw[line width=0.046875cm, -latex] (2.75, 1) -- (2.75, 2.5);

\node[text width=2.500000cm, anchor=north, align=center] at (6.25,1) {Input\\signal};
\draw[line width=0.046875cm, -latex, rounded corners=0.2cm] (6.25, 1) -- (6.25, 4) -- (4.75, 4);

\draw[line width=0.046875cm, fill=add_norm_color, rounded corners=0.1cm] (1.75, 3.5) -- (3.75, 3.5) -- (3.75, 2.5) -- (1.75, 2.5) -- cycle;
\node[align=center] at (2.75, 3) {Intra\\1D-CNN};

\node[anchor=east] at (2.25,3.75) {$M\times$};

\draw[-latex, line width=0.046875cm, rounded corners=0.2cm] (2.75, 3.5) -- (2.75, 4) -- (4.25, 4);
\node[align=center] at (4.5,4) {C};
\draw[line width=0.046875cm] (4.5, 4) circle (0.25);

\draw[-latex, line width=0.046875cm, rounded corners=0.200000cm] (4.5, 4.25) -- (4.5, 5);
\draw[line width=0.046875cm, fill=multi_head_attention_color, rounded corners=0.2cm] (3.5, 6) -- (5.5, 6) -- (5.5, 5) -- (3.5, 5) -- cycle;
\node[align=center] at (4.5,5.5) {1D-CNN};

\node[anchor=east] at (6.5,5.5) {$N\times$};

\draw[line width=0.046875cm, -latex] (4.5, 6) -- (4.5, 6.65);

\node[align=center] at (4.5,7) {NL};
\draw[line width=0.046875cm] (4.5, 7) circle (0.35);

\draw[line width=0.046875cm, fill=ff_color, rounded corners=0.1cm] (8.35, 7.5) -- (10.35, 7.5) -- (10.35, 6.5) -- (8.35, 6.5) -- cycle;
\node[text width=2.500000cm, align=center] at (9.35, 7) {Codebook};
\draw[line width=0.046875cm, -latex] (8.35, 7) -- (4.85, 7);

\draw[line width=0.046875cm, -latex] (4.5, 7.35) -- (4.5, 8);
\draw[line width=0.046875cm, fill=emb_color, rounded corners=0.1cm] (3.4, 9) -- (5.6, 9) -- (5.6, 8) -- (3.4, 8) -- cycle;
\node[text width=2.500000cm, align=center] at (4.5,8.5) {Quantized\\embeddings};

\draw[fill=gray_bbox_color, line width=0.046875cm, rounded corners=0.300000cm] (1, 15.25) -- (8, 15.25) -- (8, 9.25) -- (1, 9.25) -- cycle;
\draw[line width=0.046875cm, -latex, rounded corners=0.1cm] (4.5, 9) -- (4.5, 9.75) -- (6.25, 9.75) -- (6.25, 10.25);
\node[anchor=east] at (11,12.25) {\textsc{Decoder}};

\draw[line width=0.046875cm, fill=softmax_color, rounded corners=0.2cm] (5.25, 11.25) -- (7.25, 11.25) -- (7.25, 10.25) -- (5.25, 10.25) -- cycle;
\node[align=center] at (6.25,10.75) {1D-dCNN};
\node[anchor=east] at (7.75,11.5) {$O\times$};

\node[text width=2.500000cm, anchor=north, align=center] at (2.75,7.75) {Input\\mask};
\draw[line width=0.046875cm, -latex] (2.75, 7.75) -- (2.75, 10.25);

\draw[line width=0.046875cm, fill=add_norm_color, rounded corners=0.1cm] (1.75, 11.25) -- (3.75, 11.25) -- (3.75, 10.25) -- (1.75, 10.25) -- cycle;
\node[align=center] at (2.75, 10.75) {Intra\\1D-CNN};
\draw[line width=0.046875cm, -latex, rounded corners=0.2cm] (2.75, 11.25) -- (2.75, 11.75) -- (4.25, 11.75);
\draw[line width=0.046875cm, -latex, rounded corners=0.2cm] (6.25, 11.25) -- (6.25, 11.75) -- (4.75, 11.75);

\node[anchor=east] at (2.25,11.5) {$L\times$};

\node[align=center] at (4.5,11.75) {C};
\draw[line width=0.046875cm] (4.5, 11.75) circle (0.25);
\draw[-latex, line width=0.046875cm, rounded corners=0.2cm] (4.5, 12) -- (4.5, 12.5);

\draw[line width=0.046875cm, fill=multi_head_attention_color, rounded corners=0.1cm] (3.5, 13.5) -- (5.5, 13.5) -- (5.5, 12.5) -- (3.5, 12.5) -- cycle;
\node[align=center] at (4.5, 13) {1D-CNN};
\node[anchor=east] at (6.5,13) {$P\times$};
\draw[-latex, line width=0.046875cm, rounded corners=0.2cm] (4.5, 13.5) -- (4.5, 14);


\draw[line width=0.046875cm, fill=linear_color, rounded corners=0.1cm] (3.5, 15) -- (5.5, 15) -- (5.5, 14) -- (3.5, 14) -- cycle;
\node[align=center] at (4.5, 14.5) {Output\\activation};
\draw[-latex, line width=0.046875cm, rounded corners=0.2cm] (4.5, 15) -- (4.5, 15.75);
\node[text width=2.500000cm, anchor=north, align=center, ] at (4.5, 16.35) {Outputs};

\end{tikzpicture}}
        \caption{VQ-TSFM E2 (encoder-decoder missingness mask conditioning).}
        \label{fig:model_a2}
    \end{subfigure}

    \caption{Overview of the proposed VQ-TSFM variants.}
    \label{fig:vqvae_model_architectures}
\end{figure*}

The architectures for the base model VQ-TSFM and its extensions E1 and E2 are illustrated in Figures \ref{fig:model_a0}, \ref{fig:model_a1}, and \ref{fig:model_a2}, respectively. Throughout the network, spatial length was preserved to ensure that each time step—representing daily patient states—was captured in the embeddings.

For real-valued features such as \textit{Sleep Start}, the mean squared error (MSE) loss was employed. This loss function was extended to continuous positive variables following the transformations described in \autoref{sec:fm_vqvae}. While the counting variables (\textit{Location Clusters} and \textit{Total Steps}) could be modeled using a Poisson distribution, the broad number of unique values ($15$ and $24,849$, respectively) allowed for an approximation using the MSE loss.

Binary features, such as \textit{Weekend} and \textit{Practiced Sport}, were trained using a modified binary cross-entropy (BCE) loss to account for class imbalances. Gradient norm clipping was applied, limiting the norm to a maximum of $2.0$ to ensure stable optimization and prevent gradient explosions in the early training phases, particularly for challenging variables such as \textit{Location Distance}. The learning rate was initially set to $1 \times 10^{-3}$, with a learning rate scheduler (\texttt{ReduceLROnPlateau}) that applied a reduction factor of 0.1 when no improvement was observed over 10 epochs.

The vector quantization (VQ) mechanism plays a key role in our architecture, particularly in the extended models E1 and E2. A codebook of $256$ vectors, initialized randomly, was employed, with the embedding dimensionality set to $80$ for all variant architectures.

To combat the problem of codebook collapse---a common challenge in VQ-VAE models---a restart threshold of $0.1$ was applied. Embeddings that were underutilized (i.e., with utilization rates below this threshold) were re-initialized to improve code utilization following \cite{dhariwal2020jukebox:}. This technique effectively mitigated collapse, as demonstrated by a monotonic increase in perplexity across training epochs. Both MCAR and MNAR experiments exhibited effective embedding utilization, which contributed to the overall performance.

As discussed in \autoref{sec:fm_vqvae}, our quantization mechanism leverages an exponential moving average (EMA) to update the embedding representations during training.  This is controlled by a decay factor and the previously mentioned threshold that prevents underutilized embeddings from being excessively penalized. As part of the quantization step, a commitment loss is calculated to measure the difference between the input and its quantized representation, ensuring smooth transitions between different embeddings. For the experiments contained in this work, we used $\beta = 0.25$ in \autoref{eq:vanilla_vqvae_loss}.

To ensure the statistical rigor of our evaluation and to assess whether the observed differences between model variants are significant, we conducted a series of hypothesis tests. The analysis aims to determine whether the VQ-TSFM model variants demonstrate statistically significant performance differences when compared to the baseline model, across various metrics. For more details, see \autoref{app:extended_res_vq_vae}.

The VQ-TSFM receives the zero-imputed signal as input, which is passed through four convolutional layers, each followed by batch normalization and a ReLU activation function. These layers use filters with kernel size $3$ with stride and padding set to $1$, ensuring that the spatial dimensions are preserved. The encoder's output is then quantized using the VQ mechanism and passed to the decoder, which consists of four deconvolutional layers. Each deconvolutional layer is followed by batch normalization and ReLU, except for the last layer, where the identity function is applied to maintain the integrity of the output values for real-valued, continuous, and counting variables, and logits for binary variables. The complete architecture for the model can be seen in \autoref{tab:model_a0_architecture}.

Version E1 incorporates the missingness mask alongside the zero-imputed signal. Prior to concatenation with the input signal, the mask undergoes processing through two convolutional layers, each followed by batch normalization and ReLU. After concatenation, the combined input is passed through six convolutional layers, similar to the implicit model but with additional depth to account for the mask information. The output is then quantized using the same VQ process, and the decoder operates identically to the base model. The complete architecture for the extended version E1 is described in \autoref{tab:model_a1_architecture}.

Version E2 extends version E1 by also passing the missingness mask to the decoder. The encoder processes the input identically to E1, quantizing the result before passing it to the decoder. In the decoder, the quantized vector is processed alongside the mask, which is passed through two additional convolutional layers. These are followed by a block of four fine-tuning layers, which enable the decoder to integrate missingness information into the final reconstructed signal. The fine-tuning layers consist of convolutional layers followed by ReLU, except for the last layer, which uses the identity function. The complete architecture for model E2 is described in \autoref{tab:model_a2_architecture}.

\begin{table}[h]
  \centering
  \setlength{\tabcolsep}{3pt}
  \renewcommand{\arraystretch}{0.85}
  \begin{threeparttable}
  \caption{VQ-TSFM Architecture: Encoder, Quantizer, and Decoder}
  \label{tab:model_a0_architecture}
   \begin{tabular}{@{}c c c l@{}}
    \toprule
    \multicolumn{4}{c}{\textbf{Encoder}} \\ 
    \midrule
    \textbf{Layer Type} 
      & \textbf{Input Dimensions} 
      & \textbf{Output Dimensions} 
      & \textbf{Details} \\ 
    \midrule
    
    \textbf{Input (Signal)} 
      & $[B,\, F,\, L]$ 
      & ---
      & Model input (signal) \\ 

    Conv-Block 1
      & $[B,\, F,\, L]$
      & $[B,\, F,\, L]$
      & Convolutional block\tnote{a} \\

    Conv-Block 2
      & $[B,\, F,\, L]$
      & $[B,\, 2F,\, L]$
      & --- \\

    Conv-Block 3
      & $[B,\, 2F,\, L]$
      & $[B,\, 4F,\, L]$
      & --- \\

    Conv-Block 4
      & $[B,\, 4F,\, L]$
      & $[B,\, 8F,\, L]$
      & --- \\

    \multicolumn{4}{c}{\textbf{Quantizer}} \\ 
    \midrule
    Quantization 
      & $[B,\, 8F,\, L]$ 
      & $[B,\, 8F,\, L]$ 
      & VQ (Nearest Lookup) \\

    \addlinespace[0.75ex]
    \multicolumn{4}{c}{\textbf{Decoder}} \\ 
    \midrule
    
    Deconv-Block 1
      & $[B,\, 8F,\, L]$
      & $[B,\, 6F,\, L]$
      & --- \\

    Deconv‐Block 2 
      & $[B,\, 6F,\, L]$ 
      & $[B,\, 4F,\, L]$ 
      & —-- \\

    Deconv‐Block 3 
      & $[B,\, 4F,\, L]$ 
      & $[B,\, 4F,\, L]$ 
      & —-- \\

    Deconv‐Block 4 
      & $[B,\, 4F,\, L]$ 
      & $[B,\, 2F,\, L]$ 
      & —-- \\

    Deconv‐Block 5 
      & $[B,\, 2F,\, L]$ 
      & $[B,\, F,\, L]$ 
      & Last deconvolutional block\tnote{b} \\

      \bottomrule
 \end{tabular}
    \begin{tablenotes}
        \item [a] Unless otherwise noted, all Conv/Deconv blocks use a kernel of size 3, with stride = 1, padding = 1; BatchNorm1D; and ReLU (in that order).

        \item[b] Identity is used instead of ReLU.
    \end{tablenotes}
    \end{threeparttable}
\end{table}

\begin{table}[!ht]
  \centering
  \setlength{\tabcolsep}{3pt}
  \renewcommand{\arraystretch}{0.85}
  \begin{threeparttable}
  \caption{VQ-TSFM E1 Architecture: Encoder, Quantizer, and Decoder}
  \label{tab:model_a1_architecture}
   \begin{tabular}{@{}c c c l@{}}
    \toprule
    \multicolumn{4}{c}{\textbf{Encoder}} \\ 
    \midrule
    \textbf{Layer Type} 
      & \textbf{Input Dimensions} 
      & \textbf{Output Dimensions} 
      & \textbf{Details} \\ 
    \midrule
    
    \textbf{Input (Signal)} 
      & $[B,\, F,\, L]$ 
      & –-- 
      & Model input (signal) \\

    \textbf{Input (Mask)} 
      & $[B,\, M,\, L]$ 
      & --- 
      & Model input (mask) \\

    Mask‐Conv‐Block 1 
      & $[B,\, M,\, L]$ 
      & $[B,\, M,\, L]$ 
      & Convolutional block\tnote{a} \\

    Mask‐Conv‐Block 2 
      & $[B,\, M,\, L]$ 
      & $[B,\, M,\, L]$ 
      & — \\

    Concatenation (Signal + Mask) 
      & $[B,\, F,\, L]$, $[B,\, M,\, L]$ 
      & $[B,\, F+M,\, L]$ 
      & Note: $F=M$ \\

    Conv‐Block 1 
      & $[B,\, F+M,\, L]$ 
      & $[B,\, F,\, L]$ 
      & --- \\

    Conv‐Block 2 
      & $[B,\, F,\, L]$ 
      & $[B,\, 2F,\, L]$ 
      & --- \\

    Conv‐Block 3 
      & $[B,\, 2F,\, L]$ 
      & $[B,\, 4F,\, L]$ 
      & --- \\

    Conv‐Block 4 
      & $[B,\, 4F,\, L]$ 
      & $[B,\, 4F,\, L]$ 
      & --- \\

    Conv‐Block 5 
      & $[B,\, 4F,\, L]$ 
      & $[B,\, 6F,\, L]$ 
      & --- \\

    Conv‐Block 6 
      & $[B,\, 6F,\, L]$ 
      & $[B,\, 8F,\, L]$ 
      & --- \\

    \multicolumn{4}{c}{\textbf{Quantizer}} \\ 
    \midrule
    Quantization 
      & $[B,\, 8F,\, L]$ 
      & $[B,\, 8F,\, L]$ 
      & VQ (Nearest Lookup) \\

    \addlinespace[0.75ex]
    \multicolumn{4}{c}{\textbf{Decoder}} \\ 
    \midrule
    
    Deconv-Block 1
      & $[B,\, 8F,\, L]$
      & $[B,\, 6F,\, L]$
      & --- \\

    Deconv‐Block 2 
      & $[B,\, 6F,\, L]$ 
      & $[B,\, 4F,\, L]$ 
      & —-- \\

    Deconv‐Block 3 
      & $[B,\, 4F,\, L]$ 
      & $[B,\, 4F,\, L]$ 
      & —-- \\

    Deconv‐Block 4 
      & $[B,\, 4F,\, L]$ 
      & $[B,\, 2F,\, L]$ 
      & —-- \\

    Deconv‐Block 5 
      & $[B,\, 2F,\, L]$ 
      & $[B,\, F,\, L]$ 
      & Last deconvolutional block\tnote{b} \\

      \bottomrule
 \end{tabular}
    \begin{tablenotes}
        \item [a] Unless otherwise noted, all Conv/Deconv blocks use a kernel of size 3, with stride = 1, padding = 1; BatchNorm1D; and ReLU (in that order).

        \item[b] Identity is used instead of ReLU.
    \end{tablenotes}
    \end{threeparttable}
\end{table}

\begin{table}[t]
  \centering
  \setlength{\tabcolsep}{3pt}
  \renewcommand{\arraystretch}{0.85}
  \begin{threeparttable}
  \caption{VQ-TSFM E2 Architecture: Encoder, Quantizer, and Decoder}
  \label{tab:model_a2_architecture}
   \begin{tabular}{@{}c c c l@{}}
    \toprule
    \multicolumn{4}{c}{\textbf{Encoder}} \\ 
    \midrule
    \textbf{Layer Type} 
      & \textbf{Input Dimensions} 
      & \textbf{Output Dimensions} 
      & \textbf{Details} \\ 
    \midrule
    
    \textbf{Input (Signal)} 
      & $[B,\, F,\, L]$ 
      & –-- 
      & Model input (signal) \\

    \textbf{Input (Mask)} 
      & $[B,\, M,\, L]$ 
      & --- 
      & Model input (mask) \\

    Mask‐Conv‐Block 1 
      & $[B,\, M,\, L]$ 
      & $[B,\, M,\, L]$ 
      & Convolutional block\tnote{a} \\

    Mask‐Conv‐Block 2 
      & $[B,\, M,\, L]$ 
      & $[B,\, M,\, L]$ 
      & — \\

    Concatenation (Signal + Mask) 
      & $[B,\, F,\, L]$, $[B,\, M,\, L]$ 
      & $[B,\, F+M,\, L]$ 
      & Note: $F=M$ \\

    Conv‐Block 1 
      & $[B,\, F+M,\, L]$ 
      & $[B,\, F,\, L]$ 
      & --- \\

    Conv‐Block 2 
      & $[B,\, F,\, L]$ 
      & $[B,\, 2F,\, L]$ 
      & --- \\

    Conv‐Block 3 
      & $[B,\, 2F,\, L]$ 
      & $[B,\, 4F,\, L]$ 
      & --- \\

    Conv‐Block 4 
      & $[B,\, 4F,\, L]$ 
      & $[B,\, 4F,\, L]$ 
      & --- \\

    Conv‐Block 5 
      & $[B,\, 4F,\, L]$ 
      & $[B,\, 6F,\, L]$ 
      & --- \\

    Conv‐Block 6 
      & $[B,\, 6F,\, L]$ 
      & $[B,\, 8F,\, L]$ 
      & --- \\

    \multicolumn{4}{c}{\textbf{Quantizer}} \\ 
    \midrule
    Quantization 
      & $[B,\, 8F,\, L]$ 
      & $[B,\, 8F,\, L]$ 
      & VQ (Nearest Lookup) \\

    \addlinespace[0.75ex]
    \multicolumn{4}{c}{\textbf{Decoder}} \\ 
    \midrule

    \textbf{Input (Quantized Signal)} 
      & $[B,\, 8F,\, L]$  
      & –--                         
      & Model input (quantized signal) \\

    \textbf{Input (Mask)}   
      & $[B,\, M,\, L]$ 
      & –--                         
      & Model input (mask) \\

    Mask‐Conv‐Block 1 
      & $[B,\, M,\, L]$ 
      & $[B,\, M,\, L]$               
      & --- \\

    Mask‐Conv‐Block 2 
      & $[B,\, M,\, L]$ 
      & $[B,\, M,\, L]$               
      & —-- \\

    Deconv‐Block 1 
      & $[B,\, 8F,\, L]$ 
      & $[B,\, 6F,\, L]$              
      & --- \\

    Deconv‐Block 2 
      & $[B,\, 6F,\, L]$ 
      & $[B,\, 4F,\, L]$              
      & —-- \\

    Deconv‐Block 3 
      & $[B,\, 4F,\, L]$ 
      & $[B,\, 4F,\, L]$              
      & —-- \\

    Deconv‐Block 4 
      & $[B,\, 4F,\, L]$ 
      & $[B,\, 2F,\, L]$              
      & —-- \\

    Deconv‐Block 5 
      & $[B,\, 2F,\, L]$ 
      & $[B,\, F,\, L]$               
      & —-- \\

    Concatenation (Quantized Signal + Mask) 
      & $[B,\, F,\, L]$, $[B,\, M,\, L]$ 
      & $[B,\, F+M,\, L]$ 
      & Note: $F=M$ \\

    Fine‐tuning-Block 1 
      & $[B,\, F+M,\, L]$ 
      & $[B,\, F+M,\, L]$           
      & --- \\

    Fine‐tuning-Block 2 
      & $[B,\, F+M,\, L]$ 
      & $[B,\, F,\, L]$               
      & —-- \\

    Fine‐tuning-Block 3 
      & $[B,\, F,\, L]$ 
      & $[B,\, F,\, L]$               
      & —-- \\

    Fine‐tuning-Block 4 
      & $[B,\, F,\, L]$ 
      & $[B,\, F,\, L]$               
      & Last deconvolutional block\tnote{b} \\

    \bottomrule
 \end{tabular}
    \begin{tablenotes}
        \item [a] Unless otherwise noted, all Conv/Deconv blocks use a kernel of size 3, with stride = 1, padding = 1; BatchNorm1D; and ReLU (in that order).

        \item[b] Identity is used instead of ReLU.
    \end{tablenotes}
    \end{threeparttable}
\end{table}

\clearpage
\section{Constructing VQ-TSFM Latent Profiles for CPD}\label{app:vqvae_latent_cpd}

In preparing VQ-TSFM profiles for use in the CPD task, we leverage the inherent sparsity of the learned representations. This sparsity not only enhances the interpretability of the patient time-series embeddings but also allows for efficient and accurate change-point detection, critical in real-world applications for patient behavior monitoring for psychiatric patients.

VQ-TSFM representations often exhibit significant variations in the frequency of usage across embeddings. To capitalize on this, we introduce a ranking system based on the frequency of each embedding's occurrence. Embeddings that appear frequently within the time-series sample are ranked higher, as these are likely to represent more common patterns. Conversely, embeddings that are infrequently used (below a certain number of \enquote{most used embeddings}) are considered outliers and grouped into a special category referred to as the \enquote{dummy} embedding. This dummy embedding is more than a placeholder; it reflects rare or anomalous patterns, which may acquire specific clinical interpretations, such as periods of abnormal patient behavior or sensor malfunction. In particular, for the CPD results shown in \autoref{fig:cpd_results}, only a small number of individual embeddings ranging from 5 to 30 (depending on the specific setting)---out of the total 256 in the codebook---were considered, with the remaining, less-used instances being classified into the \enquote{dummy} embedding. An ablation study regarding the number of individual embeddings considered for the CPD algorithm is provided in \autoref{app:cpd_details}.

By categorizing uncommon embeddings into a collective representation, we enhance the robustness of downstream analysis, as this method mitigates the noise introduced by outlier embeddings (themselves caused by outlier, and often erroneous, data) while retaining the capacity to detect important deviations in patient behavior.

As mentioned in \autoref{sec:cpd}, CPD can be approached in both deterministic and probabilistic modes, depending on the level of certainty required in detecting shifts in patient behavior. To support both approaches, we compute pseudo-probabilities derived from the distances between the quantized embeddings and the original continuous outputs of the encoder. Since the latent space of VQ-TSFM is discrete, pseudo-probabilities are computed by first calculating the Euclidean distances between the continuous encoder outputs and the set of embeddings in the latent space. These distances quantify how close or far each input is from each embedding. Next, the softmax function is applied to the additive inverse of these distances, transforming them into a probability distribution over all possible embeddings. This transformation ensures that embeddings closer to the continuous encoder output (i.e., those with smaller Euclidean distances) are assigned higher pseudo-probabilities, while more distant embeddings are assigned lower pseudo-probabilities, thereby approximating a probabilistic interpretation for the otherwise discrete latent profiles.

These probabilities provide a soft assignment, offering an interpretable measure of how well an embedding fits the original data point. This is particularly useful in probabilistic CPD, where transitions between states are inherently uncertain, and the distances can be used to modulate the likelihood of a change-point. By integrating both deterministic hard-assignments and probabilistic soft-assignments, our framework allows for flexible CPD that can adapt to different levels of interpretability and precision, essential for clinical scenarios.

\newpage
\section{CPD Algorithm Details and Ablation Study} \label{app:cpd_details}

The change-point detector (CPD) model used in this work was designed with many customization options, including CPD versions, hyperparameters, and alternative methods. Some of these options are explained in detail next. The most important setting is the CPD version to be used, depending on the input data:

\begin{itemize}
    \item \textbf{Hierarchical CPD} \cite{Moreno2019}. As explained in \autoref{sec:cpd_task}, instead of directly analyzing the high-dimensional observations, the hierarchical CPD is fed with a latent variable (a profile sequence of discrete nature) and infers the posterior distribution of changes in such pseudo-observations. This approach simplifies the detection process and reduces computational complexity. However, when the distributions of the latent variables are flat or uncertain, the hierarchical CPD performance can be compromised due to noisy point estimates (i.e., the categorical estimation of the profiles is not modeled with confidence).

    \item \textbf{Multinomial CPD} \cite{Romero2022}. Adapted to work with profile distributions, this version addresses the limitation mentioned above by incorporating multinomial sampling to better characterize the uncertainty in latent variable inference. Instead of relying solely on point estimates, the multinomial CPD draws multiple samples from the posterior distribution of latent variables at each time step and constructs a counting vector representing the frequency of each latent class within the samples. By considering the uncertainty in latent variable inference, the multinomial CPD improves detection rate and enhances robustness to noisy or missing data.

    \item \textbf{Multivariate CPD}. This last version of the CPD has been designed to accept multivariate embeddings in a real space, which may correspond to raw behavioral data or real-valued numerical embeddings. To process such input, the algorithm employs a multivariate Gaussian likelihood to model the data, an inverse-Wishart for the prior conjugate, and a multivariate Student's \textit{t}-distribution to calculate the predictive probability.
\end{itemize}

\begin{figure*}[h]
    \centering
    \begin{subfigure}{0.32\textwidth}
        \centering
        \includegraphics[width=\textwidth]{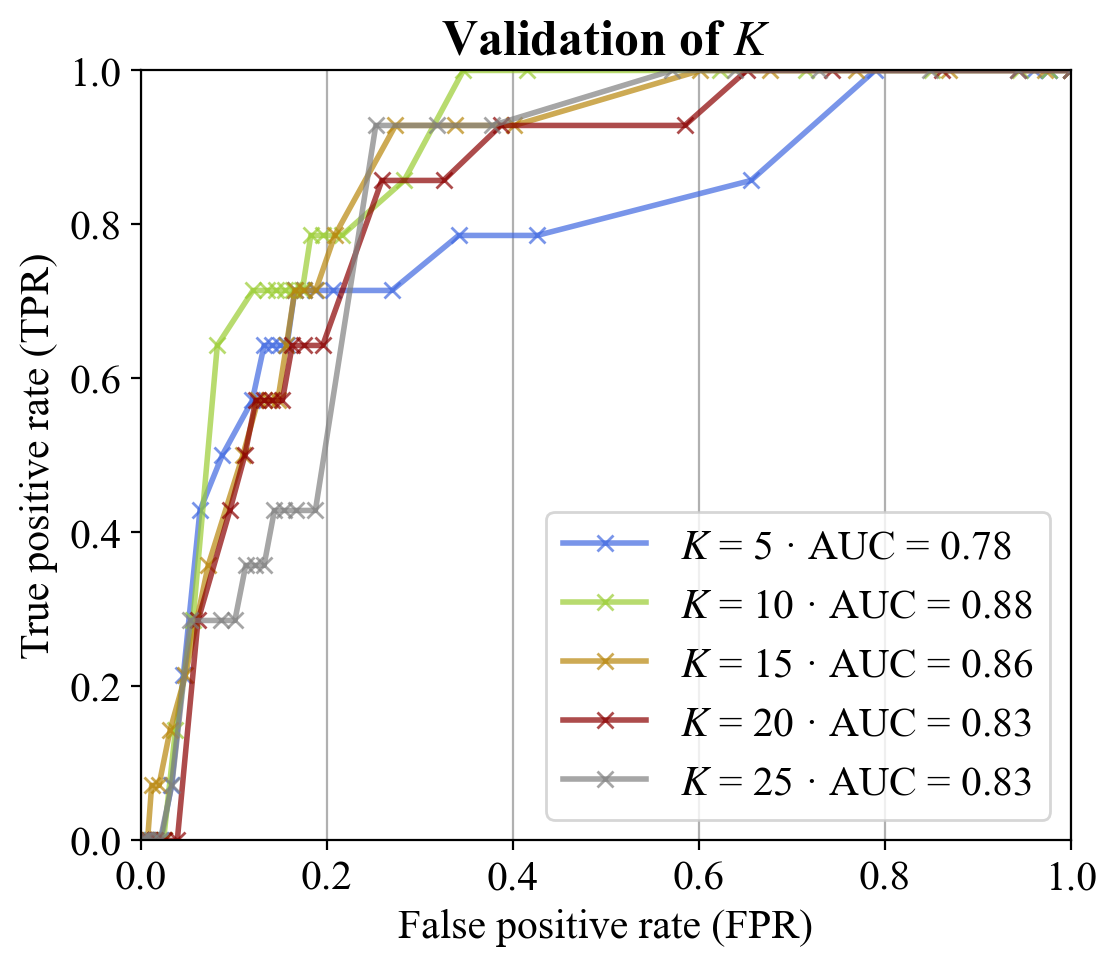}
        \caption{}\label{fig:cpd_ablation_k}
    \end{subfigure}
    \hfill
    \begin{subfigure}{0.32\textwidth}
        \centering
        \includegraphics[width=\textwidth]{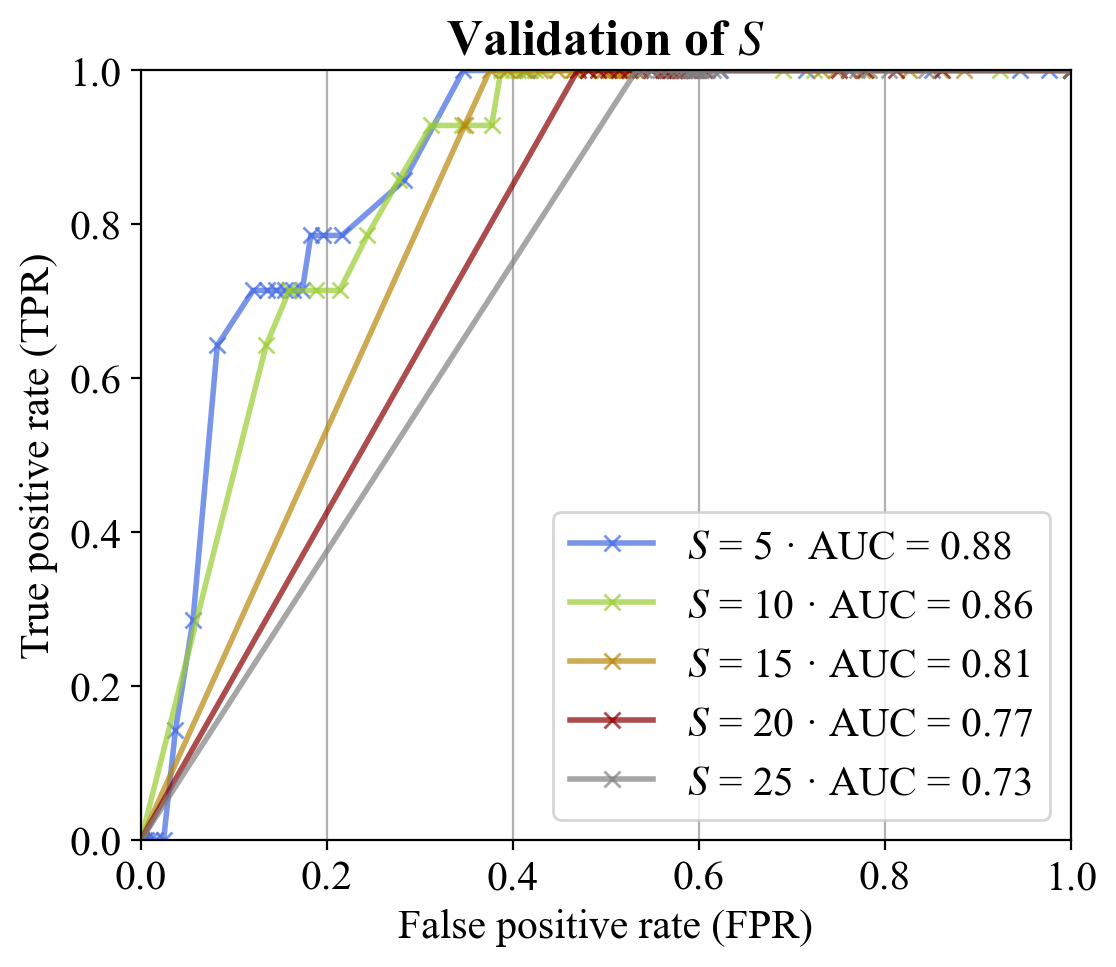}
        \caption{}\label{fig:cpd_ablation_s}
    \end{subfigure}
    \hfill
    \begin{subfigure}{0.32\textwidth}
        \centering
        \includegraphics[width=\textwidth]{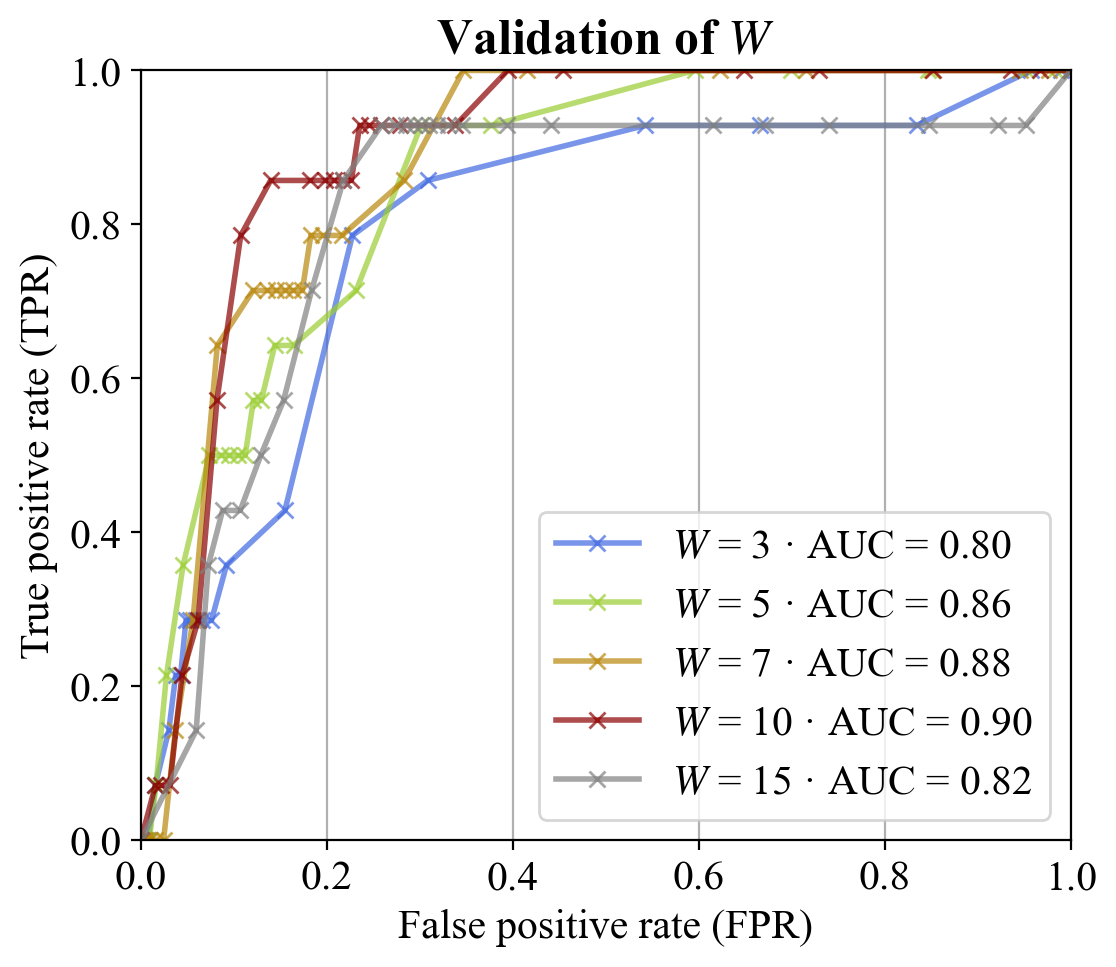}
        \caption{}\label{fig:cpd_ablation_w}
    \end{subfigure}
    \caption{ROC curves obtained from a hyperparameter analysis on the HetMM–CPD integration, testing a range of values of (a) the number of profiles $K$, (b) the number of samples $S$ and (c) the size of the temporal window $W$. The configuration of the baseline HetMM–CPD pipeline used as reference was set to 10 profiles (the best-performing value), 5 samples and a 7-day window size.}
    \label{fig:cpd_ablation}
\end{figure*}

Some of the hyperparameters involved in the downstream task were fixed based on our previous experience working with the HetMM–CPD pipeline, while others were subject to an ablation study to identify the best configuration. A brief description is given for each hyperparameter. The optimal values mentioned here were used to produce the figures and tables of this manuscript.

\begin{itemize}
    \item \textbf{Number of profiles, $K$}. While not a hyperparameter of the CPD stage (but rather involved in the previous profiling step), the number of possible profiles is a crucial setting in the downstream task. Too few profiles will fail to capture the distinct behavior patterns, but too many may introduce noisy profiles modeled with low confidence that impede the correct performance of the CPD. The value of $K$ in the heterogeneous mixture model was analyzed (\autoref{fig:cpd_ablation_k}) and chosen to be 10, the one yielding the best results. Notice that the VQ-TSFM model used to compare may use a different number of profiles. In fact, \autoref{fig:ablation_vqvae_extended} suggests that $K=20$ is the most suitable value in the VQ-TSFM context. On the other hand, the Informer approach does not have this parameter because its latent representation are real-valued embeddings.

    \item \textbf{Number of samples in multinomial distribution, $S$}. In the multinomial approach, $S$ represents the number of samples that are drawn from the posterior distribution of the latent variables at each time step. A larger value will adapt better to the latent profiles but also complicates the detection task of the CPD. As evidenced by the results shown in \autoref{fig:cpd_ablation_s}, $S=5$ provides the most suitable sample size for our task.

    \item \textbf{Prior change-point probability, $\lambda$}. As explained in \autoref{sec:cpd}, $\lambda$ is involved in the hazard function that defines the prior probability of having a change-point at any instant. This constant can be tuned to adapt the CPD's sensibility and a few values were included in the results offered in \autoref{fig:cpd_results} of \autoref{sec:cpd_task}, in order to have a richer perspective on this hyperparameter.

    \item \textbf{Size of the temporal window, $W$}. The CPD model focuses on a temporal frame to assess whether its predictions are successful or not. For example, for each true event, a true positive is returned if an alarm was given by the model within the temporal window previous to that event. If the CPD did not predict any change, then a false negative is counted. This window parameter allows therefore to select how long in advance we aim to predict suicide events. We chose a prediction period of one week ($W=7$ days), which obtained a high AUC in our analysis (see \autoref{fig:cpd_ablation_w}) and is brief enough to serve as short-term prediction.

\begin{figure*}[b!h]
    \centering
    \includegraphics[width=0.99\textwidth]{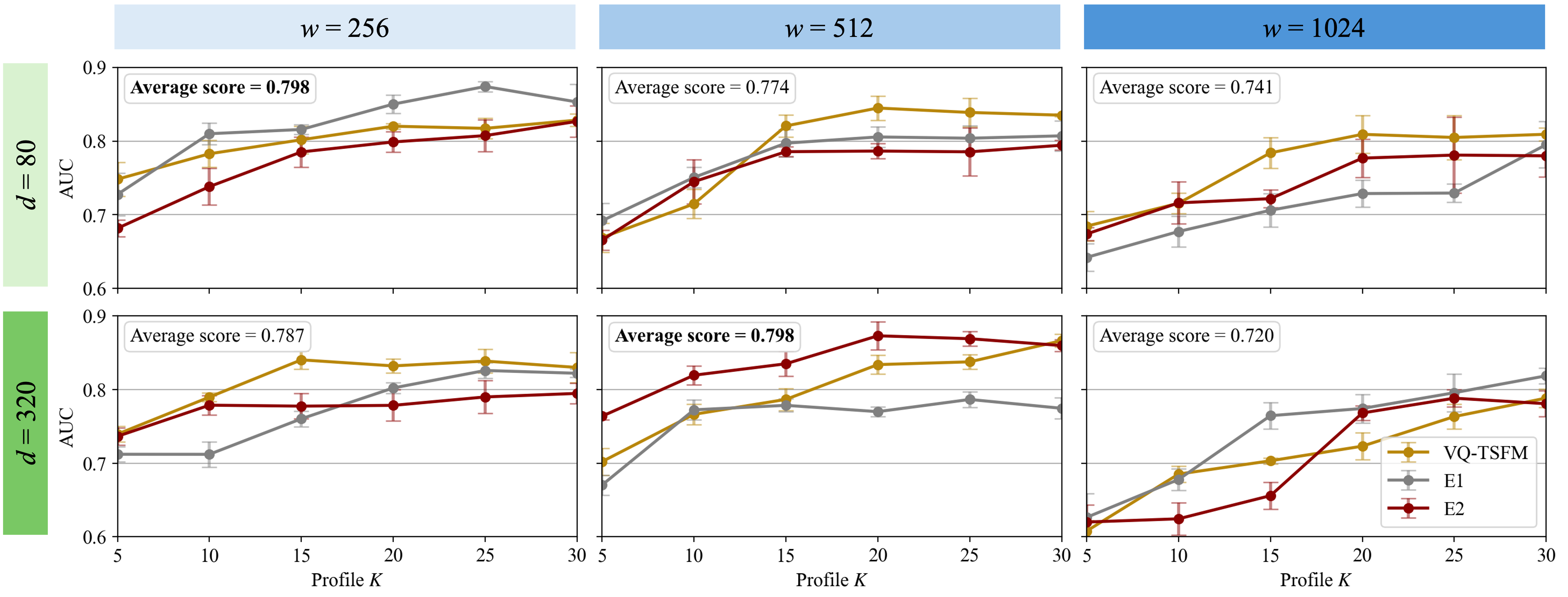}
    \caption{AUC scores of suicide prediction obtained with different VQ-TSFM variations (using the pseudo-probability output). Each subplot compares the model version (VQ-TSFM, E1 and E2) and the number of profiles $K$, whereas the whole figure displays the results for different embedding configurations, changing the embedding size ($d$) and dictionary size ($w$). A point in the plots represents, for the corresponding model, the average AUC score from the ROC curves using $\lambda=\{10,10^3,10^5,10^7\}$. All points in one subplot are averaged to compute the \textit{Average score} in the top left-hand corner.}
    \label{fig:ablation_vqvae_extended}
\end{figure*}

    \item \textbf{Threshold, $\tau$}. The last hyperparameter affects the definition of alarms or positive predictions (i.e., the conversion from run length to a binary detection vector, which is necessary to contrast model predictions against real events). By sweeping a range of values of this threshold and computing the sensitivity-specificity pair for each of them, the receiver operating characteristic (ROC) curve can be produced. However, different methods to define the decision threshold can be implemented by the CPD model. The \textit{cumulative sum} was used in this work because it gives an interpretable measure of behavioral instability during the past days that helps clinicians understand the potential risk of suicidal behavior for the patient.

    \begin{itemize}
        \item \textit{MAP ratio} (default) → based on the MAP estimates of the run length, an alarm is returned if the ratio of current $r_t$ over the previous day $r_{t-1}$ is below the threshold: $$\frac{r_t}{r_{t-1}} < \tau$$
        \item \textit{MAP difference} → based on the MAP estimates of the run length, an alarm is returned if the difference between current $r_t$ and previous $r_{t-1}$ is above the threshold: $$r_t - r_{t-1} > \tau$$
        \item \textit{Cumulative sum} → based on the cumulative probability of the run length of previous days (within the specified window of size $W$), an alarm is returned if this sum is above the threshold: $$\sum\limits_{i=0}^{W} r_{t-i} > \tau$$
    \end{itemize}
\end{itemize}

Regarding the incorporation of the VQ-TSFM encoded space as input to the CPD, we tested the implicit model against its two explicit extensions (E1 and E2, explained in \autoref{app:vqvae_details}), and for a range of numbers of embeddings (i.e., the number of possible profiles used in the subject characterization, $K$, after introducing the "dummy" profile). The results are displayed in the different subplots within \autoref{fig:ablation_vqvae_extended}, which is an extended version of \autoref{fig:ablation_vqvae} in the main body. These graphs were obtained using the VQ-TSFM's pseudo-probabilities. Poorer outcomes were obtained when less profiles were used ($K=5$, $K=10$) and the AUC score generally stabilized around $K=20$, providing a reason to set this parameter to 20. On the other hand, there is no clear model version outperforming the others: both the implicit VQ-TSFM and its variations E1 and E2 yielded optimal results in at least one of the subplots. However, while VQ-TSFM did not achieve overall peak performance, it consistently exhibited strong results and thus was chosen to elaborate \autoref{tab:cpd_trade_off} and Figures \ref{fig:cpd_vqvae_disc} and \ref{fig:cpd_vqvae_prob} in the results section. Finally, the whole \autoref{fig:ablation_vqvae_extended} compares the configuration of the VQ-TSFM embeddings: their dimension length $d$ and the size of the dictionary $w$. The most evident interpretation is that increasing the dictionary size to 1024 lead to a substantial decrease in performance. Conversely, no clear deductions can be made regarding the embedding size. The best overall performance was achieved by model E2 with $K=20$ and embeddings of $d=320$ and $w=512$, reaching an AUC score of 0.90.

\clearpage
\section{Architectural Details on the 1D CNN for Emotion Forecasting} \label{app:emotion_details}

The classifier used to predict emotions based on the output of the VQ-TSFM encoder was a one-dimensional CNN network with two convolutional layers followed by two fully-connected layers. Results in \autoref{sec:emotion} were obtained with embeddings of initial dimension 80 or 320 and based on a window of 7 days. Therefore, the size of the input tensors for this CNN was $7\times80$ or $7\times320$. The following configuration was chosen after a comprehensive ablation study focused on reducing overfitting without compromising performance.

\begin{itemize}
    \item \textbf{Architecture}. Two convolutional layers reduce the length of the input sequences by half while incrementing the number of channels (from 7 to 32, and from 32 to 64). Next, two linear layers reduce the total dimension of the signal (to 128 and then to 3), returning a three-dimensional output that corresponds to the three possible emotions: negative, neutral or positive. Each convolutional layer is followed by a ReLU activation function, max pooling (kernel size 2), batch normalization and dropout layer (25\% probability). The first linear layer is followed by ReLU and dropout (10\% probability).

    \item \textbf{Training}. The training process was made in batches of 64 entries and for a maximum of 100 epochs, although a validation set (30\% of the training set) was reserved to implement early stopping when the validation loss did not improve after ten consecutive epochs (patience = 10).

    \item \textbf{Optimization}. The network was optimized through stochastic gradient descent (Adam optimizer) with a learning rate of 0.001 and L2 regularization (weight decay of 0.001), using the cross entropy loss function.
\end{itemize}

\newpage
\section{Extended Results on the VQ-TSFM Foundation Model}
\label{app:extended_res_vq_vae}

\subsection{Signal Reconstruction and Imputation}
\label{app:signal_recons_imput}

\autoref{tab:mean_std_recons_imput_performance} presents the reconstruction performance in terms of MAE (or F1 score for the binary variables \textit{Weekend} and \textit{Practiced Sport}) for observed data, as well as for missing data under both MCAR and MNAR mechanisms. The results indicate that both VQ-TSFM and the two explicit missing-conditioned variants perform comparably across most variables, with some nuanced differences. For example, version E2 performs better on reconstructing observed instances of \textit{Sleep Start}, achieving lower mean absolute error (MAE) compared to VQ-TSFM and E1. Conversely, these two perform better than E2 for reconstructing observed instances of \textit{Time at Home} and \textit{Sleep Duration}. Additionally, the VQ-TSFM achieves the lowest error for the observed instances of \textit{Total Steps}.

\begin{table}[!h]
\centering
\setlength{\tabcolsep}{3pt}
\renewcommand{\arraystretch}{0.85}
\caption{Performance of VQ-TSFM and extended versions E1 and E2}
\label{tab:mean_std_recons_imput_performance}
\begin{tabular}{@{} llccc @{}}
\toprule
\textbf{Variable} & \textbf{Type} & \textbf{VQ-TSFM} & \textbf{E1} & \textbf{E2} \\
\midrule

\multirow{3}{*}{\textbf{Sleep Start (s)}} 
  & XO      & $1315.63 \pm 47.06$ & $1242.66 \pm 57.88$ & $\mathbf{1177.78 \pm 57.75}$ \\
  & MCAR    & $5777.24 \pm 229.41$ & $5651.99 \pm 245.31$ & $5578.96 \pm 496.26$ \\
  & MNAR    & $5896.85 \pm 492.96$ & $5718.97 \pm 417.62$ & $5607.64 \pm 593.95$ \\
\midrule

\multirow{3}{*}{\textbf{Distance (m)}} 
  & XO      & $12202.43 \pm 1296.66$ & $11627.66 \pm 937.86$ & $12874.13 \pm 836.27$ \\
  & MCAR    & $17008.33 \pm 7488.46$ & $16681.98  \pm 13920.55$ & $15190.03 \pm 3520.84$ \\
  & MNAR    & $15100.38 \pm 2035.91$ & $14232.06 \pm 1821.58$ & $15175.21 \pm 2363.39$ \\
\midrule

\multirow{3}{*}{\textbf{Time at Home (m)}} 
  & XO      & $\mathbf{146.17 \pm 4.95}$ & $\mathbf{143.58 \pm 8.58}$ & $174.94 \pm 9.70$ \\
  & MCAR    & $289.52 \pm 17.03$ & $290.18 \pm 17.87$ & $291.85 \pm 18.18$ \\
  & MNAR    & $287.52 \pm 16.05$ & $282.68 \pm 15.94$ & $286.16 \pm 13.35$ \\
\midrule

\multirow{3}{*}{\textbf{Sleep Duration (s)}} 
  & XO      & $\mathbf{4149.40 \pm 120.98}$ & $\mathbf{4055.13 \pm 151.20}$ & $5005.76 \pm 211.03$ \\
  & MCAR    & $6563.44 \pm 282.73$ & $6615.74 \pm 309.10$ & $6738.00 \pm 398.30$ \\
  & MNAR    & $6422.58 \pm 340.45$ & $6373.11 \pm 232.31$ & $6585.21 \pm 300.78$ \\
\midrule

\multirow{3}{*}{\textbf{Time Walking (s)}} 
  & XO      & $1341.44 \pm 65.39$ & $1298.03 \pm 61.20$ & $1279.72 \pm 67.14$ \\
  & MCAR    & $1779.98 \pm 145.89$ & $1742.47 \pm 101.91$ & $1734.54 \pm 73.66$ \\
  & MNAR    & $1676.90 \pm 82.56$ & $1657.30 \pm 96.37$ & $1744.46 \pm 105.72$ \\
\midrule

\multirow{3}{*}{\textbf{App Usage (s)}} 
  & XO      & $3784.17 \pm 348.70$ & $3714.48 \pm 315.91$ & $3968.00 \pm 357.25$ \\
  & MCAR    & $5045.95 \pm 528.72$ & $4973.86 \pm 558.61$ & $4946.72 \pm 744.72$ \\
  & MNAR    & $4436.77 \pm 669.15$ & $4303.00 \pm 760.17$ & $4310.54 \pm 655.41$ \\
\midrule

\multirow{3}{*}{\textbf{Location Clusters}} 
  & XO      & $1.0887 \pm 0.0716$ & $1.0746 \pm 0.0833$ & $1.2469 \pm 0.0987$ \\
  & MCAR    & $1.3234 \pm 0.1120$ & $1.3143 \pm 0.1094$ & $1.3980 \pm 0.1100$ \\
  & MNAR    & $1.3210 \pm 0.1887$ & $1.2900 \pm 0.1907$ & $1.3835 \pm 0.1645$ \\
\midrule

\multirow{3}{*}{\textbf{Total Steps}} 
  & XO      & $\mathbf{2101.48 \pm 348.70}$ & $3714.48 \pm 315.91$ & $3968.00 \pm 357.25$ \\
  & MCAR    & $3056.67 \pm 137.87$ & $3002.53 \pm 230.60$ & $2993.74 \pm 204.87$ \\
  & MNAR    & $3042.64 \pm 130.44$ & $2986.37 \pm 175.30$ & $2986.15 \pm 164.41$ \\
\midrule

\multirow{1}{*}{\textbf{Weekend}} 
  & XO      & $0.9950 \pm 0.0010$ & $0.9960 \pm 0.0015$ & $0.9967 \pm 0.0013$ \\
\midrule

\multirow{1}{*}{\textbf{Practiced Sport}} 
  & XO            & $0.9932 \pm 0.0016$ & $0.9941 \pm 0.0023$ & $0.9929 \pm 0.0021$ \\
\bottomrule
\multicolumn{5}{@{}p{0.8\columnwidth}@{}}{Metrics for Variables 0-7 are reported in MAE (lower is better), and Variables 8-9 are evaluated using F1 (higher is better).}
\end{tabular}
\end{table}

Despite not being explicitly optimized for imputation, the models performed competently in this task. These results highlight the models' ability to generalize beyond their training objective, particularly under the MNAR condition, where missingness is more structured and challenging. This is compounded by the fact that the discrete profile representation provided by VQ-TSFM is sparse, i.e., out of the total 256 embeddings in the codebook, only a few were used for each patient, thereby enhancing interpretability (see Appendix \ref{app:embed_usage_hist} for embedding utilization histograms).

It is important to note that no synthetic missingness was applied to the variables \textit{Weekend} and \textit{Practiced Sport}, as these were fully observed across the dataset. Consequently, the MCAR and MNAR scenarios were not applicable for these variables. Nonetheless, the consistently high F1 scores (close to $1.0$) achieved by all models for these categorical variables reinforce the robustness of the learned representations, even for variables without missing data.

Hypothesis testing was performed for a more in-depth analysis to assess the statistical significance of the observed differences between the models. We began by testing the normality of the data using the Shapiro-Wilk test. The null hypothesis ($H_0$) for this test states that the data comes from a normally distributed population. Conversely, the alternative hypothesis ($H_1$) posits that the data is not normally distributed. We employed a significance level of $\alpha = 0.05$. If the $p$-value from the Shapiro-Wilk test is greater than $0.05$, we fail to reject the null hypothesis, indicated that the data can be assumed to follow a normal distribution.\footnote{The significance levels used in these tests ensure that any rejection of the null hypothesis corresponds to a less than 5\% probability of a Type I error, i.e., that it is rejected while being true. In the case of the Shapiro-Wilk and Wilcoxon signed-rank tests this would represent the scenario in which it is incorrectly concluded that the models differ when they do not.}

\begin{table*}[ht]
\centering
\setlength{\tabcolsep}{3pt}
\renewcommand{\arraystretch}{0.85}
\caption{Shapiro-Wilk test for normality for VQ-TSFM and extended versions E1 and E2}

\label{tab:shapiro_wilk_test}
\begin{tabular}{@{} llcccccc @{}}
\toprule
\textbf{Variable} & \textbf{Type} & \textbf{VQ-TSFM (W)} & \textbf{VQ-TSFM (\textit{p})} & \textbf{E1 (W)} & \textbf{E1 (\textit{p})} & \textbf{E2 (W)} & \textbf{E2 (\textit{p})} \\
\midrule

\multirow{3}{*}{\textbf{Sleep Start (s)}} 
  & XO      & $0.9870$ & $0.9197$ & $0.9515$ & $0.0854$ & $0.9639$ & $0.2274$ \\
  & MCAR    & $0.9654$ & $0.2542$ & $0.9877$ & $0.9358$ & $0.9758$ & $0.5371$ \\
  & MNAR    & $0.9544$ & $0.1074$ & $0.9352$ & $0.0240$ (\ding{55}) & $0.9839$ & $0.8290$ \\
\midrule

\multirow{3}{*}{\textbf{Distance (m)}} 
  & XO      & $0.7935$ & $5\times 10^{-6}$ (\ding{55}) & $0.9768$ & $0.5723$ & $0.9827$ & $0.7863$ \\
  & MCAR    & $0.4596$ & $5.9\times 10^{-11}$ (\ding{55}) & $0.2506$ & $5\times 10^{-13}$ (\ding{55}) & $0.4973$ & $1.6\times 10^{-10}$ (\ding{55})\\
  & MNAR    & $0.9714$ & $0.3969$ & $0.9756$ & $0.5311$ & $0.9748$ & $0.5023$ \\
\midrule

\multirow{3}{*}{\textbf{Time at Home (m)}} 
  & XO      & $0.9645$ & $0.2387$ & $0.9537$ & $0.1016$ & $0.9589$ & $0.1530$ \\
  & MCAR    & $0.9862$ & $0.8978$ & $0.9402$ & $0.0351$ (\ding{55}) & $0.9700$ & $0.3595$ \\
  & MNAR    & $0.9668$ & $0.2833$ & $0.9604$ & $0.1734$ & $0.9576$ & $0.1387$ \\
\midrule

\multirow{3}{*}{\textbf{Sleep Duration (s)}} 
  & XO      & $0.9720$ & $0.4141$ & $0.9548$ & $0.1113$ & $0.9639$ & $0.2270$ \\
  & MCAR    & $0.9658$ & $0.2636$ & $0.9640$ & $0.2292$ & $0.9803$ & $0.7008$ \\
  & MNAR    & $0.9654$ & $0.2545$ & $0.9782$ & $0.6245$ & $0.9484$ & $0.0668$ \\
\midrule

\multirow{3}{*}{\textbf{Time Walking (s)}} 
  & XO      & $0.9682$ & $0.3155$ & $0.9617$ & $0.1913$ & $0.9706$ & $0.3751$ \\
  & MCAR    & $0.7455$ & $5.9\times 10^{-7}$ (\ding{55}) & $0.9734$ & $0.4593$ & $0.9868$ & $0.9138$ \\
  & MNAR    & $0.9747$ & $0.4988$ & $0.8987$ & $0.0017$ (\ding{55}) & $0.9864$ & $0.9046$ \\
\midrule

\multirow{3}{*}{\textbf{App Usage (s)}} 
  & XO      & $0.9629$ & $0.2106$ & $0.9611$ & $0.1821$ & $0.9596$ & $0.1620$ \\
  & MCAR    & $0.9700$ & $0.3602$ & $0.9782$ & $0.6242$ & $0.7979$ & $6.1\times 10^{-6}$ (\ding{55})\\
  & MNAR    & $0.9259$ & $0.0119$ (\ding{55}) & $0.9248$ & $0.010$ (\ding{55}) & $0.9733$ & $0.4549$ \\
\midrule

\multirow{3}{*}{\textbf{Location Clusters}} 
  & XO      & $0.9576$ & $0.1386$ & $0.9642$ & $0.2321$ & $0.9838$ & $0.8272$ \\
  & MCAR    & $0.9754$ & $0.5245$ & $0.9567$ & $0.1290$ & $0.9443$ & $0.0487$ (\ding{55}) \\
  & MNAR    & $0.9612$ & $0.1841$ & $0.9717$ & $0.4063$ & $0.9742$ & $0.4836$ \\
\midrule

\multirow{3}{*}{\textbf{Total Steps}} 
  & XO      & $0.9574$ & $0.1366$ & $0.9696$ & $0.3496$ & $0.9790$ & $0.6536$ \\
  & MCAR    & $0.9745$ & $0.4929$ & $0.9057$ & $0.0028$ (\ding{55}) & $0.9232$ & $0.0097$ (\ding{55}) \\
  & MNAR    & $0.9800$ & $0.6911$ & $0.9818$ & $0.7552$ & $0.9487$ & $0.0683$ \\
\midrule

\multirow{1}{*}{\textbf{Weekend}} 
  & XO      & $0.9849$ & $0.9849$ & $0.9752$ & $0.5162$ & $0.9617$ & $0.9617$ \\
\midrule

\multirow{1}{*}{\textbf{Practiced Sport}} 
  & XO            & $0.9397$ & $0.0338$ (\ding{55}) & $0.7819$ & $2.9\times 10^{-6}$ (\ding{55}) & $0.9503$ & $0.0779$ \\
\bottomrule
\multicolumn{8}{@{}p{0.95\columnwidth}@{}}{The table reports the test statistic (W) and \textit{p}-values for each model and variable under different conditions (XO, MCAR, and MNAR). $\alpha = 0.05$ was used and \ding{55} denotes the rejection of the null at the $\alpha$ significance level, implying non-normal distribution.}
\end{tabular}
\end{table*}

The Shapiro-Wilk test results are provided in \autoref{tab:shapiro_wilk_test}. If both models' result (i.e., the variant model and baseline) for a given variable and type passed the normality test, we proceeded with the paired Welch t-test. If the null hypothesis was rejected for either one of the two models (i.e., the data is not normally distributed), we opted for the non-parametric Wilcoxon signed-rank test.

When the data for both the baseline and the variant model were found to be normally distributed, we used the paired Welch's t-test to compare their means. The null hypothesis for this test asserts that there is not difference between the means of the two models, whereas the alternative hypothesis suggests a significant difference between them. We again used a significance level of $\alpha = 0.05$, rejecting the null hypothesis if the p-value was below this threshold. The results for the paired Welch t-tests are summarized in \autoref{tab:paired_welch_ttest}.

For cases where the data for one or both models did not pass the Shapiro-Wilk normality test, we employed the Wilcoxon signed-rank test. This non-parametric test does not assume normality.\footnote{A requirement of the Wilcoxon signed-rank test is symmetry.} The null hypothesis here is that the distributions of the two models are identical, whereas the alternative hypothesis suggests a significant difference between them. Similar to the Welch t-test, we used $\alpha =0.05$ as the significance level. \autoref{tab:paired_wilcoxon} provides a detailed summary of the Wilcoxon signed-rank test results.

\begin{figure*}[t]
    \centering
    \begin{subfigure}[b]{0.37\textwidth}
        \centering
        \includegraphics[width=\textwidth]{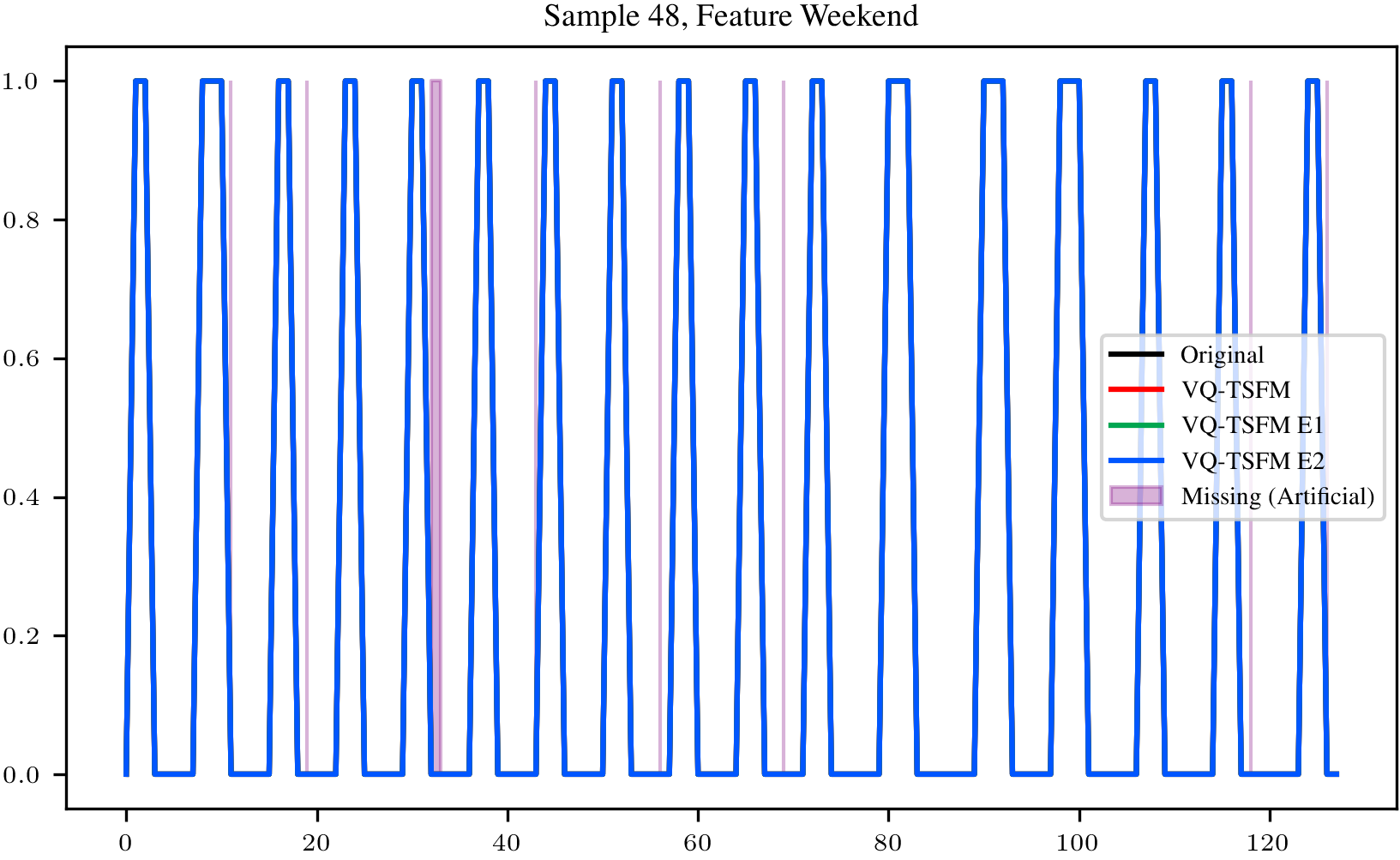}
        \caption{Recons. of sample 48 for Weekend.}
        \label{fig:reco_vqvae_s_48_weekend}
    \end{subfigure}
    \begin{subfigure}[b]{0.37\textwidth}
        \centering
        \includegraphics[width=\textwidth]{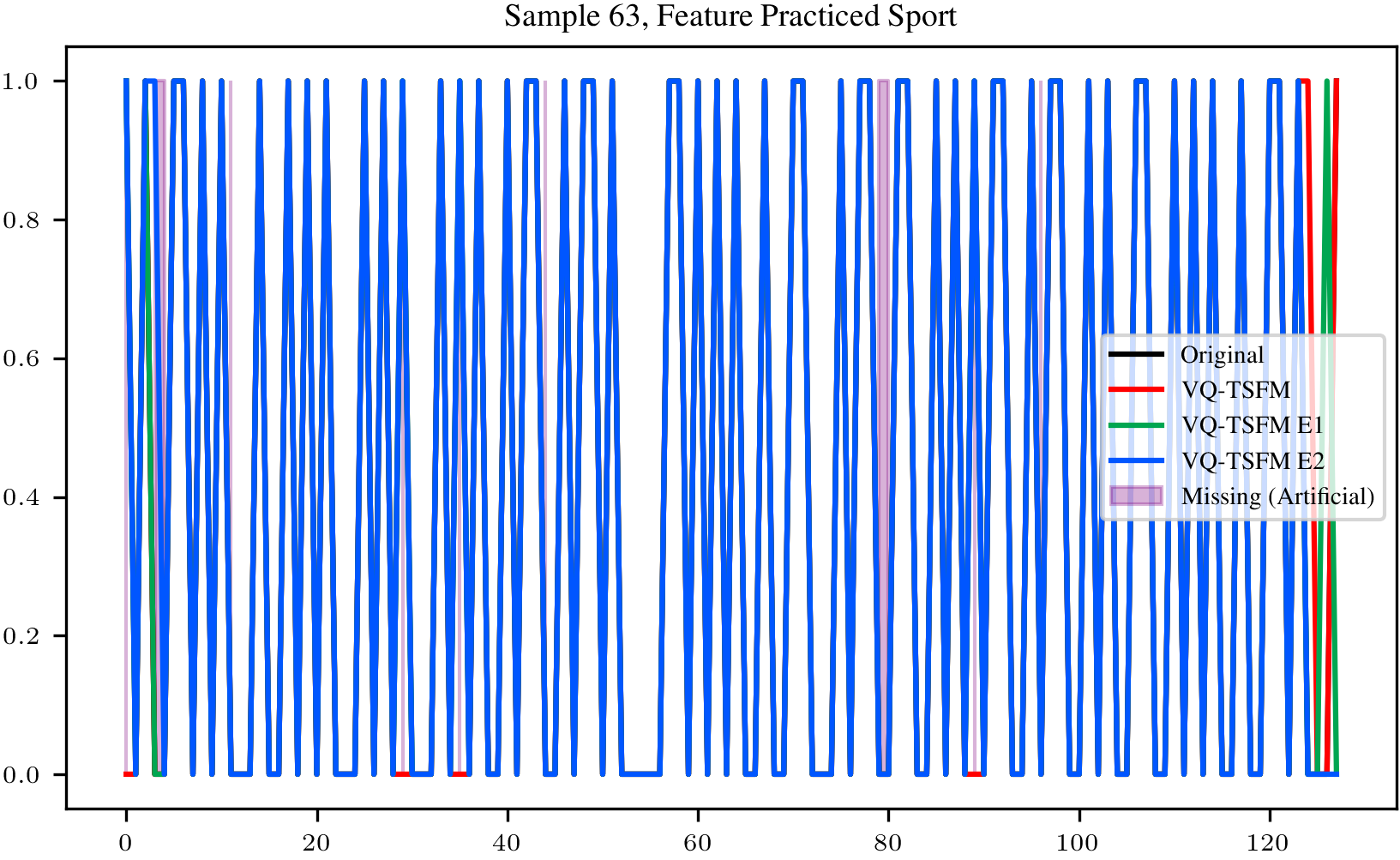}
        \caption{Recons. of sample 63 for Practiced Sport.}
        \label{fig:reco_vqvae_s_63_practiced_sport}
    \end{subfigure}
    
    \vspace{0.05cm}
    
    \begin{subfigure}[b]{0.37\textwidth}
        \centering
        \includegraphics[width=\textwidth]{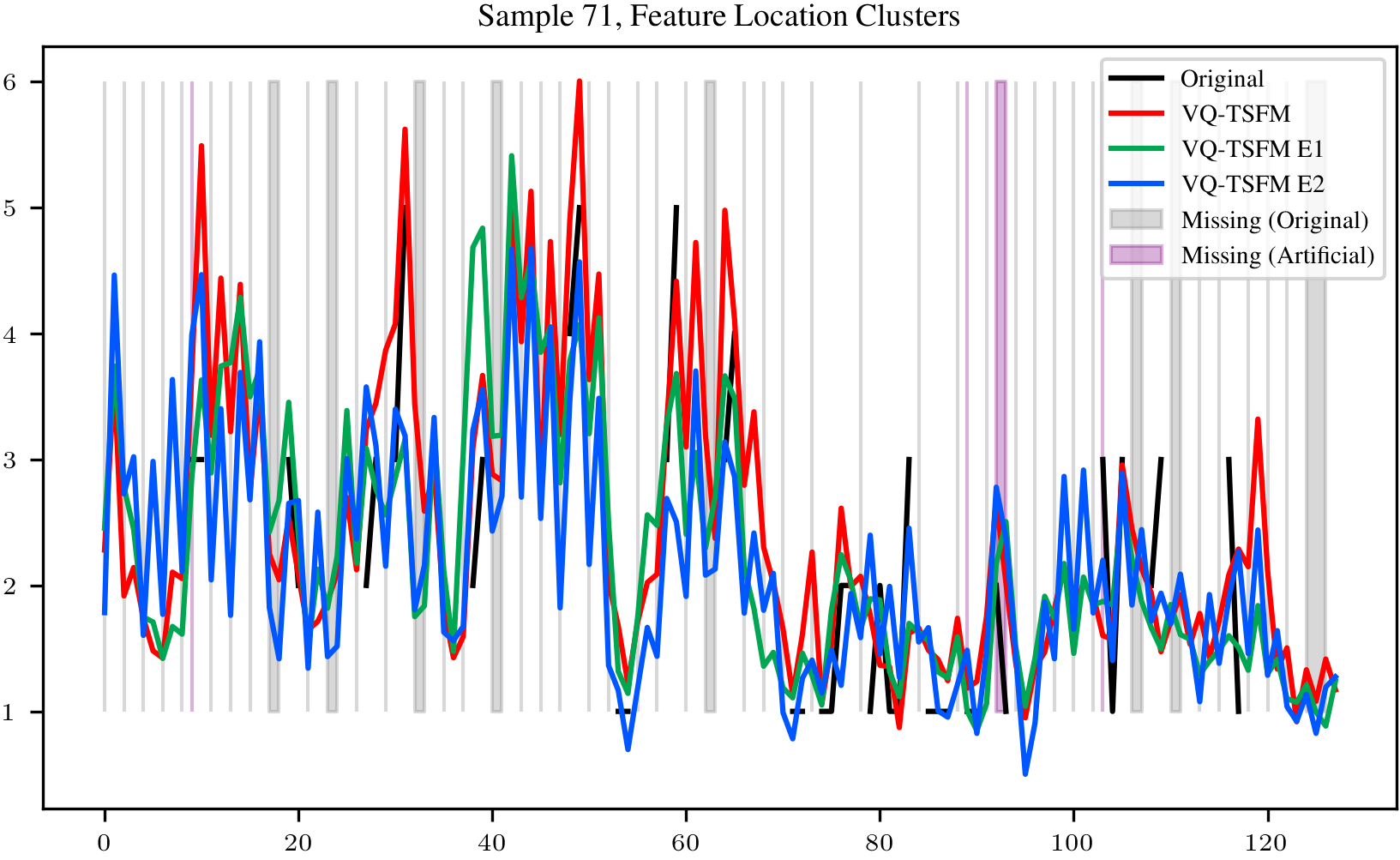}
        \caption{Recons. of sample 71 for Location Clusters.}
        \label{fig:reco_vqvae_s_71_location_clusters}
    \end{subfigure}
    \begin{subfigure}[b]{0.37\textwidth}
        \centering
        \includegraphics[width=\textwidth]{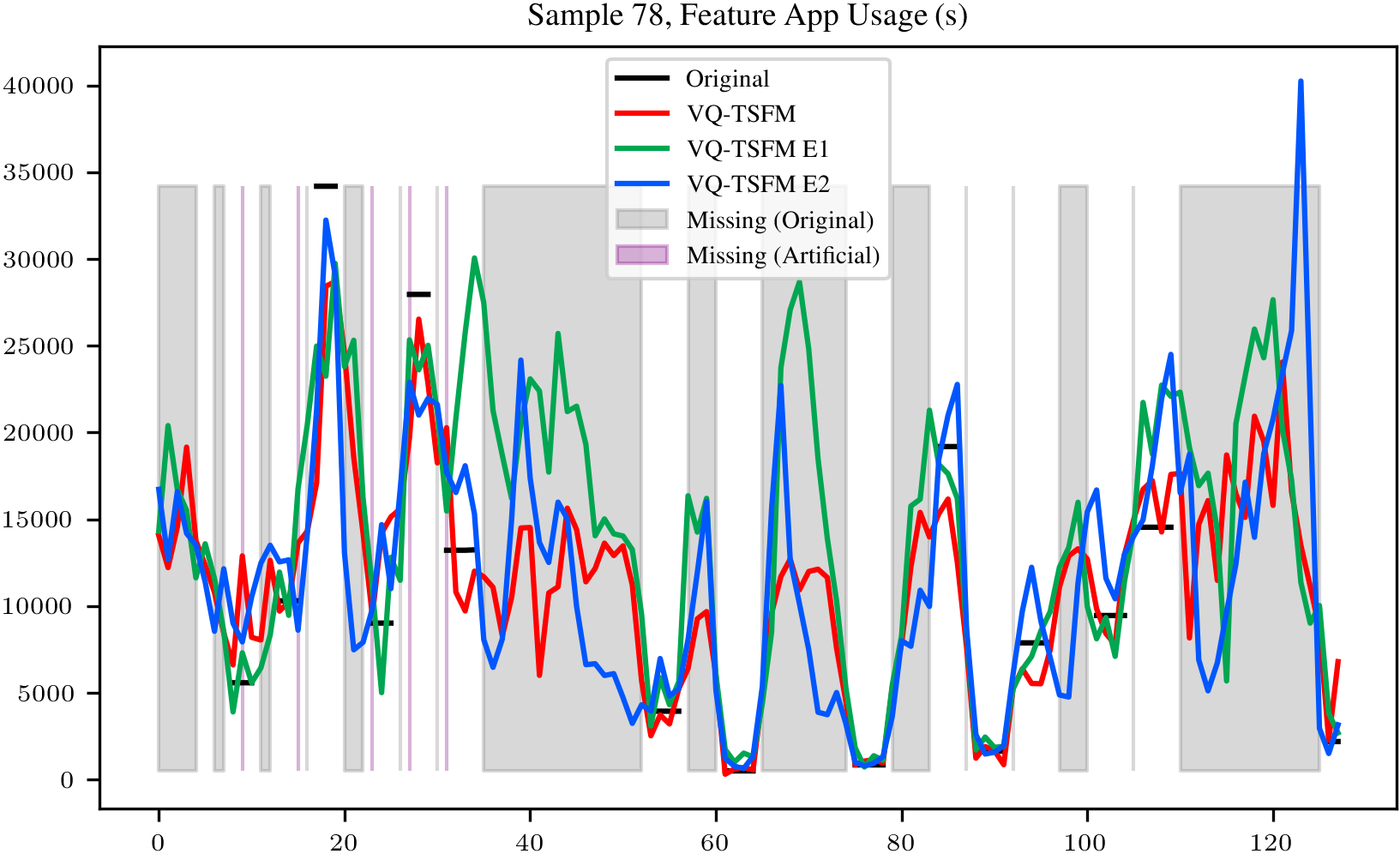}
        \caption{Recons. of sample 78 for App Usage.}
        \label{fig:reco_vqvae_s_78_app_usage}
    \end{subfigure}
    
    \vspace{0.05cm}

    \begin{subfigure}[b]{0.37\textwidth}
        \centering
        \includegraphics[width=\textwidth]{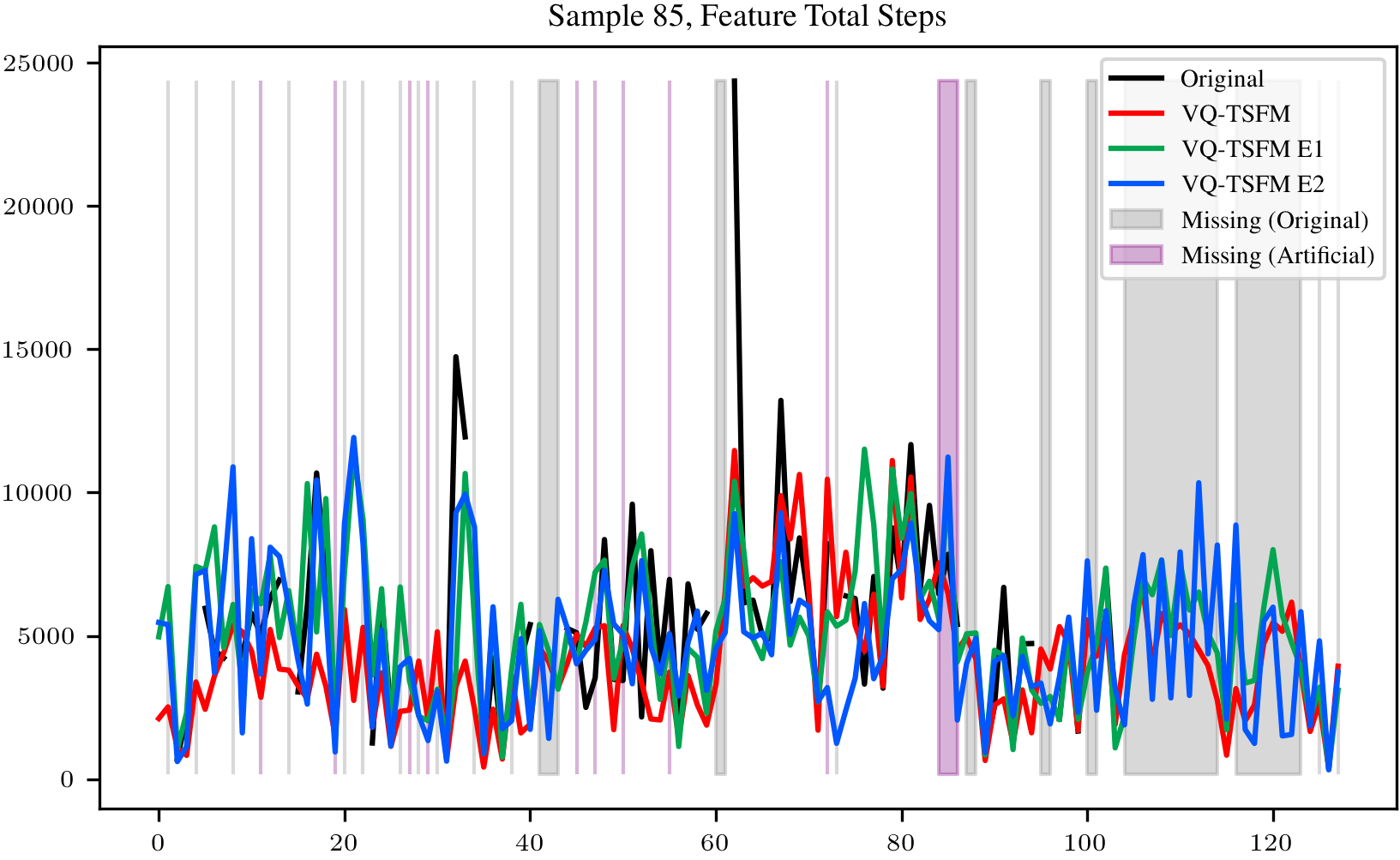}
        \caption{Recons. of sample 85 for Total Steps.}
        \label{fig:reco_vqvae_s_85_total_steps}
    \end{subfigure}
    \begin{subfigure}[b]{0.37\textwidth}
        \centering
        \includegraphics[width=\textwidth]{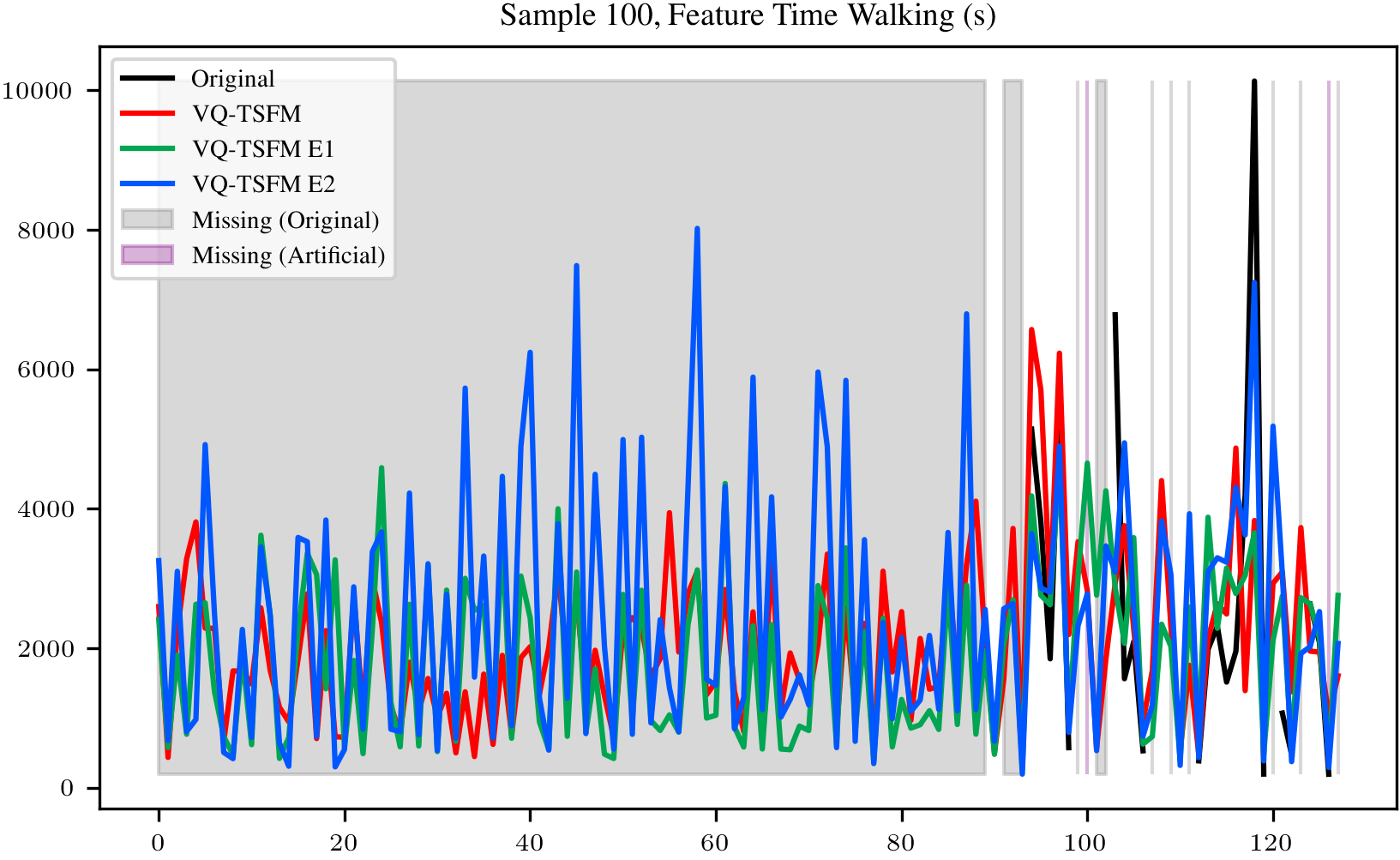}
        \caption{Recons. of sample 100 for Time Walking.}
        \label{fig:reco_vqvae_s_100_time_walking}
    \end{subfigure}
    
    \vspace{0.05cm}

    \begin{subfigure}[b]{0.37\textwidth}
        \centering
        \includegraphics[width=\textwidth]{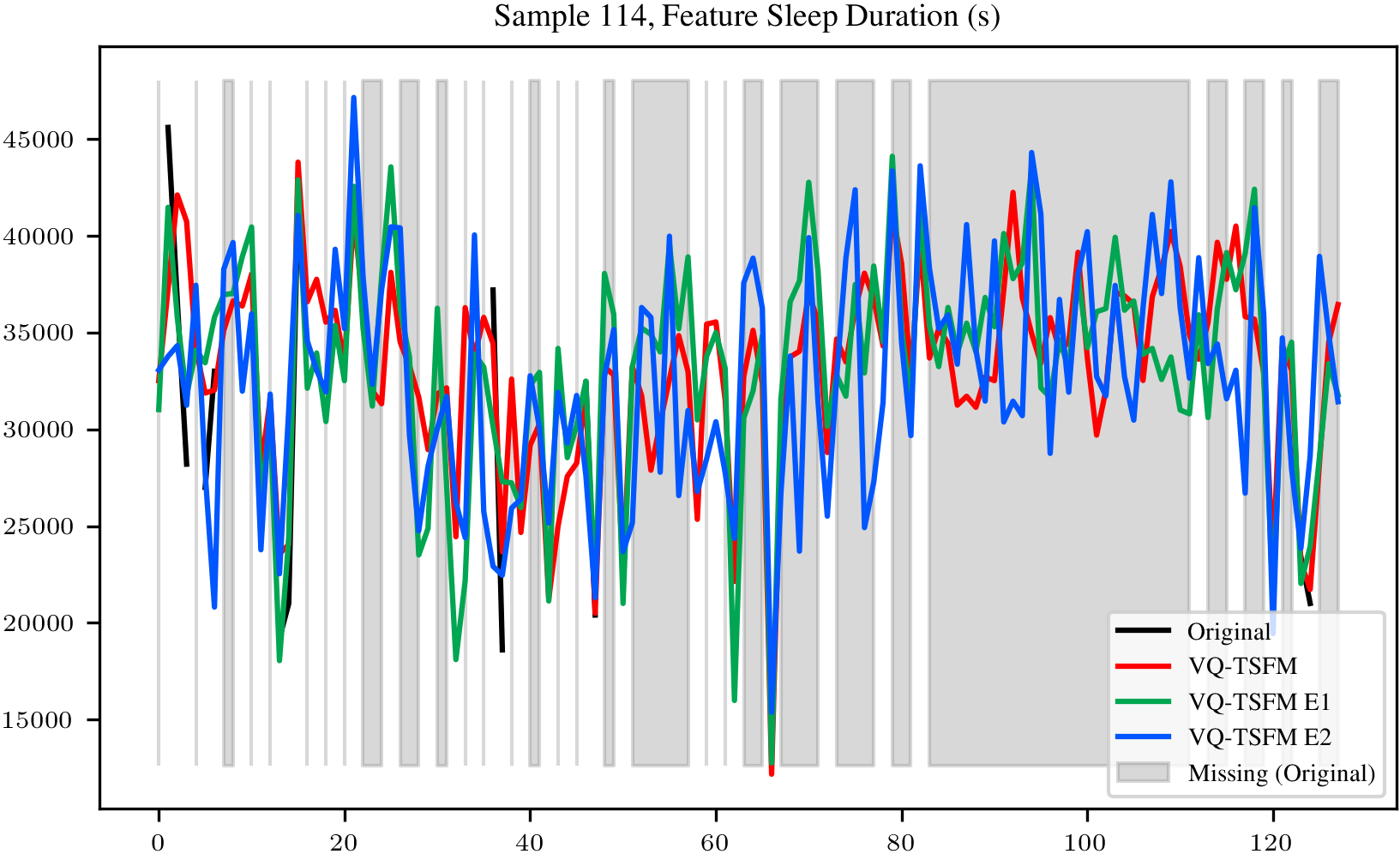}
        \caption{Recons. of sample 114 for Sleep Duration.}
        \label{fig:reco_vqvae_s_114_sleep_duration}
    \end{subfigure}
    \begin{subfigure}[b]{0.37\textwidth}
        \centering
        \includegraphics[width=\textwidth]{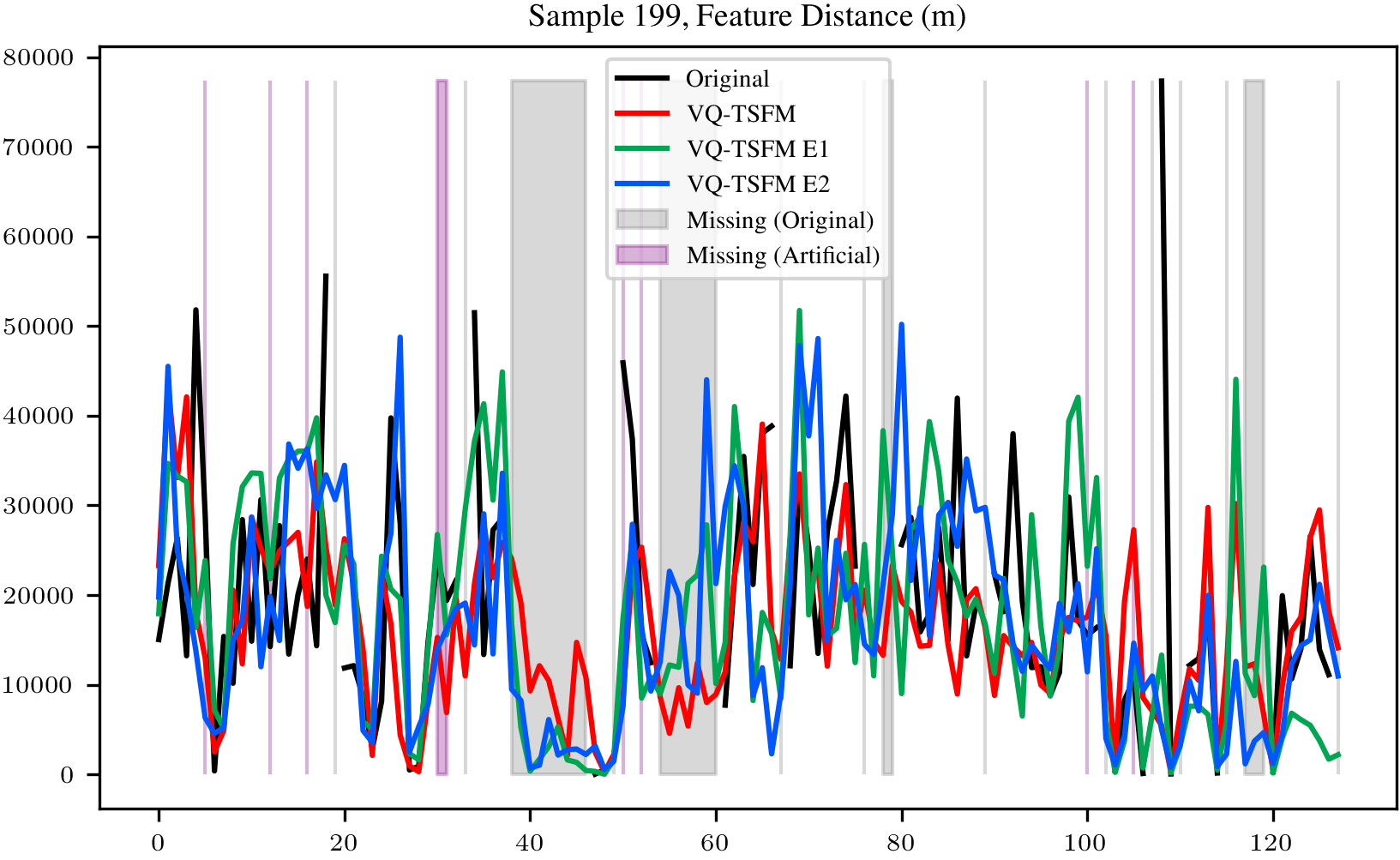}
        \caption{Recons. of sample 199 for Distance.}
        \label{fig:reco_vqvae_s_199_distance}
    \end{subfigure}
    
    \vspace{0.05cm}

    \begin{subfigure}[b]{0.37\textwidth}
        \centering
        \includegraphics[width=\textwidth]{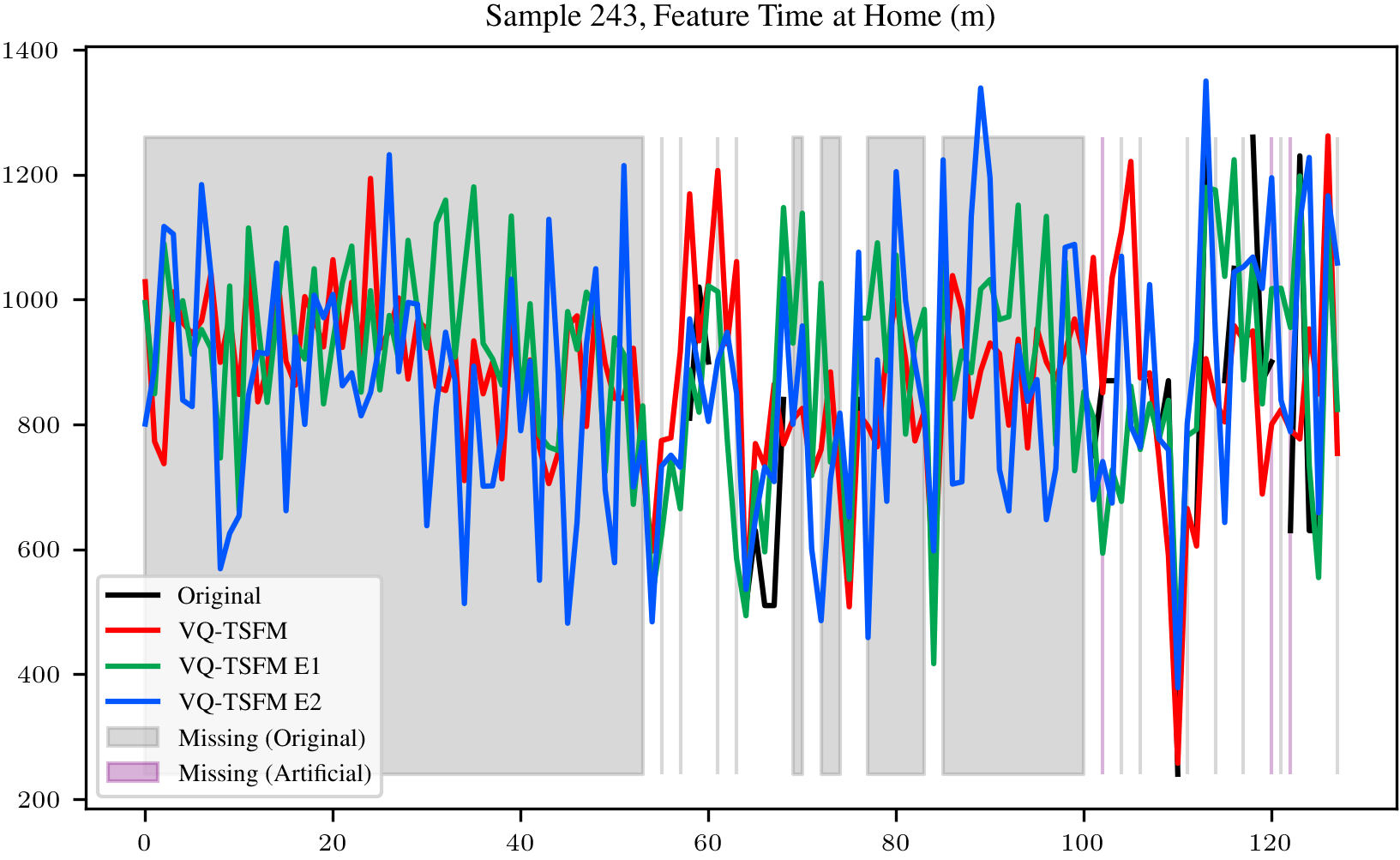}
        \caption{Recons. of sample 243 for Time at Home.}
        \label{fig:reco_vqvae_s_243_time_at_home}
    \end{subfigure}
    \begin{subfigure}[b]{0.37\textwidth}
        \centering
        \includegraphics[width=\textwidth]{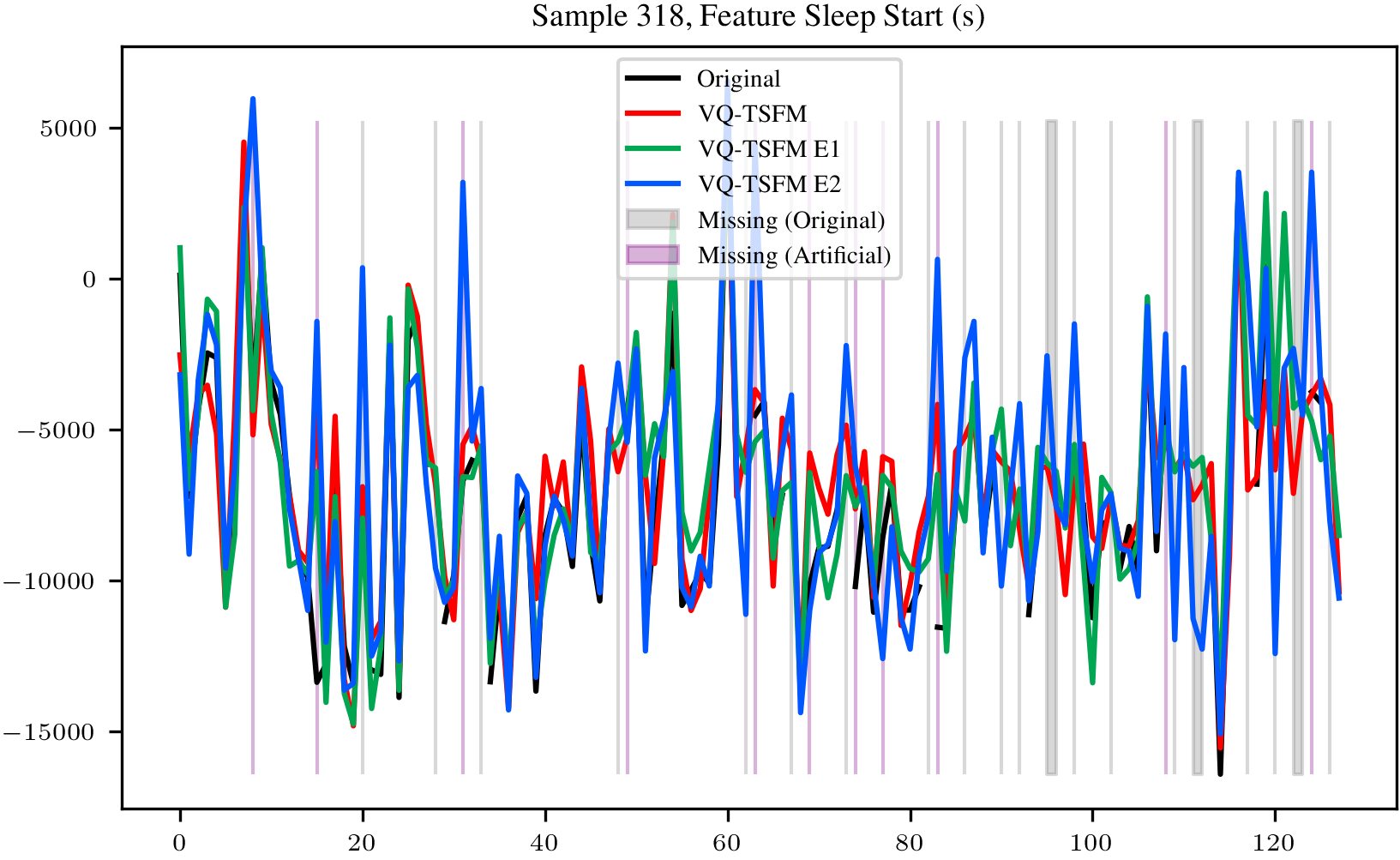}
        \caption{Recons. of sample 318 for Sleep Start.}
        \label{fig:reco_vqvae_s_318_sleep_start}
    \end{subfigure}
    
    \caption{Representative signal reconstructions for observed and imputed instances. In cases where the original signal is not explicitly shown, it is because one or more of the models (whose reconstructions are plotted) overlap the true signal precisely, obscuring the original data.}
    \label{fig:reco_vqvae_plots_app}
\end{figure*}

\autoref{fig:reco_vqvae_plots_app} presents reconstructed and imputed sample examples, where white shading indicates observed data, gray shading denotes originally missing data, and purple shading represents synthetically induced missingness. The remaining time steps (in this case, days) are fully visible to the model. When the original signal is obscured in observed intervals, it is due to one or more model reconstructions perfectly overlapping the true signal, demonstrating accurate recovery. As shown in Figures \ref{fig:reco_vqvae_s_48_weekend} and \ref{fig:reco_vqvae_s_63_practiced_sport} all models perform well with binary variables. 

Notably, the proposed VQ-TSFM and its variants E1 and E2 exhibit strong imputation capabilities even under high proportions of missingness, as evidenced by Figures \ref{fig:reco_vqvae_s_78_app_usage}, \ref{fig:reco_vqvae_s_100_time_walking}, \ref{fig:reco_vqvae_s_243_time_at_home}, and \ref{fig:reco_vqvae_s_114_sleep_duration}. Whether the missing data spans large temporal segments (e.g., the first three-quarters of the sample in \autoref{fig:reco_vqvae_s_100_time_walking}), appears centrally (\autoref{fig:reco_vqvae_s_243_time_at_home}), or is intermittently distributed (\autoref{fig:reco_vqvae_s_114_sleep_duration}), the models consistently maintain robust representations and plausible imputations. This performance generalizes across all variable types---continuous real-valued, continuous positive, count data, and binary---highlighting the versatility of the models across different data ranges and types. The implicit model---the simplest of the three proposed variants---exhibited performance on par with or better than that of E1 and E2. This indicates that, even without explicit conditioning on the missingness mask, VQ-TSFM effectively captures and reconstructs the underlying patterns of missing data.

\subsection{Embedding Usage Histograms} \label{app:embed_usage_hist}

\begin{figure*}[t]
    \centering
    \begin{subfigure}[b]{0.45\textwidth}
        \centering
        \includegraphics[width=\textwidth]{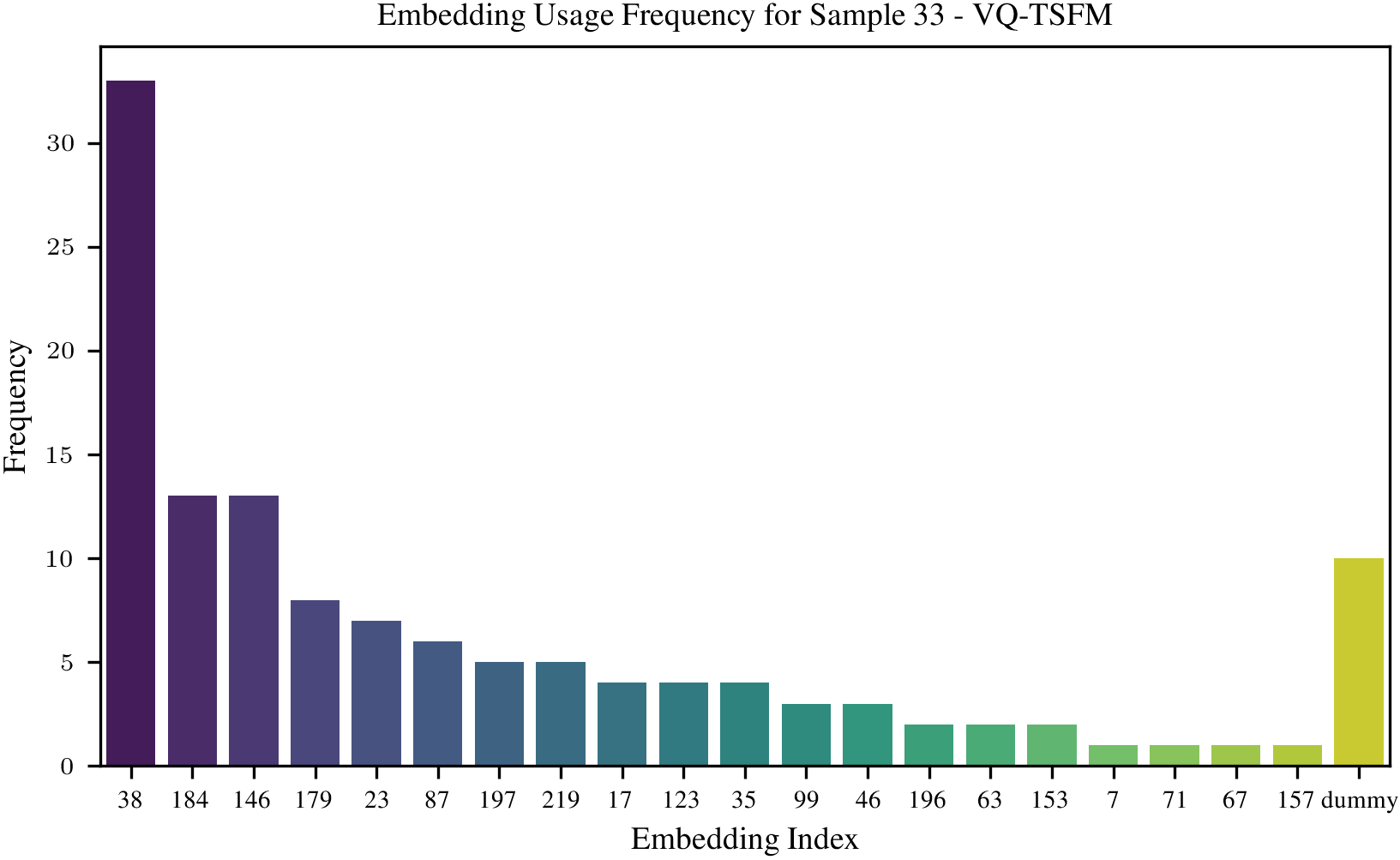}
        \caption{Embedding usage for sample 33 by VQ-TSFM.}
        \label{fig:embed_usage_s_33_a0}
    \end{subfigure}
    \begin{subfigure}[b]{0.45\textwidth}
        \centering
        \includegraphics[width=\textwidth]{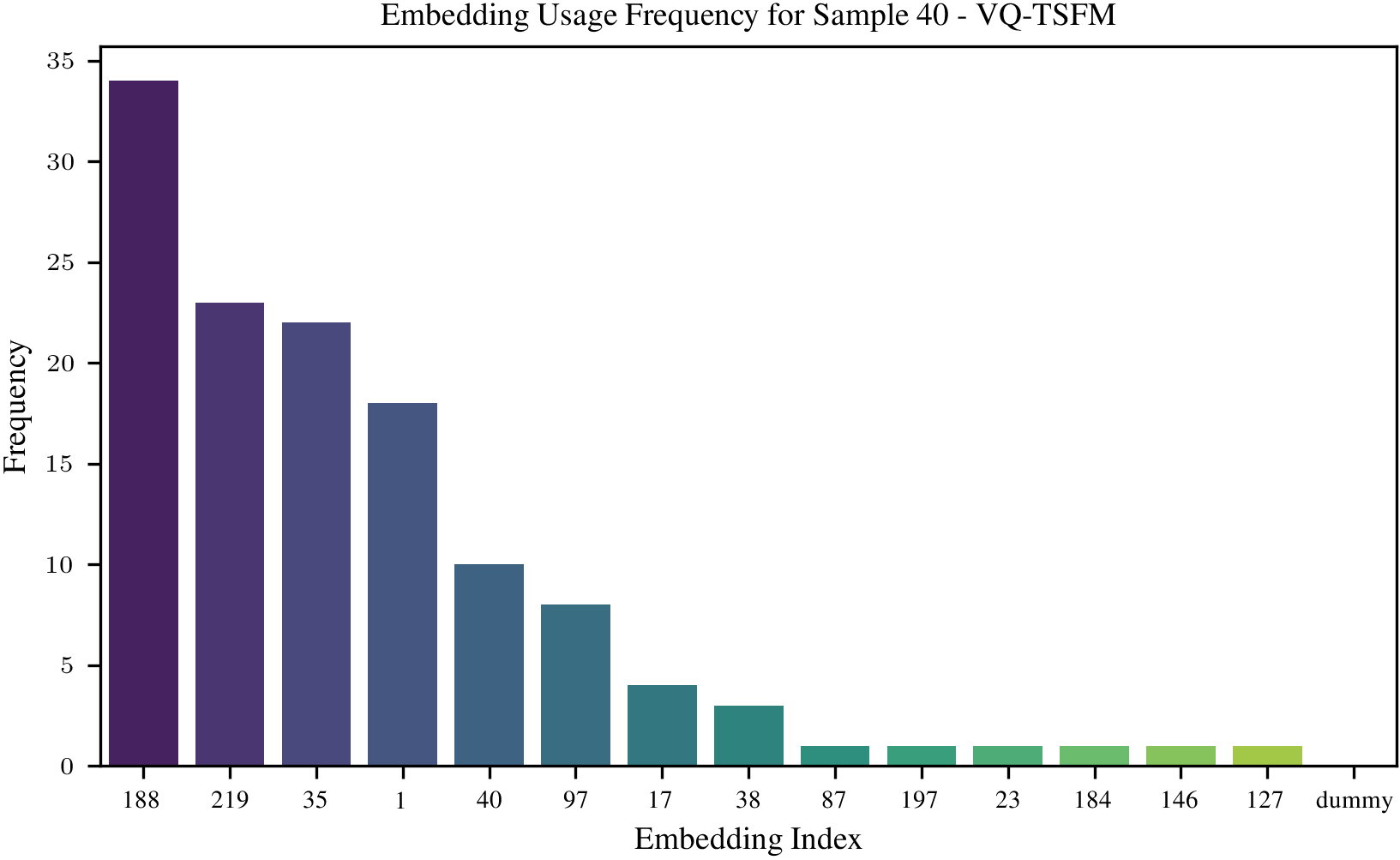}
        \caption{Embedding usage for sample 40 by VQ-TSFM.}
        \label{fig:embed_usage_s_40_a0}
    \end{subfigure}

    \vspace{0.08cm}

    \begin{subfigure}[b]{0.45\textwidth}
        \centering
        \includegraphics[width=\textwidth]{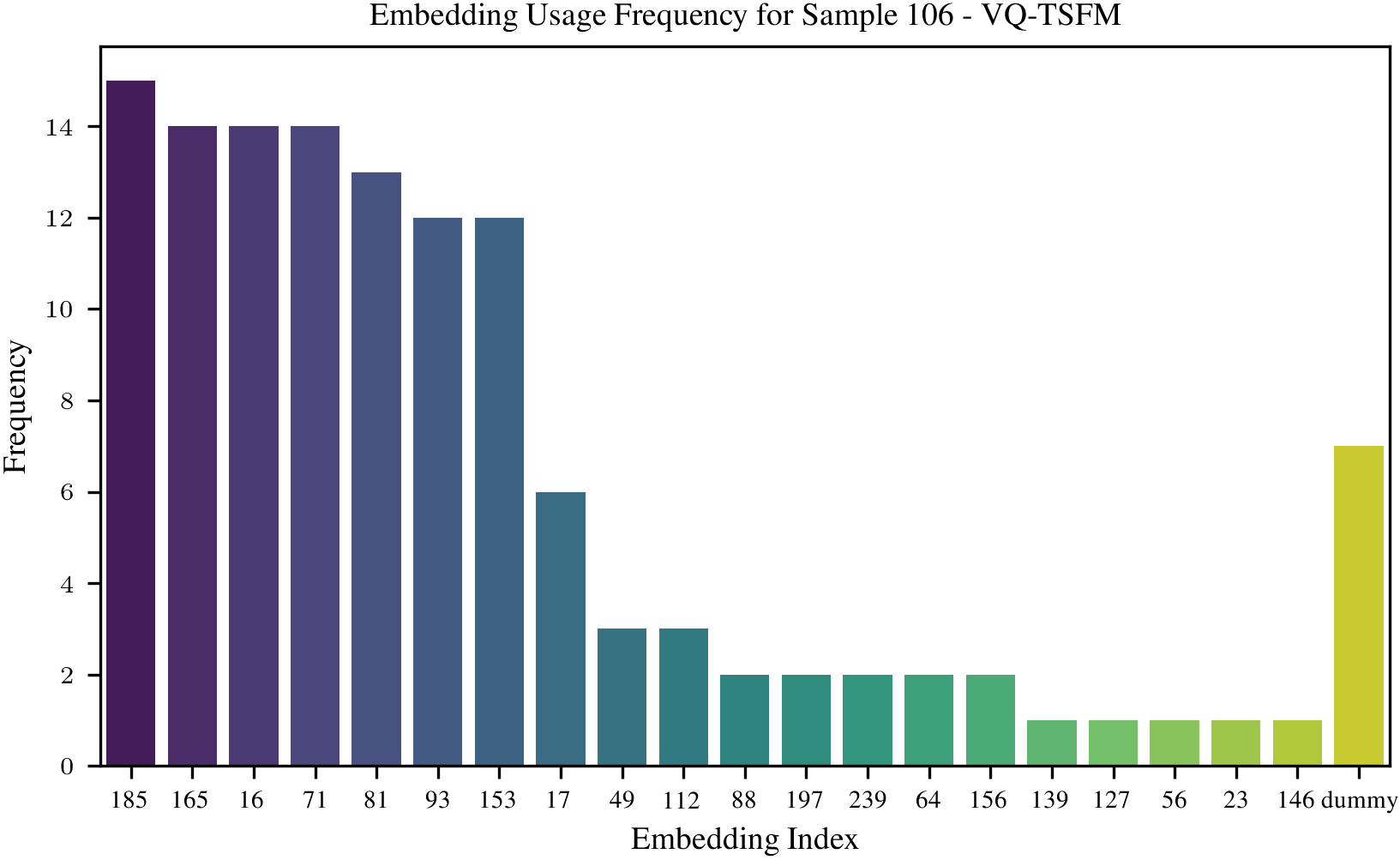}
        \caption{Embedding usage for sample 106 by VQ-TSFM.}
        \label{fig:embed_usage_s_106_a0}
    \end{subfigure}
    \begin{subfigure}[b]{0.45\textwidth}
        \centering
        \includegraphics[width=\textwidth]{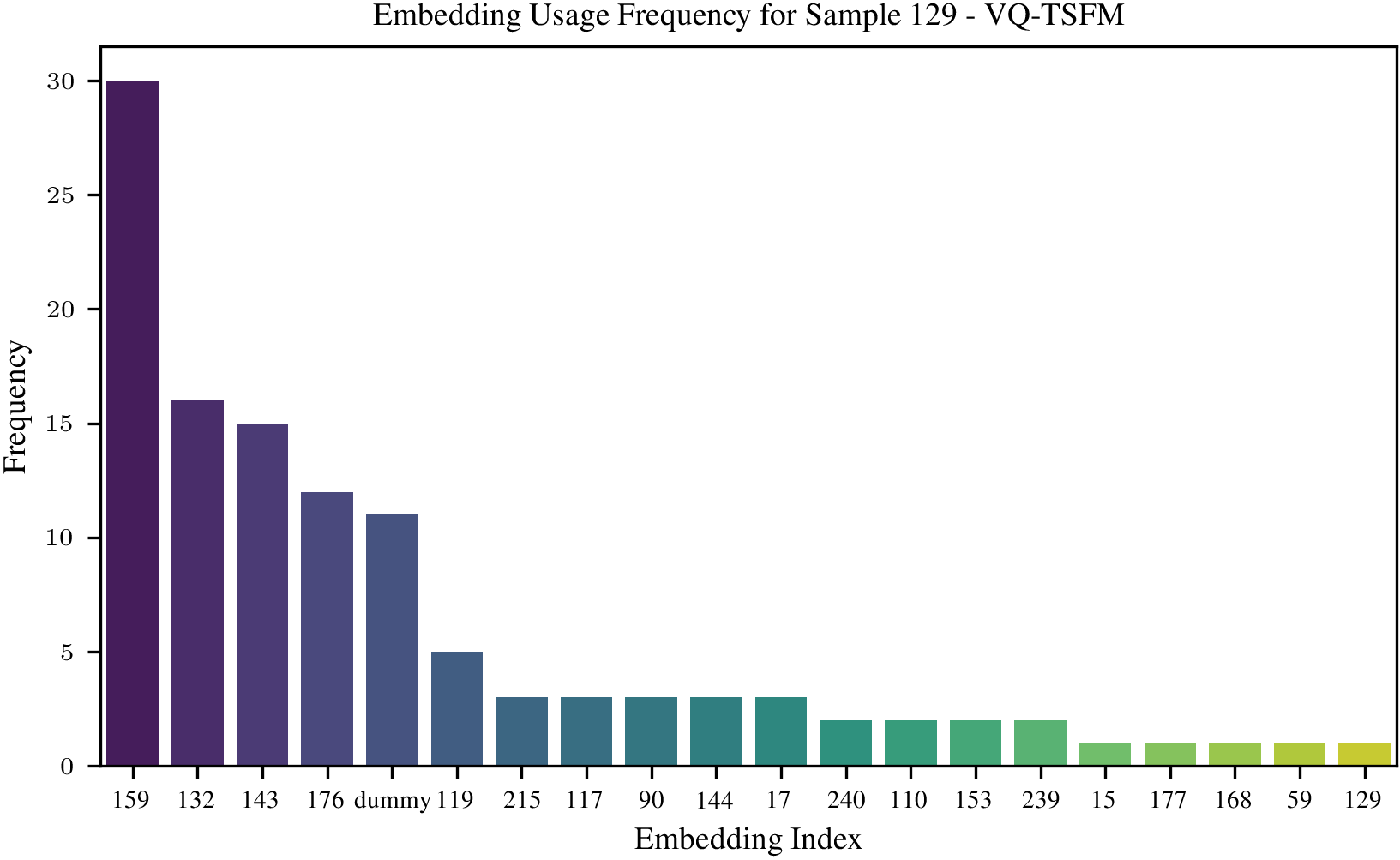}
        \caption{Embedding usage for sample 129 by VQ-TSFM.}
        \label{fig:embed_usage_s_129_a0}
    \end{subfigure}

    \vspace{0.08cm}

    \begin{subfigure}[b]{0.45\textwidth}
        \centering
        \includegraphics[width=\textwidth]{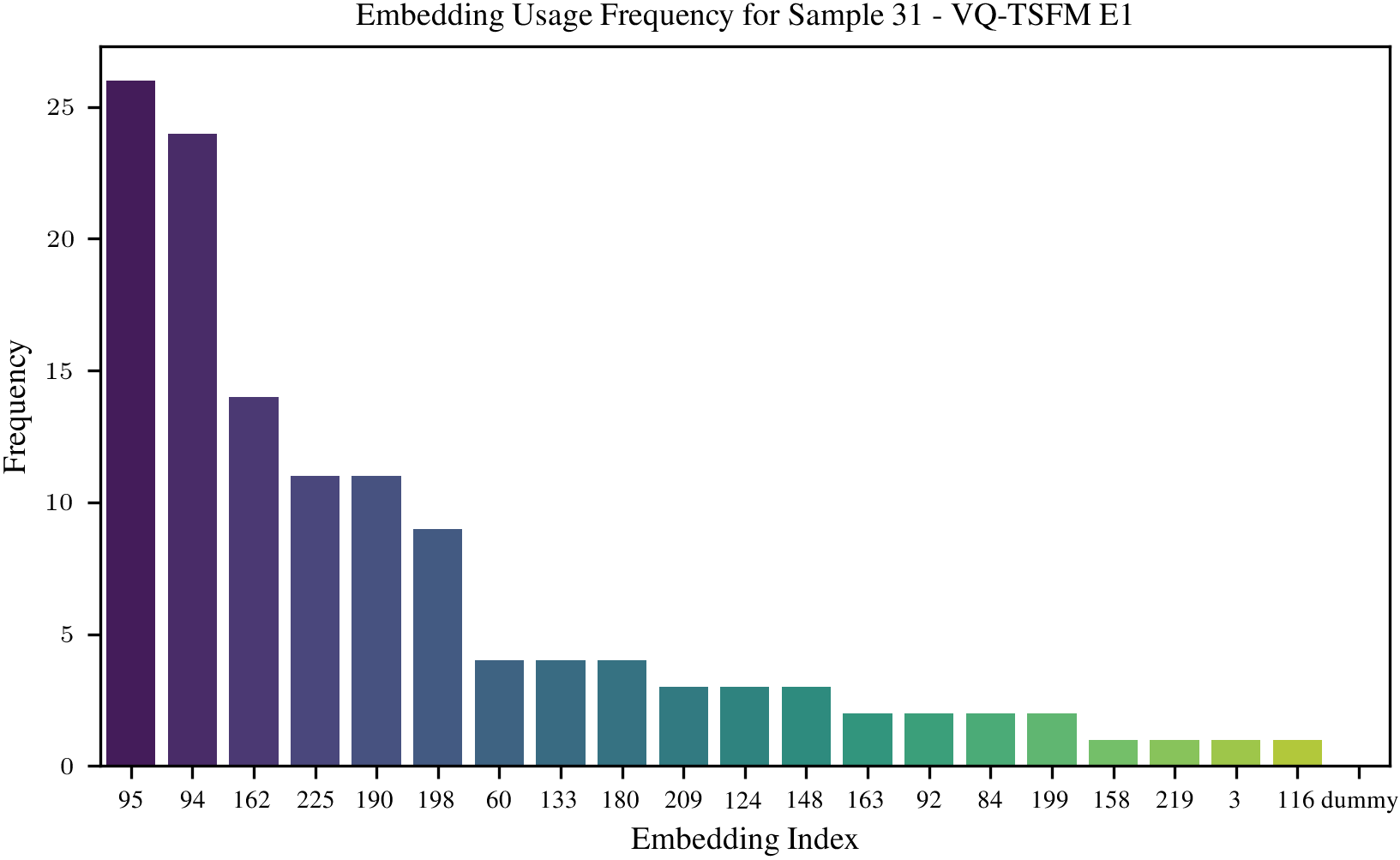}
        \caption{Embedding usage for sample 31 by E1.}
        \label{fig:embed_usage_s_31_a1}
    \end{subfigure}
    \begin{subfigure}[b]{0.45\textwidth}
        \centering
        \includegraphics[width=\textwidth]{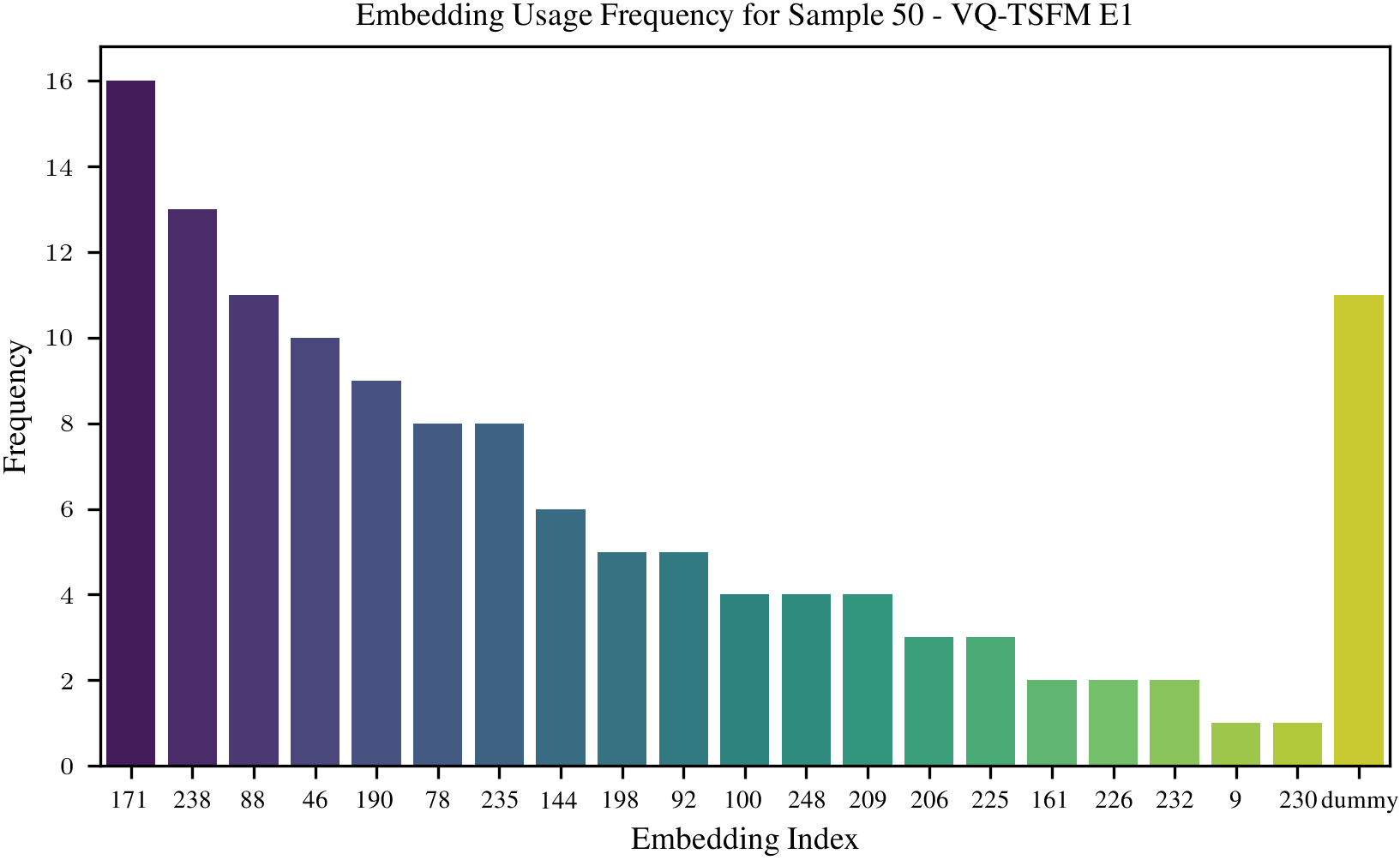}
        \caption{Embedding usage for sample 50 by E1.}
        \label{fig:embed_usage_s_50_a1}
    \end{subfigure}

    \vspace{0.08cm}

    \begin{subfigure}[b]{0.45\textwidth}
        \centering
        \includegraphics[width=\textwidth]{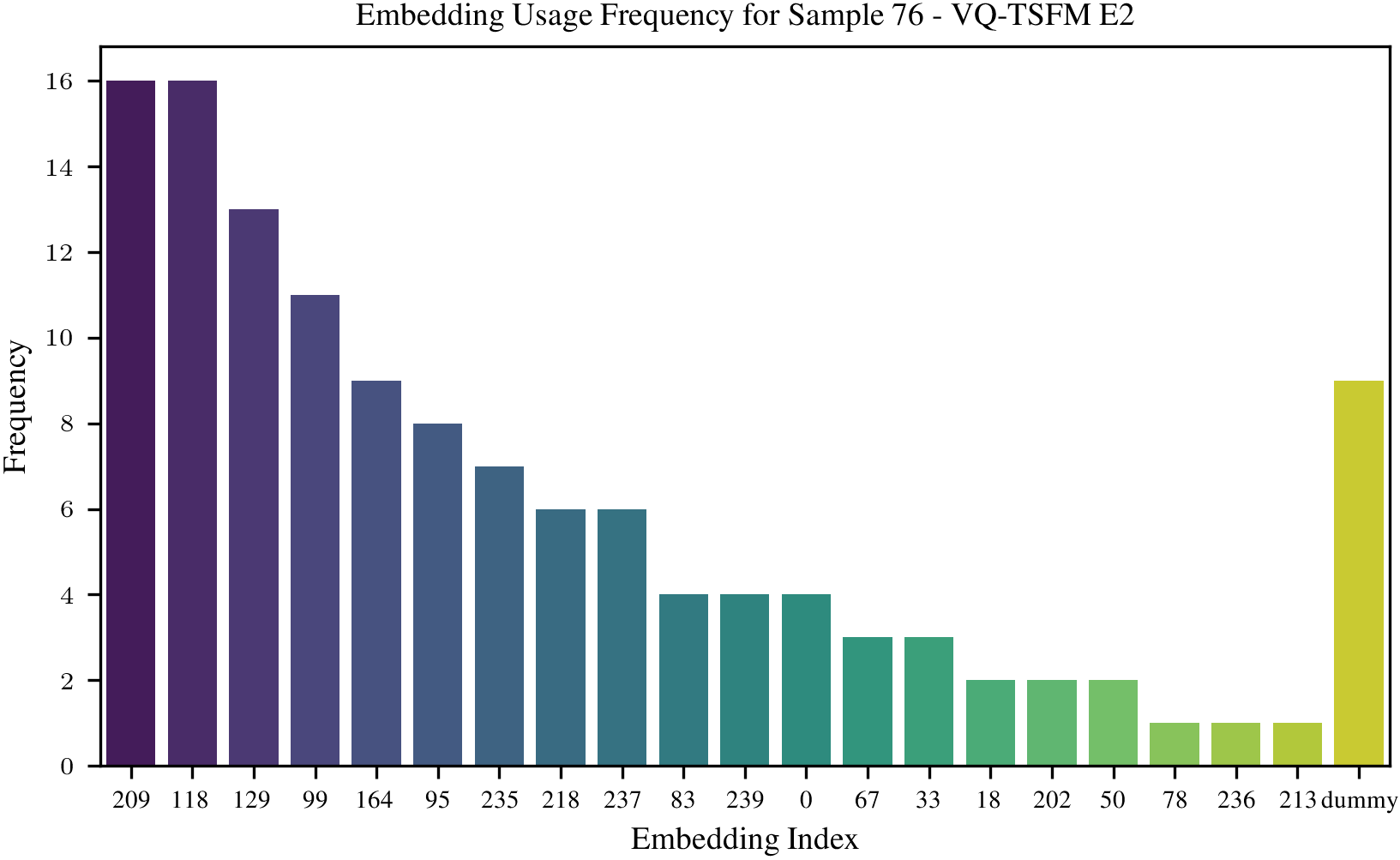}
        \caption{Embedding usage for sample 76 by E2.}
        \label{fig:embed_usage_76_a2}
    \end{subfigure}
    \begin{subfigure}[b]{0.45\textwidth}
        \centering
        \includegraphics[width=\textwidth]{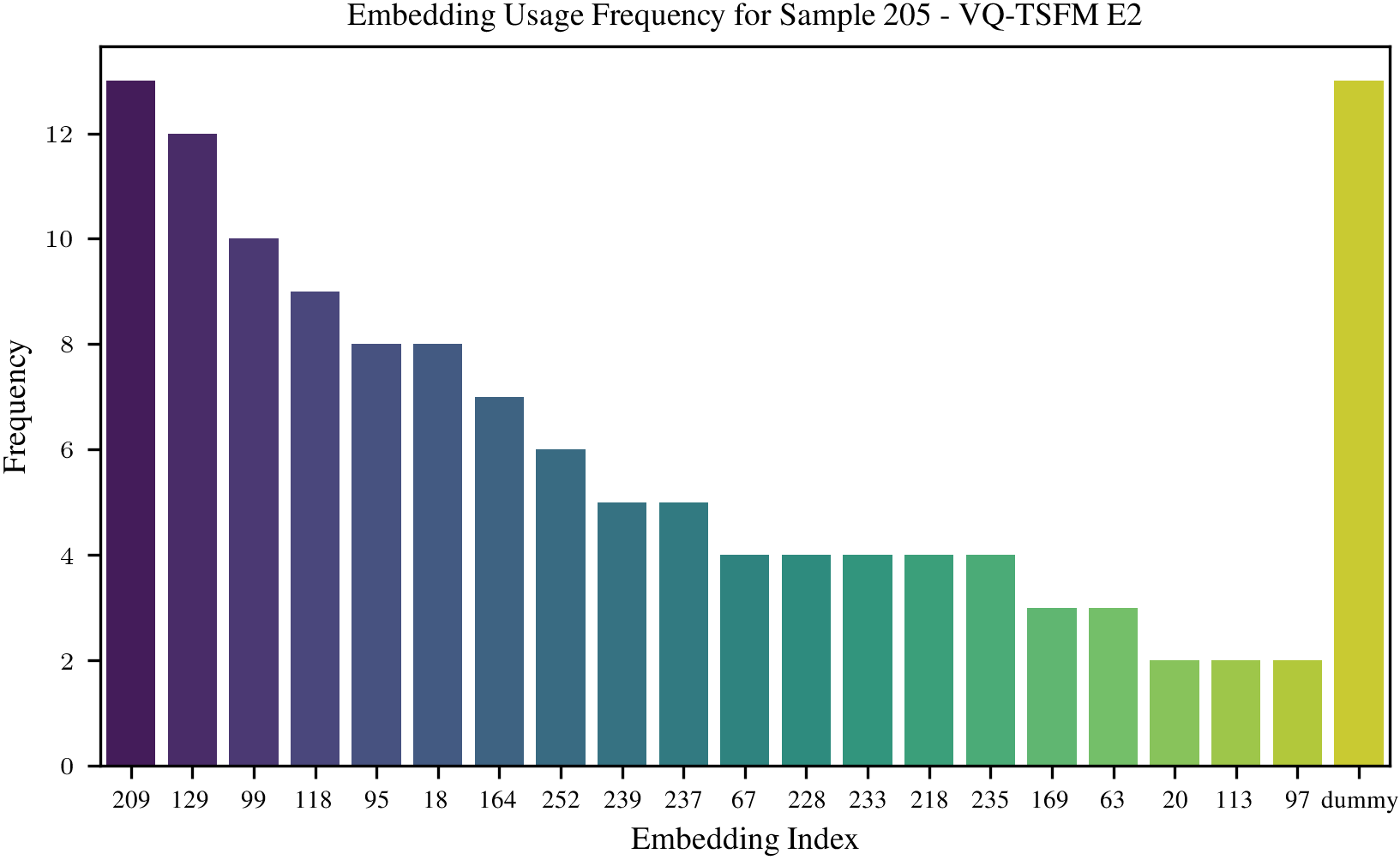}
        \caption{Embedding usage for sample 205 by E2.}
        \label{fig:embed_usage_205_a2}
    \end{subfigure}

    \caption{Embedding usage histograms for different samples. Out of the total 256 available embeddings, we observe that only a small subset is typically used, resulting in a sparse and more interpretable solution. Embeddings that are individually uncommon are categorized as belonging to the "dummy" embedding, emphasizing the model’s focus on a limited number of relevant embeddings.}
    \label{fig:embedding_histograms}
\end{figure*}

The discrete quantization of VQ-TSFM facilitates the construction of latent representations, making it particularly suited for applications that benefit from codifying instances, as demonstrated in this work. Unlike traditional methods that rely on handcrafted features---often tailored to individual patients and limiting generalizability---VQ-TSFM learns patient-agnostic embeddings, enabling generalization across subpopulations and tasks. These embeddings can be effectively applied to tasks such as time-series data imputation and extended to critical downstream tasks, such as identifying critical health events or suicide risk detection. As illustrated in \autoref{fig:embedding_histograms}, the usefulness of these embeddings is enhanced by their sparsity---typically, only a small subset of the 256 available embeddings is used per sample. This results in a more interpretable solution, with infrequent embeddings classified as "dummy" embeddings, which can themselves acquire meaningful interpretations (e.g., representing rare or unstable states). In turn, this sparsity in then leveraged to provide contained, yet expressive profiles sequences for the CPD algorithm, as discussed in \autoref{app:vqvae_latent_cpd}.

\begin{table}[ht]
\centering
\setlength{\tabcolsep}{3pt}
\renewcommand{\arraystretch}{0.85}
\caption{Paired Welch's t-test results comparing variant models E1 and E2 to the VQ-TSFM baseline}

\label{tab:paired_welch_ttest}
\begin{tabular}{@{} llcccccc @{}}
\toprule
\textbf{Variable} & \textbf{Type} & \textbf{VQ-TSFM vs E1 ($t$)} & \textbf{VQ-TSFM vs E1 ($p$)} & \textbf{VQ-TSFM vs E2 ($t$)} & \textbf{VQ-TSFM vs E2 ($p$)} \\
\midrule

\multirow{3}{*}{\textbf{Sleep Start (s)}} 
  & XO      & $-6.1860$ & $3\times 10^{-8}$ (\ding{55})  & $-11.7016$  & $1.4\times 10^{-18}$ (\ding{55}) \\
  & MCAR    & $-2.3585$  & $0.0209$ (\ding{55}) & $ -2.2937$  & $0.0257$ (\ding{55}) \\
  & MNAR    & ---  & ---  & $-2.3697$  & $0.0203$ (\ding{55}) \\
\midrule

\multirow{3}{*}{\textbf{Distance (m)}} 
  & XO      & ---  & --- & ---  & --- \\
  & MCAR    & ---  & --- & ---  & --- \\
  & MNAR    & $-2.0102$  & $0.0479$ (\ding{55}) & $0.1517$  & $0.8798$ \\
\midrule

\multirow{3}{*}{\textbf{Time at Home (m)}} 
  & XO      & $-1.6511$  & $0.1037$  & $16.7191$  & $7.4\times 10^{-24}$ (\ding{55}) \\
  & MCAR    & ---  & ---  & $0.5906$  & $0.5564$ \\
  & MNAR    & $-1.0755$  & $0.2854$  & $-0.4124$  & $0.6812$ \\
\midrule

\multirow{3}{*}{\textbf{Sleep Duration (s)}} 
  & XO      & $-3.0788$  & $0.0029$ (\ding{55})  & $22.2654$  & $2.6\times 10^{-31}$ (\ding{55}) \\
  & MCAR    & $0.7896$  & $0.4322$  & $2.2603$  & $0.0268$ (\ding{55}) \\
  & MNAR    & $-0.7592$  & $0.4503$  & $2.2641$  & $0.0264$ (\ding{55}) \\
\midrule

\multirow{3}{*}{\textbf{Time Walking (s)}} 
  & XO      & $-3.0425$  & $0.0031$ (\ding{55})  & $-4.1449$  & $8.6\times 10^{-5}$ (\ding{55}) \\
  & MCAR    & ---  & --- & ---  & --- \\
  & MNAR    & ---  & ---  & $3.1853$  & $0.0021$ (\ding{55}) \\
\midrule

\multirow{3}{*}{\textbf{App Usage (s)}} 
  & XO      & $-0.9368$  & $0.3518$  & $2.3289$  & $0.0225$ (\ding{55}) \\
  & MCAR    & $-0.5927$  & $0.5551$  & ---  & --- \\
  & MNAR    & ---  & --- & ---  & --- \\
\midrule

\multirow{3}{*}{\textbf{Location Clusters}} 
  & XO      & $-0.8132$  & $0.4186$  & $8.2048$  & $6.9\times 10^{-12}$ (\ding{55}) \\
  & MCAR    & $-0.3650$  & $0.7160$  & ---  & --- \\
  & MNAR    & $-0.7398$  & $0.4616$ & $1.5771$  & $0.1189$ \\
\midrule

\multirow{3}{*}{\textbf{Total Steps}} 
  & XO      & $-0.1357$  & $0.8924$  & $5.2860$  & $1.1\times 10^{-6}$ (\ding{55}) \\
  & MCAR    & ---  & --- & ---  & --- \\
  & MNAR    & $-1.6286$  & $0.1078$  & $-1.7023$  & $0.0929$ \\
\midrule

\multirow{1}{*}{\textbf{Weekend}} 
  & XO      & $3.6438$  & $0.0005$ (\ding{55})  & $6.3882$  & $1.5\times 10^{-8}$ (\ding{55}) \\
\midrule

\multirow{1}{*}{\textbf{Practiced Sport}} 
  & XO      & ---  & ---  & ---  & --- \\
\bottomrule
\multicolumn{6}{p{0.95\linewidth}}{The table reports the test statistic (\textit{t}) and \textit{p}-values for each model and variable under different conditions (XO, MCAR, and MNAR). $\alpha = 0.05$ was used and \ding{55} denotes the rejection of the null hypothesis at the $\alpha$ significance level.}
\end{tabular}
\end{table}

\begin{table}[ht]
\centering
\setlength{\tabcolsep}{3pt}
\renewcommand{\arraystretch}{0.85}
\caption{Wilcoxon signed-rank test results comparing variant models E1 and E2 to the VQ-TSFM baseline}
\label{tab:paired_wilcoxon}
\begin{tabular}{@{} llcccccc @{}}
\toprule
\textbf{Variable} & \textbf{Type} & \textbf{VQ-TSFM vs E1 ($t$)} & \textbf{VQ-TSFM vs E1 ($p$)} & \textbf{VQ-TSFM vs E2 ($t$)} & \textbf{VQ-TSFM vs E2 ($p$)} \\
\midrule

\multirow{3}{*}{\textbf{Sleep Start (s)}} 
  & XO      & --- & ---  & ---  & --- \\
  & MCAR    & ---  & --- & ---  & --- \\
  & MNAR    & $272.0$  & $0.0641$  & ---  & --- \\
\midrule

\multirow{3}{*}{\textbf{Distance (m)}} 
  & XO      & $217.0$ & $0.0086$ (\ding{55})  & $200.0$  & $0.0041$ (\ding{55}) \\
  & MCAR    & $263.0$  & $0.0482$ (\ding{55}) & $353.0$  & $0.4517$ \\
  & MNAR    & ---  & ---  & ---  & --- \\
\midrule

\multirow{3}{*}{\textbf{Time at Home (m)}} 
  & XO      & --- & ---  & ---  & --- \\
  & MCAR    & $394.0$  & $0.8368$ & ---  & --- \\
  & MNAR    & ---  & ---  & ---  & --- \\
\midrule

\multirow{3}{*}{\textbf{Sleep Duration (s)}} 
  & XO      & --- & ---  & ---  & --- \\
  & MCAR    & ---  & --- & ---  & --- \\
  & MNAR    & ---  & ---  & ---  & --- \\
\midrule

\multirow{3}{*}{\textbf{Time Walking (s)}} 
  & XO      & --- & ---  & ---  & --- \\
  & MCAR    & $333.0$  & $0.3074$ & $310.0$  & $0.1831$ \\
  & MNAR    & $301.0$  & $0.1461$  & ---  & --- \\
\midrule

\multirow{3}{*}{\textbf{App Usage (s)}} 
  & XO      & --- & ---  & ---  & --- \\
  & MCAR    & ---  & --- & $301.0$  & $0.1460$ \\
  & MNAR    & $330.0$  & $0.2887$  & $369.0$  & $0.5900$ \\
\midrule

\multirow{3}{*}{\textbf{Location Clusters}} 
  & XO      & --- & ---  & ---  & --- \\
  & MCAR    & ---  & --- & $206.0$  & $0.0053$ \\
  & MNAR    & ---  & ---  & ---  & --- \\
\midrule

\multirow{3}{*}{\textbf{Total Steps}} 
  & XO      & --- & ---  & ---  & --- \\
  & MCAR    & $283.0$  & $0.0892$ & $280.0$  & $0.0817$ \\
  & MNAR    & ---  & ---  & ---  & --- \\
\midrule

\multirow{1}{*}{\textbf{Weekend}} 
  & XO      & ---  & ---  & ---  & --- \\
\midrule

\multirow{1}{*}{\textbf{Practiced Sport}} 
  & XO      & $236.0$  & $0.0185$ (\ding{55})  & $353.0$  & $0.5360$ \\
\bottomrule
\multicolumn{6}{p{0.95\linewidth}}{The table reports the test statistic (\textit{t}) and \textit{p}-values for each model and variable under different conditions (XO, MCAR, and MNAR). $\alpha = 0.05$ was used and \ding{55} denotes the rejection of the null hypothesis at the $\alpha$ significance level.}
\end{tabular}
\end{table}

\newpage

\end{document}